\documentclass{article} \usepackage{iclr2022_conference,times}

\usepackage{amsmath,amsfonts,bm}

\def\Figref#1{Figure~\ref{#1}}

\def\eqref#1{equation~\ref{#1}}
\def\Eqref#1{Equation~\ref{#1}}

\def\1{\bm{1}}

\DeclareMathAlphabet{\mathsfit}{\encodingdefault}{\sfdefault}{m}{sl}
\SetMathAlphabet{\mathsfit}{bold}{\encodingdefault}{\sfdefault}{bx}{n}

\usepackage{hyperref}
\usepackage{url}

\usepackage{wrapfig}
\usepackage{subcaption}
\usepackage{overpic}
\usepackage{graphicx}

\title{Wavelet-based Loss for High-frequency Interface Dynamics}

\author{Lukas Prantl \\
Technical University of Munich\\
\texttt{lukas.prantl@tum.de} \\
\And
\hspace{-50mm}
Jan Bender \\
\hspace{-50mm}
RWTH Aachen University\\
\AND
Tassilo Kugelstadt \\
RWTH Aachen University\\
\And
\hspace{48mm}
Nils Thuerey \\
\hspace{48mm}
Technical University of Munich\\
}

\definecolor{nilscol}{rgb}{0.1,0.75,0.08}

\definecolor{lukcol}{rgb}{0.1,0.15,0.8}

\iclrfinalcopy \begin{document}

\maketitle

\begin{abstract}
Generating highly detailed, complex data is a long-standing and frequently considered problem in the machine learning field. However, developing detail-aware generators remains an challenging and open problem. Generative adversarial networks are the basis of many state-of-the-art methods. However, they introduce a second network to be trained as a loss function, making the interpretation of the learned functions much more difficult. As an alternative, we present a new method based on a wavelet loss formulation, which remains transparent in terms of what is optimized. The wavelet-based loss function is used to overcome the limitations of conventional distance metrics, such as L1 or L2 distances, when it comes to generate data with high-frequency details. 
We show that our method can successfully reconstruct high-frequency details in an illustrative synthetic test case. Additionally, we evaluate the performance when applied to more complex surfaces based on physical simulations.
Taking a roughly approximated simulation as input, our method infers corresponding spatial details while taking into account how they evolve. We consider this problem in terms of spatial and temporal frequencies, and leverage generative networks trained with our wavelet loss to learn the desired spatio-temporal signal for the surface dynamics. We 
test the capabilities of our method with a set of synthetic wave function tests and complex 2D and 3D dynamics of elasto-plastic materials.
\end{abstract}

\section{Introduction}

In the context of generative neural networks, simple distance metrics such as L1 or L2 distances play an important role. They often serve as a loss function to validate the output in relation to the ground truth. However, these conventional distance metrics usually act like a low-pass filter.
Small high-frequency details are thus weighted too weakly and the results become smooth and can sometimes be very different from sharp, detailed ground-truth data.
This is especially true for multi-modal problems that are ill-conditioned and where features cannot be unambiguously reconstructed from the information-poor input. Representative for this are super-resolution tasks, where high-resolution data are to be reconstructed from low-resolution data \citep{johnson2016perceptual, dosovitskiy2016perceptual, bruna2016sres}.
To counteract the limitations of conventional distance functions, we introduce a novel wavelet-based loss function.

Our central motivation is to put more emphasis on high-frequency features.
For the proposed wavelet loss, we therefore first transform the data into wavelet space and then scale it such  that the focus lies on the high-frequency content. This successfully negates the low-pass effect of the simpler distance metrics.
We choose the wavelet transform \citep{daubechies1992ten} because it better fits the nature of many natural data sets,
where high-frequency details do not appear uniformly, but instead are sporadic and isolated. The first choice for frequency analysis would usually be a Fourier transform, but this usually expects periodic signals whose frequency does not change significantly over time. An adaptation of the Fourier transform, the so-called \textit{short-time} Fourier transform, attempts to circumvent this limitation by applying Fourier transforms to localized, smoothed time slices of the signal. However, this leads to limitations in the frequency and time resolution, which the wavelet transform does not have.
Due to its advantages, the wavelet transform is widely used in the field of image compression (e.g. JPEG 2000). The superiority of the wavelet transform in this area has accordingly been demonstrated many times \citep{santacruz2002jpeg, woodring2011compress, lewis1992compress}.
In summary, the wavelet transform allows us to process non-periodic signals, independent of position and scaling with respect to their frequency. This allows us to compensate for negative effects that affect frequency, such as the low-pass effect mentioned above.

To 
evaluate the effectiveness of our method, we consider challenging super-resolution problems with complex data sets from physical elasto-plastic simulations. 
Super-resolution methods are a common technique 
to speed up the necessary calculations and to allow for more control.
This can be seen as a form of post-processing where one simulates only a low-resolution simulation and uses an up-sampling technique to approximate the behavior of a costly high-resolution simulation. 
Neural networks are of special interest here because of their capability to efficiently approximate the strongly nonlinear behavior of physical simulations. 
Applying neural networks to space-time data sets 
of physical simulations has seen strongly growing interest in recent years \citep{ladicky2015data,kim2020lagrangian}, and is particularly interesting in this context to incorporate additional constraints, e.g., for temporal coherence \citep{tempoGAN}, or for physical plausibility \citep{tompson2016accelerating,kim2019}.

To summarize, the central contributions of our work are:
(1) A novel frequency aware wavelet-based loss formulation,
(2) The processing of non-periodic signals with respect to their frequency,
(3) The identification of details independent of position and scale,
(4) A time consistent spatio-temporal up-sampling of complex physical surfaces.
We 
tested the capabilities of our approach with three sets of spatio-temporal
data from physical simulations.
The experiments show an improvement in synthetic test cases, but the performance was hardly increased in more complex cases with our setup.

\paragraph{Related Work}

The idea of exploiting the frequency spectrum of data in Deep Learning is not new. 
For example, \citet{mildenhall2020nerf} and \citet{tancik2020fftfeats} use a Fourier transform as a feature mapping to increase the performance of MLPs. They show that by providing high-frequency features, the spectral bias can be reduced, allowing better reconstruction of high-frequency details.
The Fourier transform was also used in the generative domain to reconstruct details from low-information data \citep{ichimura2018fftae}, or to reduce noise in perceptional data \citep{yoo2018fftdenoise, yadav2021fftloss}. 
Wavelet transformation also found application in recent publications.
For example, wavelets have been used to augment the provided features to neural networks \citep{gao2016wltcnn, liu2019wltcnn} as well to variational autoencoders \citep{ichimura2018fftae} and to generative adversarial networks \citep{liu2018wltgan, wang2020wvlt, gal2021swagan}. This led to improvements in a variety of generative tasks, such as denoising \citep{lin2020wltdenoise, zhang2021wggan}, style transfer \citep{yoo2019wltstyle} or for super-resolution tasks \citep{zhang2019wltsr}.
In \citet{gal2021swagan} wavelet-based features were provided to the generator and discriminator architecture, which forces a frequency-aware latent representation.
Other methods attempted to generate wavelet decompositions directly \citep{huang2019wvltdgan, wang2020wvlt, zhang2019wltsr}.

Most of the generative methods mentioned above focus on perceptual data, such as photos or videos. However, we focus on structured, noise-free data from physical simulations, whose properties can be optimally exploited by our method.
Deep learning methods in conjunction with physical models were 
employed in variety of contexts, ranging from learning models for physical intuition
\citep{battagliaInteractionNetworksLearning2016,sanchez2018graph}, over 
robotic control \citep{schenckSPNetsDifferentiableFluid2018,hu2019difftaichi} to
engineering applications \citep{ling2016reynolds,morton2018deep}.
In the following, we focus on fluid-like materials with continuous descriptions,
which encompass a wide range of behavior and pose challenging tasks for learning methods
\citep{mrowcaFlexibleNeuralRepresentation2018,liLearningParticleDynamics2019}.
For fluid flows in particular, a variety of learning methods were proposed 
\citep{tompson2016accelerating,rtliquids2017,um2018mlflip}.
A common approach to reduce the high computational cost of a simulation is to employ super-resolution techniques \citep{dong2016image,chu2017cnnpatch,bai2019dynamic}.
In this context, our work targets the up-sampling for physics-based animations, similar to the approach proposed by \citet{tempoGAN}.
However, in contrast to this work, we target phenomena with clear interfaces,
which motivates the frequency-based viewpoint of our work.

For sharp interfaces, Lagrangian models are a very popular discretization of continuum mechanical systems. 
E.g., smoothed particle hydrodynamics (SPH) \citep{gingold1977smoothed,KBST19} is a widely-used particle-based simulation method. 
While points and particles are likewise frequently used representations for 
physical deep learning  \citep{liLearningParticleDynamics2019,ummenhofer2019lagrangian,sanchez2020learning},
Eulerian, i.e., grid-based representations offer advantages in terms of efficient and robust kernel evaluations.
A further advantage that results from the observation in wave space is the ability to process complex time phenomena.
The time dimension was also taken into account in natural imaging works,
e.g., by Saito et al. in the form of a temporal generator \citep{saito2017temporal},
or via a stochastic sequence generator \citep{yu2017seqgan}.
Other works have included direct $L_2$ loss terms as temporal regularizers \citep{bhattacharjee2017temporal,Chen2017ICCV},
which, however, typically strongly restricts the changes over time.
Similar to flow advection, video networks also often use warping information 
to align data over time \citep{liu2017VideoSR,de2017deep}.
We will demonstrate that recurrent architectures similar to those used for
video super-resolution \citep{sajjadi2018FRVSR} are likewise very amenable for physical problems
over time.

\section{Method}

The input for our method is a coarsely approximated source simulation, with the learning objective 
to infer the surface of a target simulation over space and time. This target is typically computed via 
a potentially very costly, finely resolved simulation run for the same physical setup.
When it comes to the possibilities of simulation representations, there is a great variance. In our case we have chosen an implicit representation of the data, by a signed-distance field (SDF) denoted by $g: \mathbb{R}^3 \rightarrow \mathbb{R}$. An SDF returns, for a given point, the signed distance to the surface, with negative being inside the medium.
Such a function is realized in practice by a grid $X \in \mathbb{R}^{M_x\times M_y\times M_z}$, storing the pre-computed signed distance values, where $M_*, * \in \{x, y, z\}$ specifies the size of the grid in the respective dimension $x$, $y$ or $z$.
We have chosen this representation because most neural network layers are designed for array-like representations, and the loss functions on grid-based data are very efficient to evaluate. Additionally, an implicit representation via a grid can leverage tools from the field of level-set processing \citep{adalsteinsson1999fast}, and facilitates the frequency viewpoint via a wavelet transformation.
Additional values, like the velocity, are also mapped on a grid $V \in \mathbb{R}^{M_x\times M_y\times M_z\times 3}$.
Our goal is to let a generative network $\mathcal{G} : \mathbb{R}^{M_x\times M_y\times M_z\times 4} \rightarrow \mathbb{R}^{N_x\times N_y\times N_z}$ infer a grid $\tilde{Y}$ which approximates a desired high-resolution simulation $Y \in \mathbb{R}^{N_x\times N_y\times N_z}$ with $N_* = kM_*, N_* \in N_{\{x,y,z\}}$ and up-sample factor $k \in \mathbb{N}$, i.e. $\mathcal{G}(X) = \tilde{Y} \approx Y$. 
As our method only requires position and velocity data from a simulation, it is largely agnostic to the 
type of solver or physical model for generating the source and target particle data.

\subsection{Neural Network Formulation}

\begin{wrapfigure}{t}{0.56\linewidth} 
	\centering
	\vspace{-5mm}
    	\includegraphics[width=\linewidth]{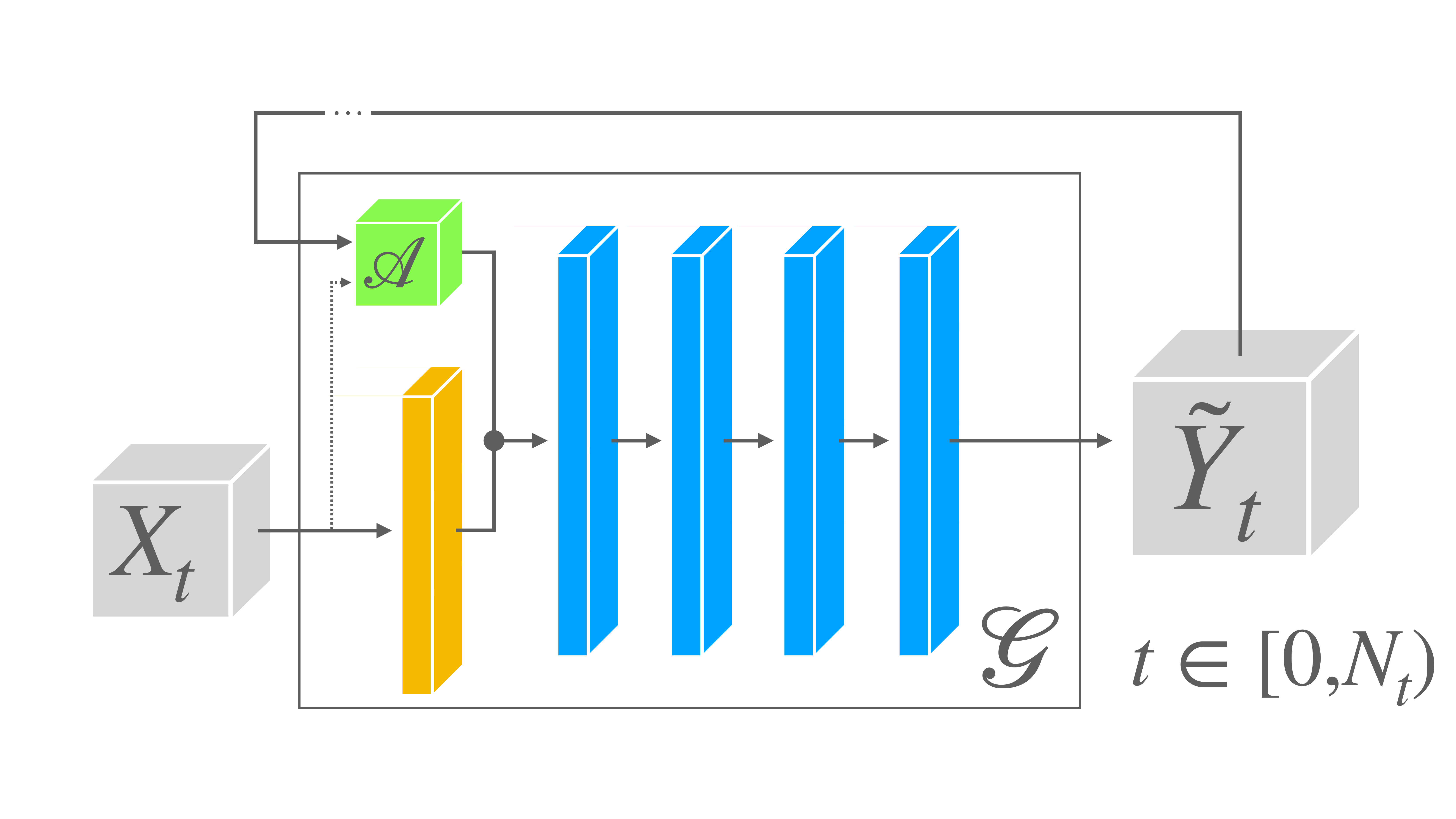}
		    					\vspace{-3mm}
	\caption{ 
	   The generator 
	   	   takes low-resolution data $X_t$ as input, together with a high-resolution version, from the previous time step $\tilde{Y}_{t-1}$. The low-resolution data is tri-linearly up-scaled  (orange layer), while the data with the previous state is advected with the velocity of the input (green layer). The processed inputs are concatenated and processed via four residual blocks (in blue).
	   A high-resolution frame $\tilde{Y}_t$ is generated as output.
	   	}\label{fig:gen}
	\vspace{-3mm}
\end{wrapfigure}

Our method is based on a generative, neural network with a 3D fully-convolutional ResNet architecture \citep{he2016resnet} that produces an output field at a single instance in time.
The low-resolution input data is first up-sampled with a tri-linear up-sampling and then processed with several convolutional layers, as shown in \Figref{fig:gen}. We use leaky ReLU as activation function after each layer, except for the last layer, where we use a no activation.
In our case, the input data consists of the implicitly represented geometry data $X_t$, the velocity $V_t$ of the simulation as well as the results of a previous pass $\tilde{Y}_t$.
The previously generated data is advected with the low-resolution velocity before further processing.
Through this feedback loop we train our network recurrently by iterating over a sequence of $T=10$ frames. 
This yields stability over longer periods of time and gives better insights about temporal behaviour.
Furthermore, the recurrent training is important to enable persistent behavior over time, such as the progression of fine surface waves.
Unlike the process for generating the input data, the network training cannot resort to a physical simulation with full resolution, and hence cannot uniquely determine the evolution of future states. Therefore, its main learning objective is to capture the dynamics of the target simulations beyond that basic motion computed with an advection step.
For initialization of the undefined first frame $\tilde{Y}_{-1}$ we use a tri-linear up-sampled version of the input.

\subsection{Loss Formulation}
\begin{figure}[t]
	\centering
	\includegraphics[width=\textwidth, trim= 0 100 0 100 clip]{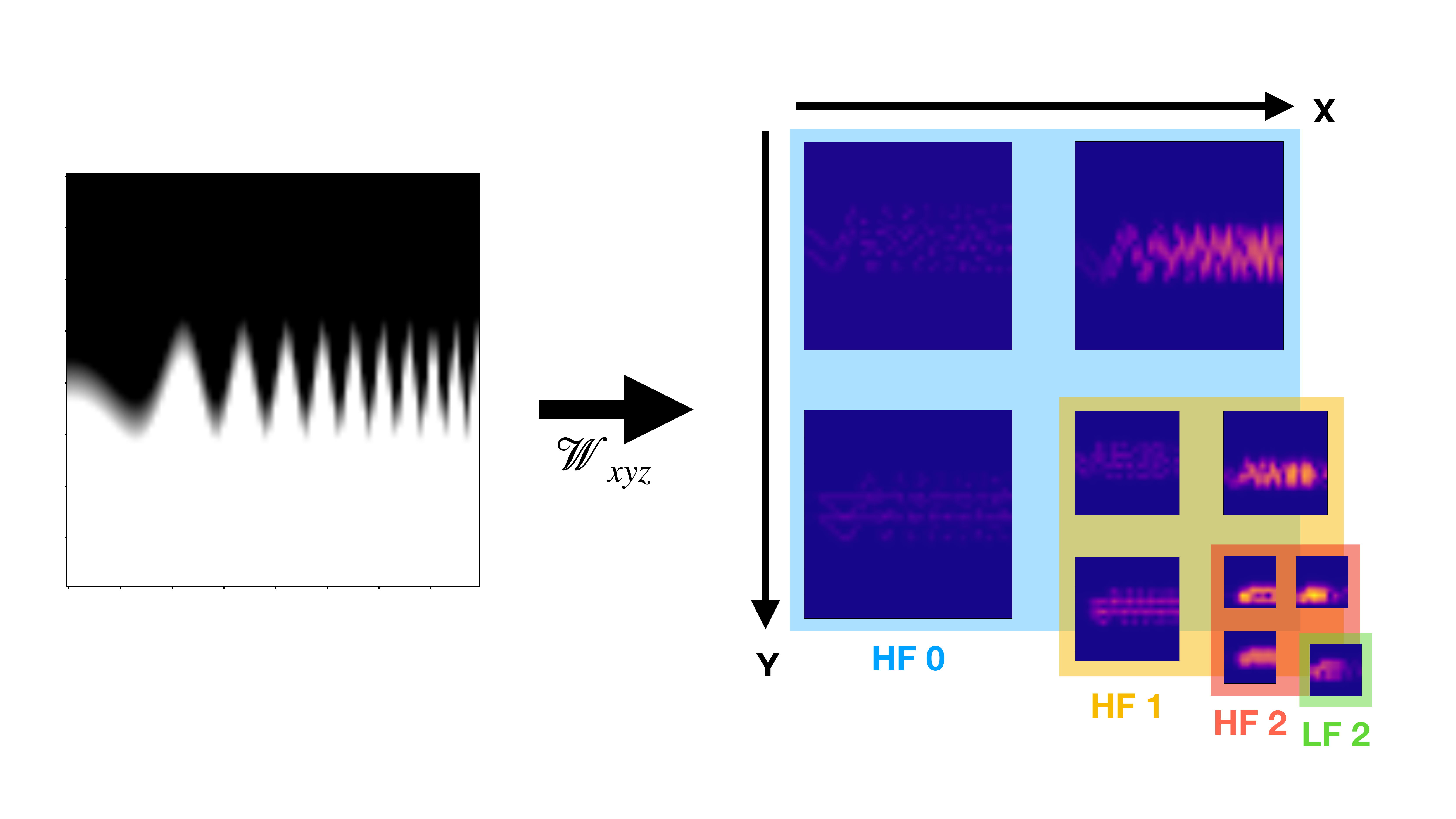}    
	\caption{ 
	    The process of discrete wavelet transform, shown with a chirp signal. For each level, the given data is processed with a low-pass and a high-pass filter. The filters are applied along the $x$-axis (to the right), the $y$-axis (downwards) and the $x$-$y$-axis (diagonally). The result of the low-pass filter is used to generate the components in the next lower level. This gives a hierarchical gradation. The results of the high-pass filter are the wavelet coefficients used for the loss (HF0 - HF2). The chosen representation clearly shows that lower levels represent low frequencies (HF2), while higher levels filter high frequencies (HF0).
	}
	\label{fig:wlt}
\end{figure}

In the following, we keep the generator network constant 
to evaluate the differences for a choice of the loss function to measure the differences between generated and ground-truth data.
The most basic loss function is a simple mean absolute error (MAE):
\begin{equation}
    \mathcal{L}_s = |Y-\tilde{Y}| .
\end{equation}
This choice has the central disadvantage that it is ill-suited to measure the similarity or differences of solutions in a multi-modal settings. Multi-model here means that multiple solutions for a given input exist.
A method that trains with an MAE loss in this setting will learn the expected value of the output distribution, i.e, the average of the different solutions. However, the average is typically not a part of the solution set. Thus, the MAE loss often does not correspond to the correct distance in solution space, based on significant factors corresponding to the distribution of the solutions. Our super-sampling setup is such a problem: Due to the low resolution input, the high resolution details cannot be determined uniquely, resulting in a variety of possible solutions when up-sampling. Via physical properties of the material and its temporal sequence, some solutions can be eliminated, but nonetheless the space of solutions typically remains infinitely large. If a MAE loss is used, all such samples from the training data set are simply averaged to obtain a mean value, so that the result does no longer reflect the level of detail of the ground-truth data.
The MAE loss nevertheless gives a rough direction, and provides a stable learning target. Hence, we still use it as a component in the final loss formulation, 
in combination with a wavelet loss. 

Considering the problem in frequency space, the sought after detail consists of high-frequency features that cannot be represented by the low-resolution simulation. 
The mean-squared distance can be seen in our problem similar to a low-pass filter. To increase the level of detail again, we have to try to include higher frequencies in the loss calculation. A simple way would be a Fourier transformation of the data and a L2 distance in Fourier space. A Fourier transformation expects periodic data, which can be circumvented with a short time Fourier transformation. For this one transforms always only a partial area of the space, smoothed with a window function. Thus one receives a space-frequency representation, whereby the resolution in the space dimension depends on the size of the subrange. 

A next step is the use of a wavelet transformation. Unlike the Fourier transform, the wavelet transform also explicitly takes into account the spatial dimensions. Thus, it gives a larger amount of variance in the spatial dimensions, which in turn means that spatial details can be isolated and filtered.

For the wavelet loss we transform data into a space-frequency representation using a multi-scale discrete wavelet transform $\mathcal{W}(x) = [w_i, i \in [0, I)]$:
\begin{equation}
\begin{split}
    HF_{i} &= \langle \psi_{hi},  LF_{i-1} \rangle \\
    LF_i        &= \langle \psi_{li}, LF_{i-1} \rangle \\
    LF_0        &= x,
\end{split}
\end{equation}
where $\psi_{hi}$ and $\psi_{li}$ is a high-pass and low-pass filter of a wavelet filter bank. In our case, we use Daubechies-2 wavelets \citep{daubechies1992ten}.
The first pass starts processing the input data $x$. 
The high pass filter generates the wavelet component of the respective level, the low pass filter on the other side generates a scaled down version of the data, which is then used in the next pass. The process is repeated to generate the wavelet components in a hierarchical fashion. Each level represents a different frequency, resulting in a list of differently scaled wavelet components. \Figref{fig:wlt} shows an example of the discrete wavelet transformation of a 2D \textit{cirp} signal. A signal whose frequency increases progressively over time. The number of levels $I$ is given by $\lfloor log_2 d \rfloor$, where $d$ is the smallest input dimension.
In addition we scale the values using the logarithm:
\begin{equation}
    w_{i} = log_2(|HF_i| + \epsilon).
\end{equation}
This way we compensate for the fact that high-frequency components have less energy, making low-frequency components very dominant. The added $\epsilon$ is used to avoid very high values and to keep the scaling stable. 
For multi-dimensional grids, the wavelet transform is applied multiple times, with each pass applying the filter to the respective dimension. We denote the wavelet transform by $\mathcal{W}_d$ where $d$ denote the dimensions from which the wavelet transform is formed.
Using the wavelet transform, the L1 distance is calculated from the scaled and transformed data and used as the loss:
\begin{equation}
    L_{ws} = \sum\limits_{t=0}^{T} |\mathcal{W}_{xyz}(Y_t) - \mathcal{W}_{xyz}(\tilde{Y}_t)|
\end{equation}

While we have primarily focused on spatial content so far, i.e., the surface of the material, 
the temporal behavior likewise plays a crucial role, and poses similar difficulties in our multi-modal setting.
On the one hand, the generation of details can quickly lead to temporally incoherent results, which is characterized by unappealing flickering. On the other hand, our network also should be able to match and recreate spatial solutions over time that 
reflect the physical behavior. 
Similar to the spatial loss, we can also resort to a wavelet transformation. One possible solution would be, to increase the dimensionality of the wavelet transform by one, giving a space-time-frequency representation. However, we chose to consider space and time separately. The separation of spatial and temporal frequencies gives us more flexibility by adjusting the weighting. Using the wavelet transform, we generate a time-frequency representation, similar to the spatial loss:
\begin{equation}
    L_{wt} = |\mathcal{W}_{t}(Y) - \mathcal{W}_{t}(\tilde{Y})|.
\end{equation}

The distances in wavelet space are included in the loss formulation of the generator which gives the final loss function: 
\begin{equation}
    \label{eq:total}
    \mathcal{L}_G = \mathcal{L}_s + \alpha \mathcal{L}_{ws} + \beta \mathcal{L}_{wt} ,
\end{equation}
where $\alpha$ and $\beta$ indicate the weighting of the individual loss terms.

To indicate the focus on surface structures, we refer to the final version of our generative network as \textit{surfNet}. For a more detailed description of the training and the network architecture we refer to the appendix \ref{sec:impl_details} and \ref{sec:train_details}.

\begin{figure*}[t]
	\centering
    \begin{subfigure}[c]{0.42\textwidth}
    \begin{overpic}[width=0.49\linewidth, trim= 0 20 0 50]{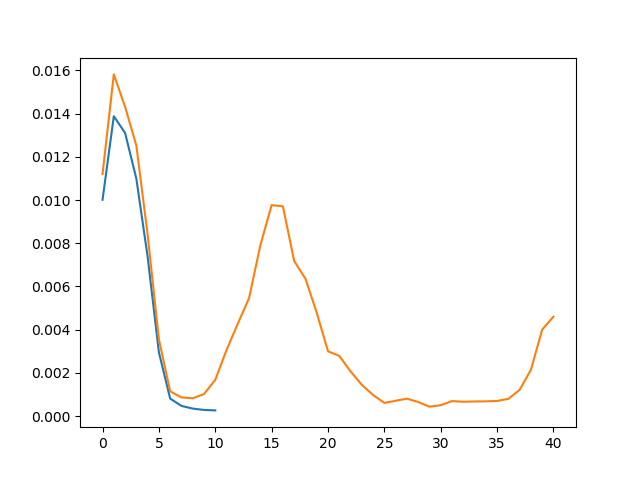}    
    	\put(-8,15){\rotatebox{90}{\tiny \color{black}{amplitude}}}
    	\put(30,-10){\tiny \color{black}{frequency [Hz]}}\end{overpic}~
    \begin{overpic}[width=0.49\linewidth, trim= 0 20 0 50]{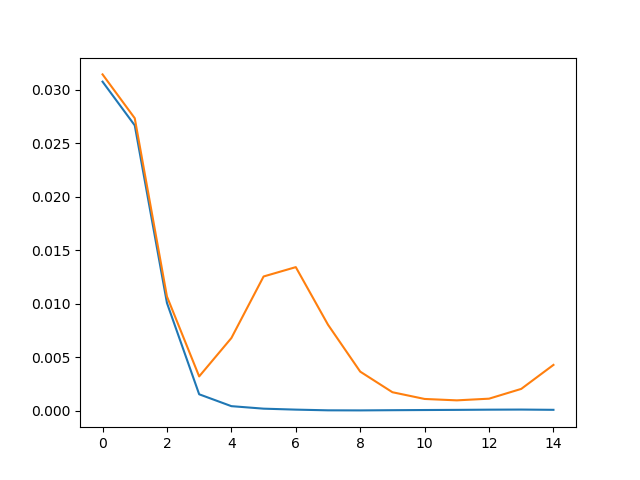}    
    	\put(-8,15){\rotatebox{90}{\tiny \color{black}{amplitude}}}
    	\put(30,-10){\tiny \color{black}{frequency [Hz]}}\end{overpic}~
    \vspace{3mm}
	\subcaption{Frequency histograms (spatial and temp.).}
	\end{subfigure}
    \begin{subfigure}[c]{0.57\textwidth}
	\includegraphics[width=0.47\linewidth]{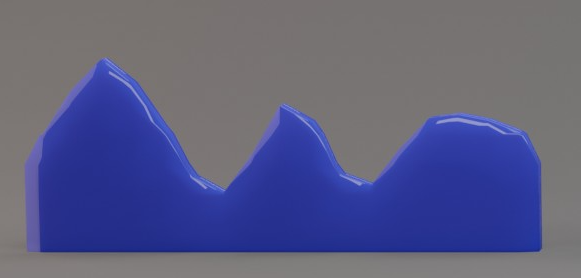} 
	\includegraphics[width=0.51\linewidth]{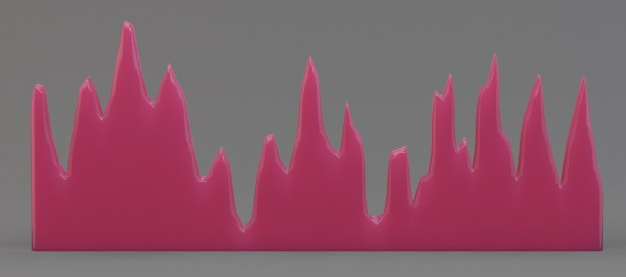} 
	\subcaption{Synthetic data sample.}
	\end{subfigure}
	\caption{ 
	    The averaged frequency spectrum of our data set (a) for the surface, spatial (left), and for the temporal behavior (right). The spectrum of the low frequency input data is shown in blue, while in orange the spectrum of the ground-truth data is shown. In the spatial spectrum, it's clearly visible that the spectrum of the input data only covers a quarter of the frequencies, while a large peak of the ground-truth is visible for higher frequencies. This highlights what the generator needs to reproduce.
	    	    A similar shape can be seen in the time spectrum as the time response is strongly coupled to the surface frequency. Here, the spectrum of the input data (blue) continues as the time discretization is the same for both.
	    	    Figure (b) shows an example of a surface from our synthetic data set: 
	    the input wave in blue, and the corresponding ground-truth target in purple.
	}
\label{fig:2d_sample}
\end{figure*}

\section{Results}
For evaluation we consider three different data sets: a synthetic 2D case, a 2D simulation based on a mass-spring system and a 3D particle-based simulation. 
The synthetic data is fully controllable such that the frequency spectrum of the surface can be evaluated reliably.
The basis is a wavy surface, based on a sine wave with varying frequency (\Figref{fig:2d_sample}). This forms a wide range of analysis to isolate problems in the generative process and illustrate the aspects of the proposed method. 
We then evaluate the established methodology with data generated from physical simulations for a more complex scenario. We use elastic bodies based on a mass-spring system in 2D and we also consider a simulation with plastic material with data generated from a highly viscous SPH simulation \citep{WKBB18}.
For more information on the three data sets, refer to the appendix \ref{sec:synth_data}, \ref{sec:sim_data_ms} and \ref{sec:sim_data}.

\begin{figure*}[t]
	\centering
    				    \begin{subfigure}[c]{0.245\textwidth}
    \begin{overpic}[width=\textwidth, trim= 50 50 50 50, clip]{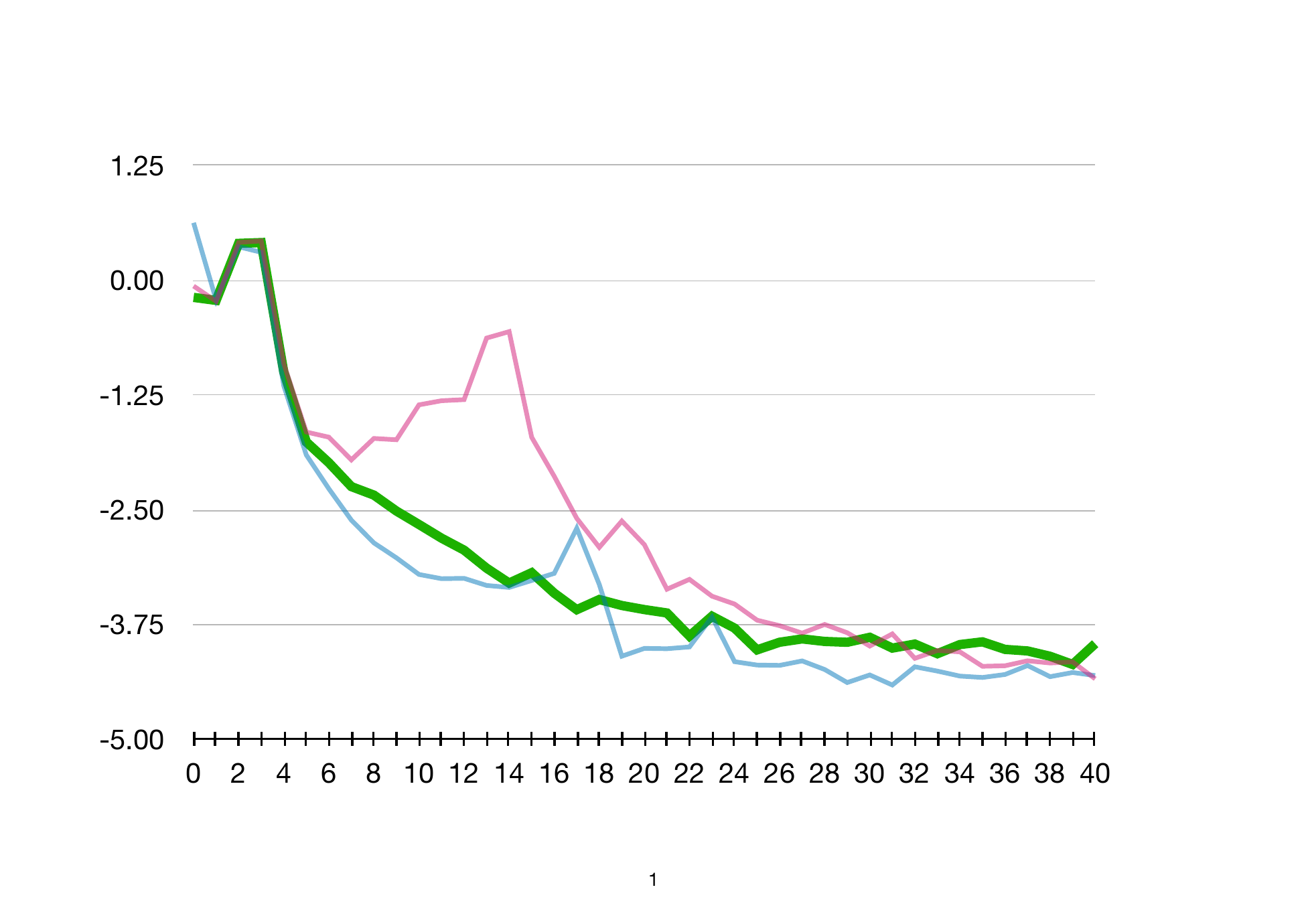}
    	\put(-8,20){\rotatebox{90}{\tiny \color{black}{log. amplitude}}}\end{overpic}~
    	
    \begin{overpic}[width=\textwidth, trim= 50 50 50 50, clip]{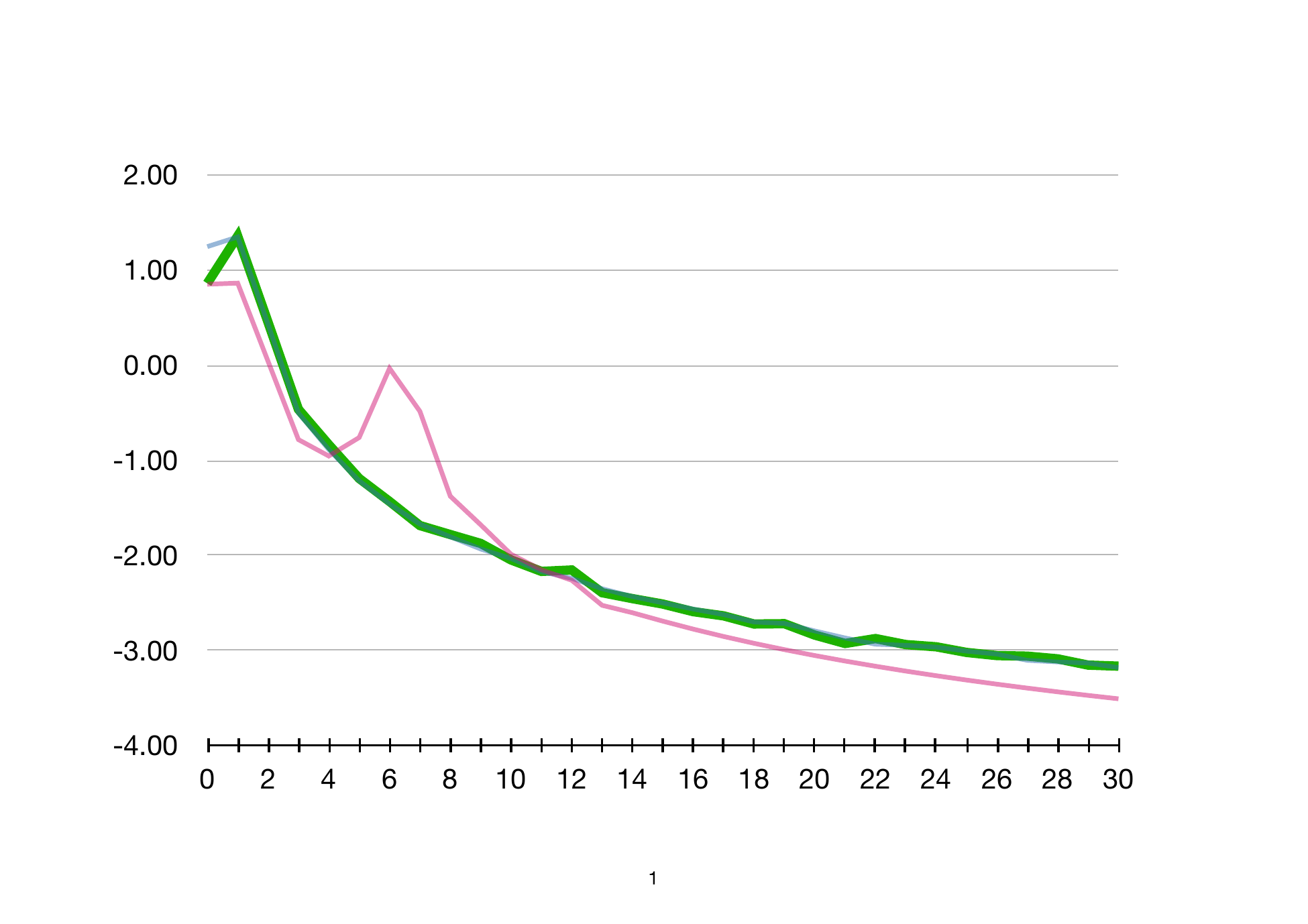}    
    	\put(-8,20){\rotatebox{90}{\tiny \color{black}{log. amplitude}}}
    	\put(30,-8){\tiny \color{black}{frequency [Hz]}}\end{overpic}~
    \vspace{3mm}
	\subcaption{MAE.}
	\end{subfigure}
    \begin{subfigure}[c]{0.245\textwidth}
	\includegraphics[width=\textwidth, trim= 50 50 50 50, clip]{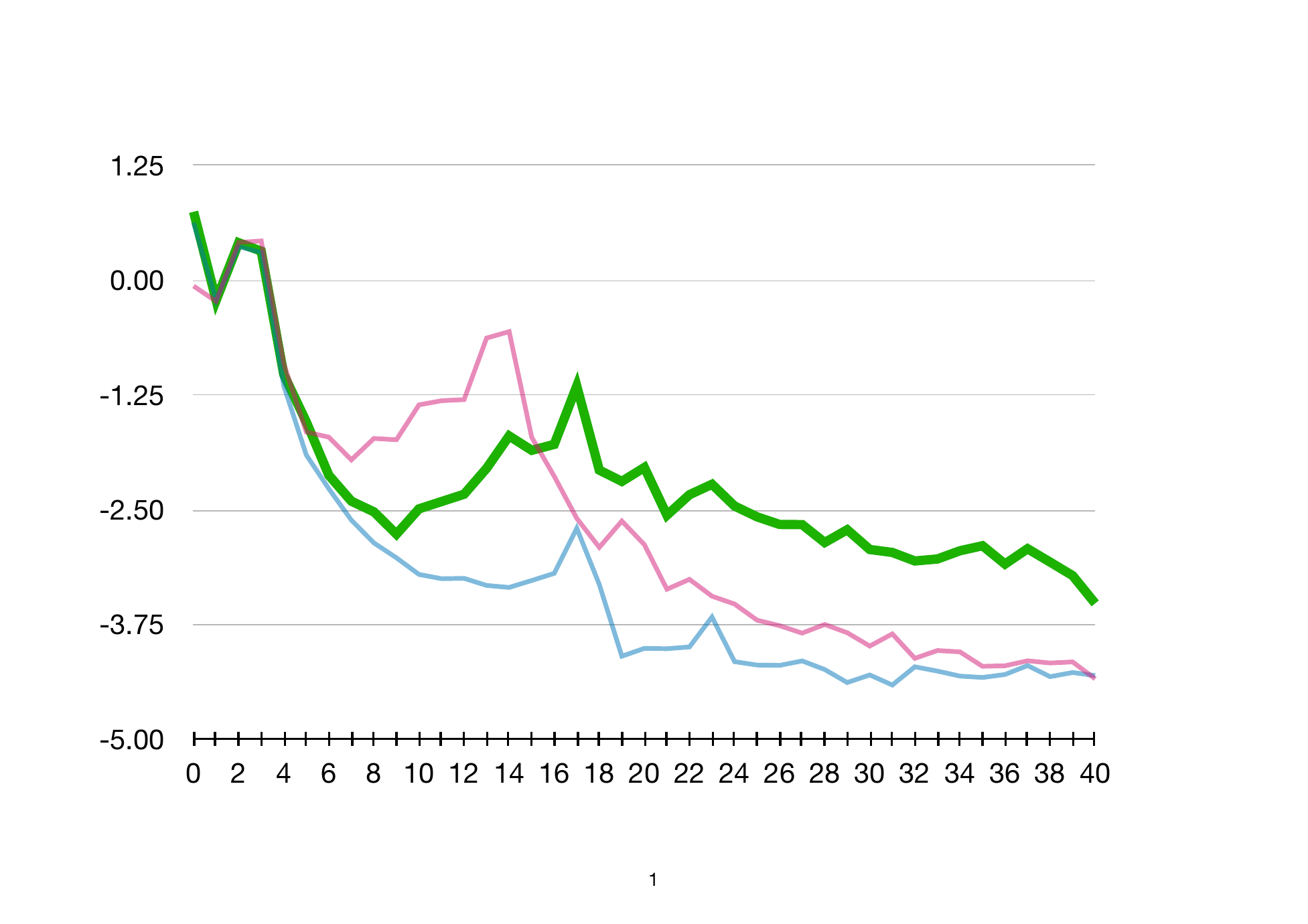} 
	
    \begin{overpic}[width=\textwidth, trim= 50 50 50 50, clip]{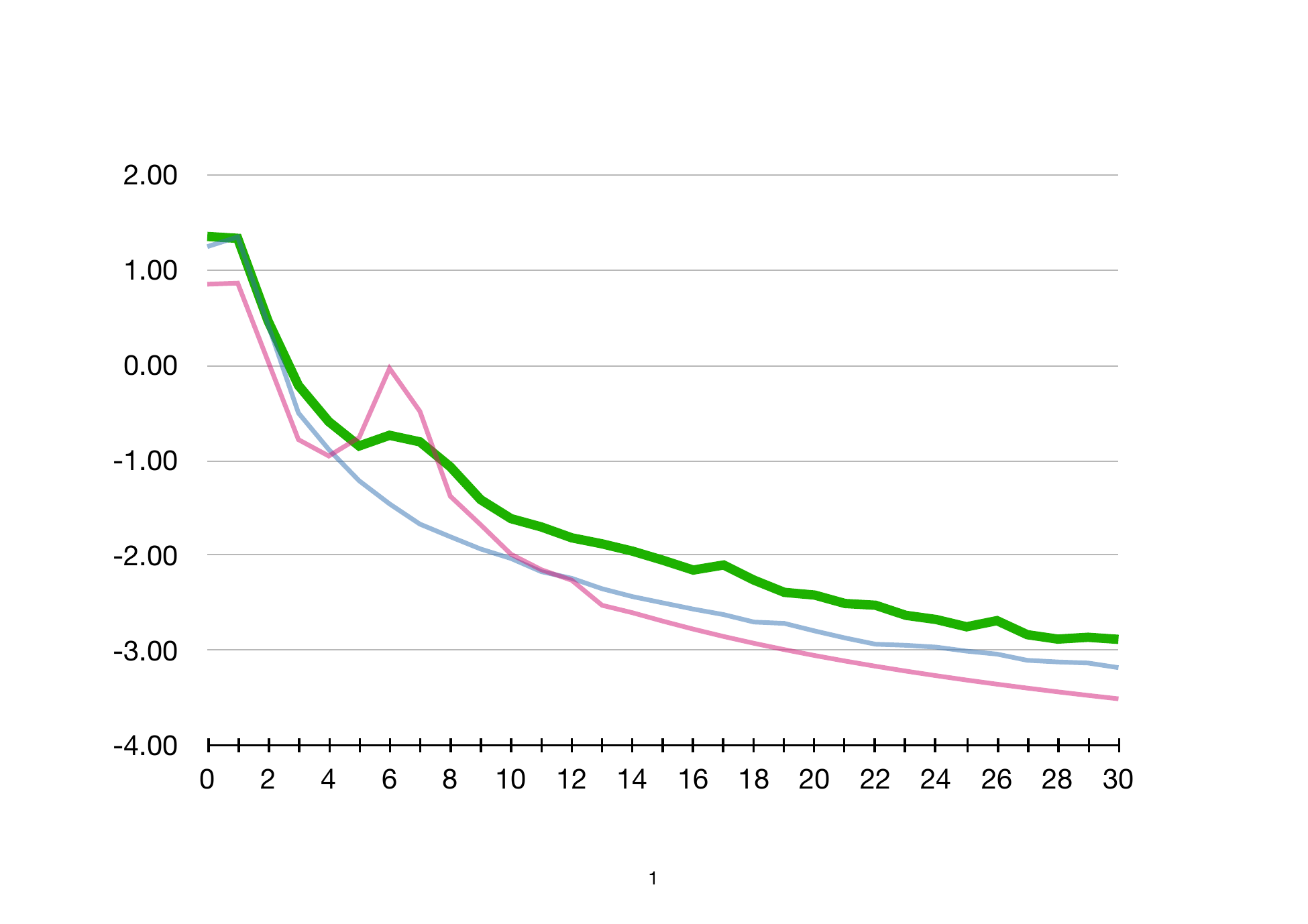}
    	\put(30,-8){\tiny \color{black}{frequency [Hz]}}\end{overpic}~
    \vspace{3mm}
	\subcaption{RFFT loss.}
	\end{subfigure}
    \begin{subfigure}[c]{0.245\textwidth}
	\includegraphics[width=\textwidth, trim= 50 50 50 50, clip]{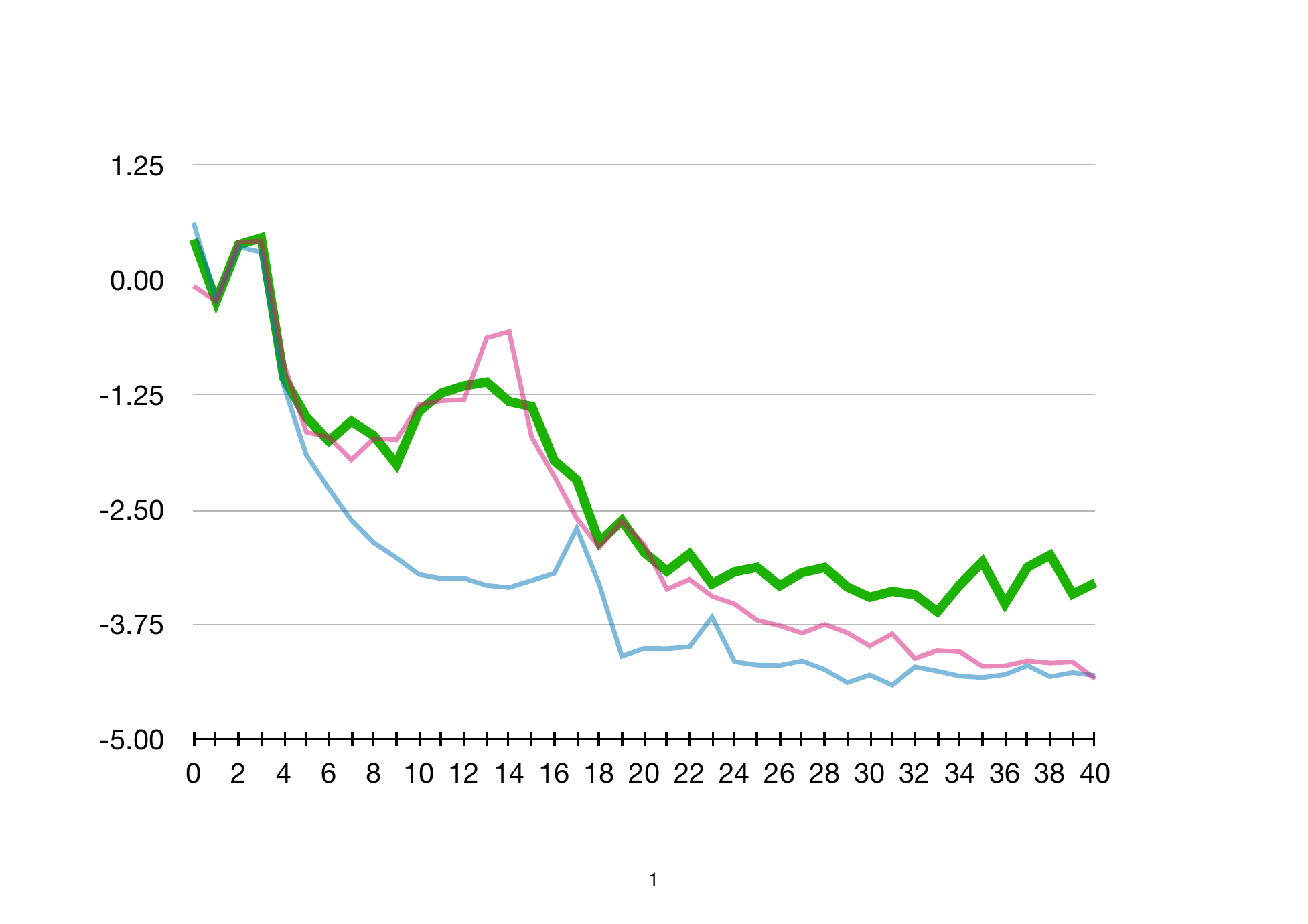}
	
    \begin{overpic}[width=\textwidth, trim= 50 50 50 50, clip]{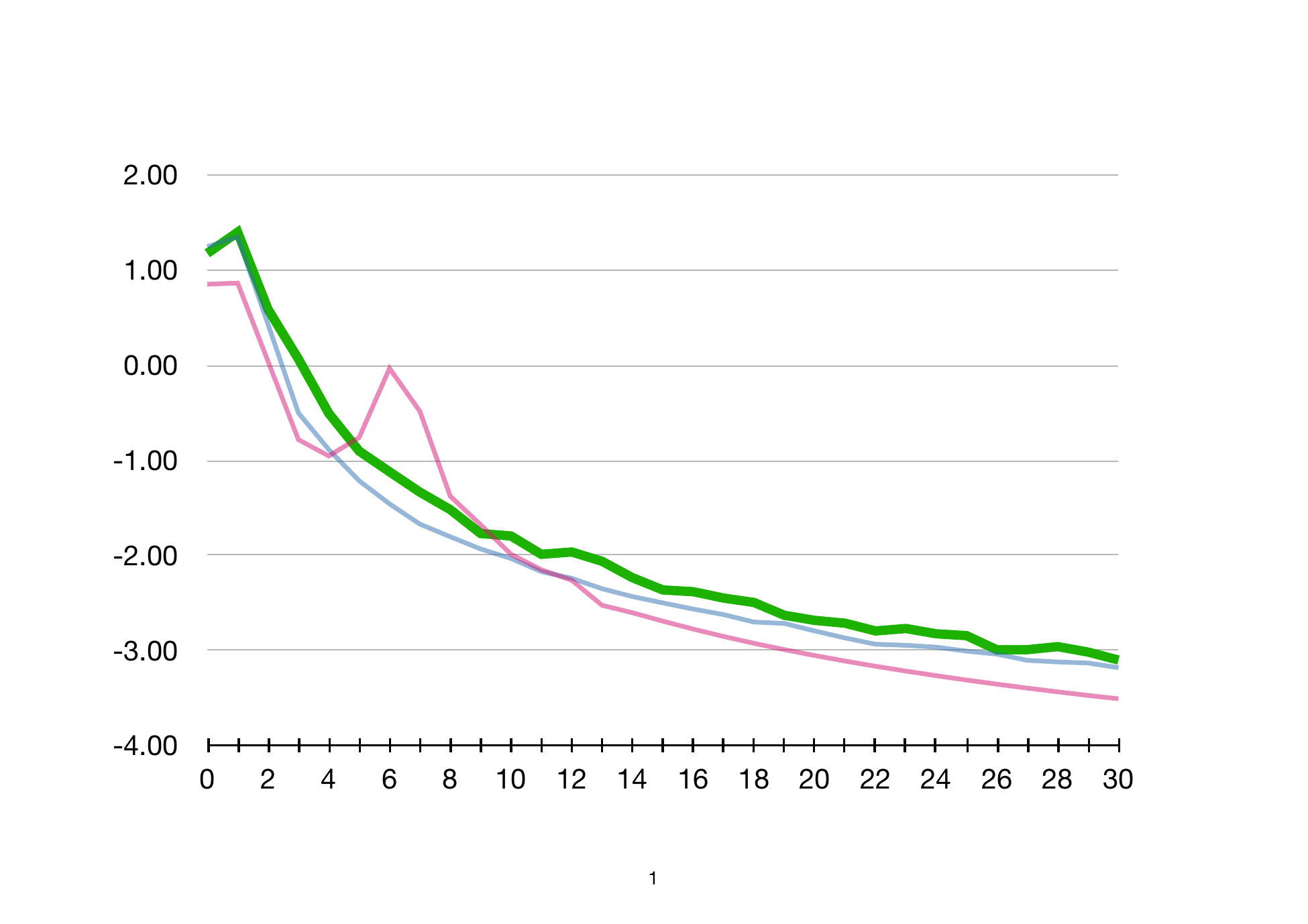}
    	\put(30,-8){\tiny \color{black}{frequency [Hz]}}\end{overpic}~ 
    \vspace{3mm}
	\subcaption{GAN.}
	\end{subfigure}
    \begin{subfigure}[c]{0.245\textwidth}
	\includegraphics[width=\textwidth, trim= 50 50 50 50, clip]{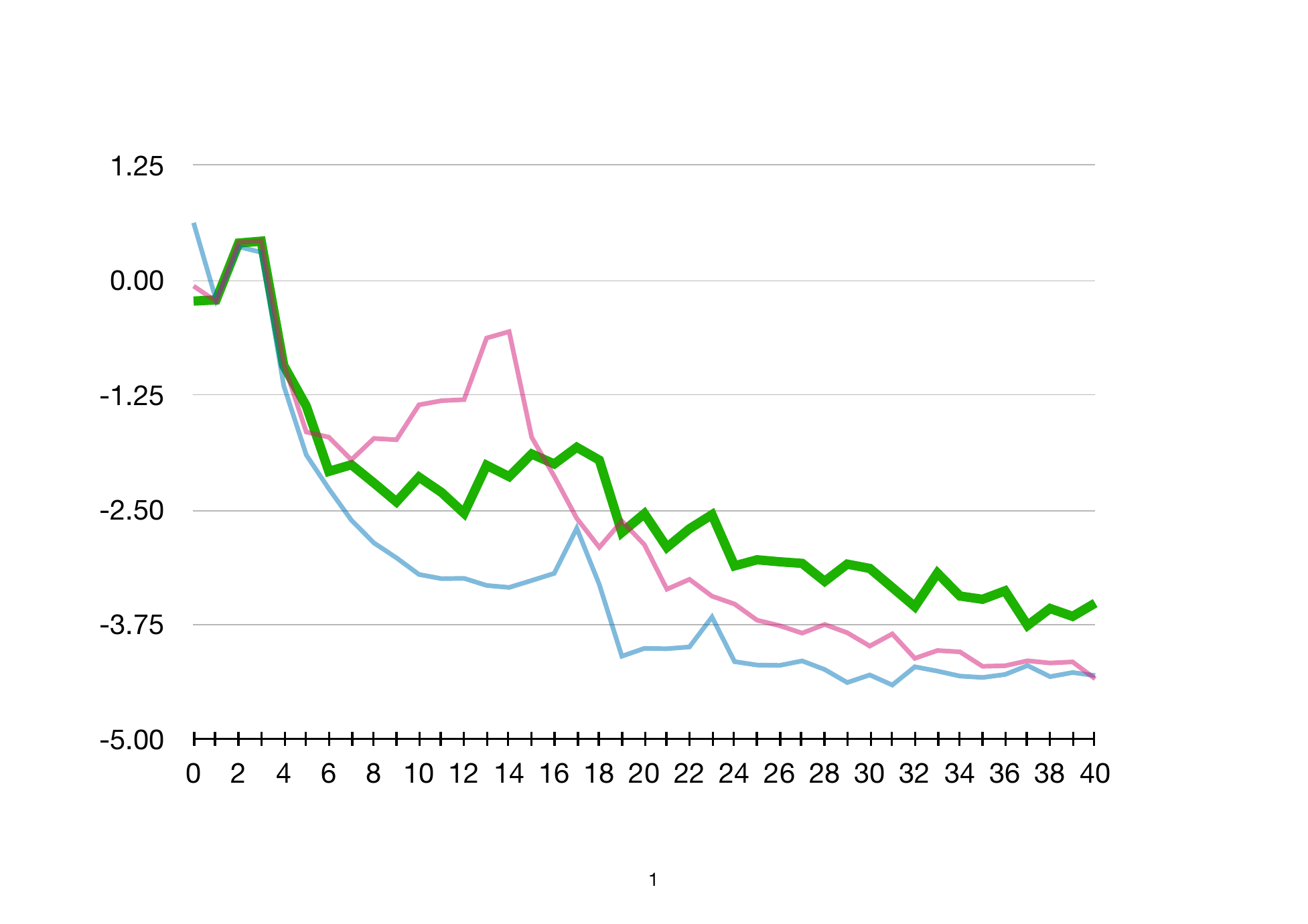}  
	
    \begin{overpic}[width=\textwidth, trim= 50 50 50 50, clip]{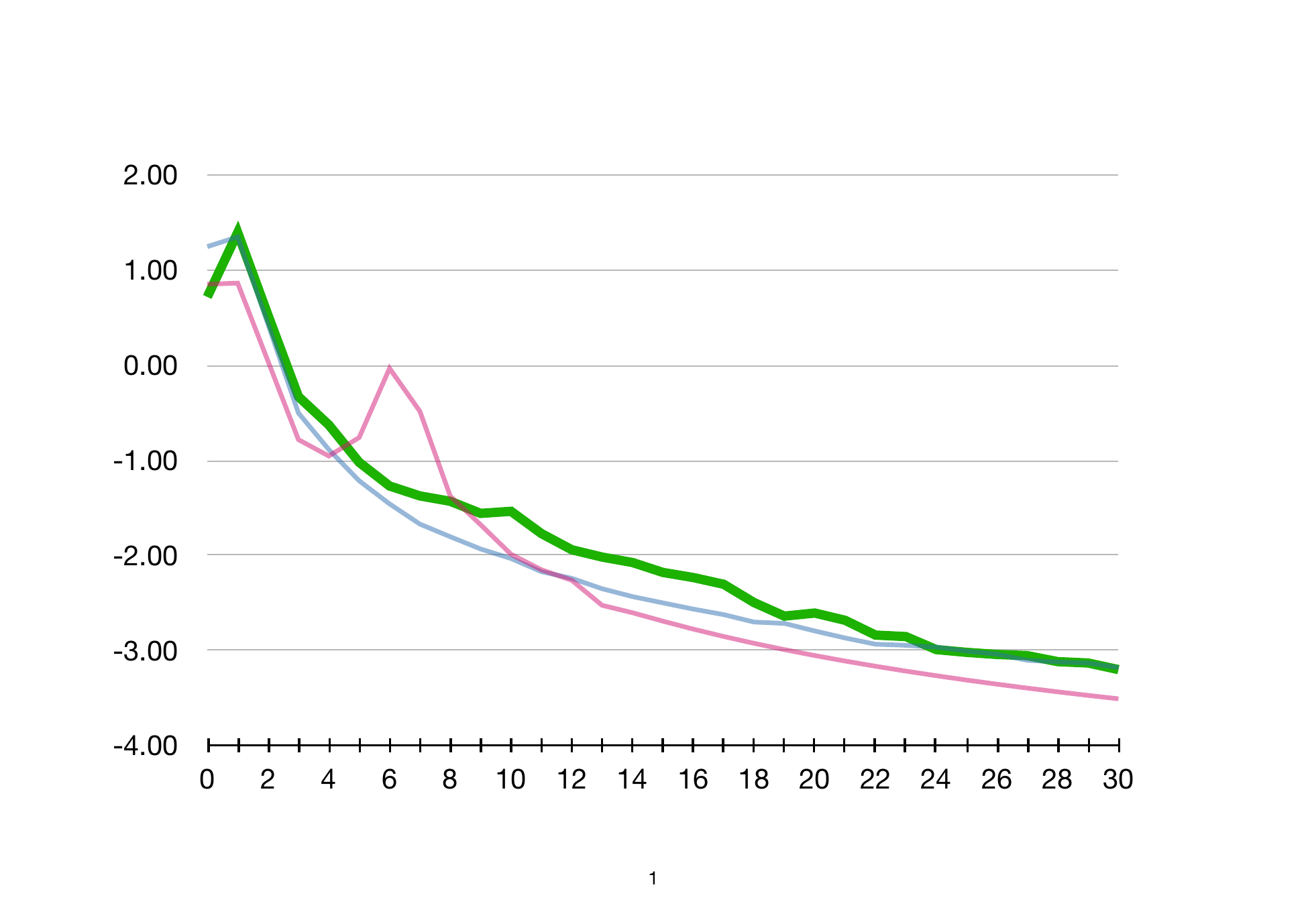}
    	\put(30,-8){\tiny \color{black}{frequency [Hz]}}\end{overpic}~
    \vspace{3mm}
	\subcaption{surfNet.}
	\end{subfigure}
	
	\caption{ 
	   Frequency spectrum comparison of different versions of our network for the synthetic 2D data set. The top row always shows the spatial frequency, while the bottom row shows the temporal frequency. A logarithmic scaling is used for the amplitude. The spectrum of the input is shown in blue, the ground-truth in pink, and the predictions in green. 
	   Version (a) was trained with a simple MAE loss, while (b) was extended with a RFFT loss; (c) was trained with an adversarial loss and (d) was trained with our wavelet loss.
	}
	\label{fig:freq_eval}
\end{figure*}
\begin{table}[t]
	\centering \footnotesize
\begin{tabular}{l|c|c|c|c}        & \ \ MAE \ \ & RFFT \ & GAN & \ surfNet \\
    \hline
    \hline
    \textbf{MAE}    
	&  0.039 & 0.074 & 0.052 & 0.047 \\
	\\	\\	\\    	\\	\\	\\    \hline
	\textbf{Spatial freq. MAE} 	& 0.990 & 1.110 & \textbf{0.819} & \textbf{0.884} \\
        \hline
	\\	\\	\\    	\textbf{Temp. freq. MAE} 	& 0.881 &  0.997 & 0.978 & \textbf{0.738} \\
\end{tabular}
\caption{
    Mean errors for 2D variants. The spatial and temporal frequency MAE values represent the averaged difference of the generated and the target residuals in frequency space.
}
\vspace{-3mm}
\label{tab:freq_eval}
\end{table}

\begin{figure*}[t]
	\centering
    \begin{subfigure}[c]{0.082\textwidth}
	\begin{overpic}[width=\linewidth, trim= 640 200 640 400, clip]{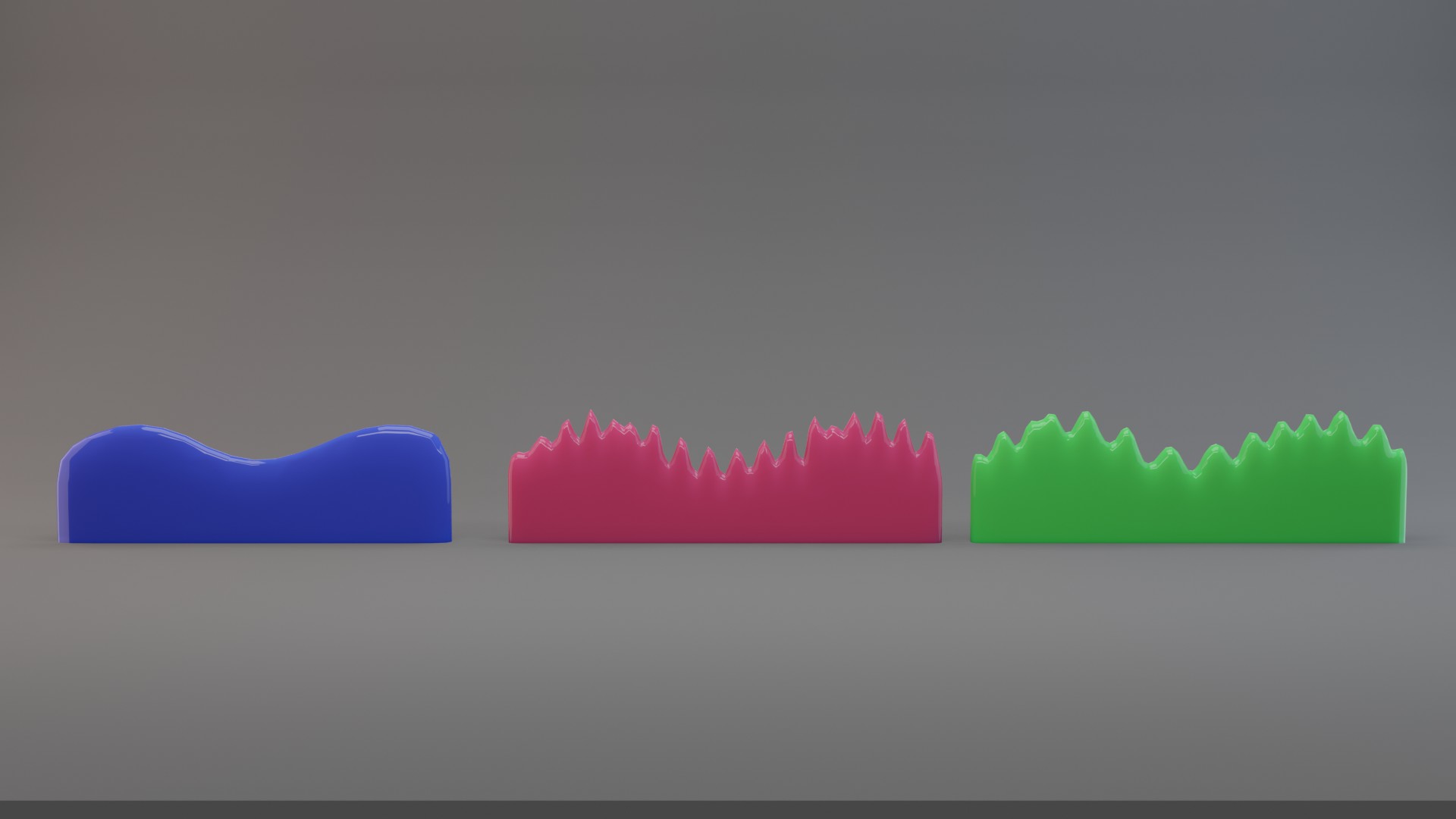} 
	    \put(-50,50){\small \color{black}{$a)$}}\end{overpic}
	    
    \vspace{0.5mm}
	\begin{overpic}[width=\linewidth, trim= 1260 200 20 400, clip]{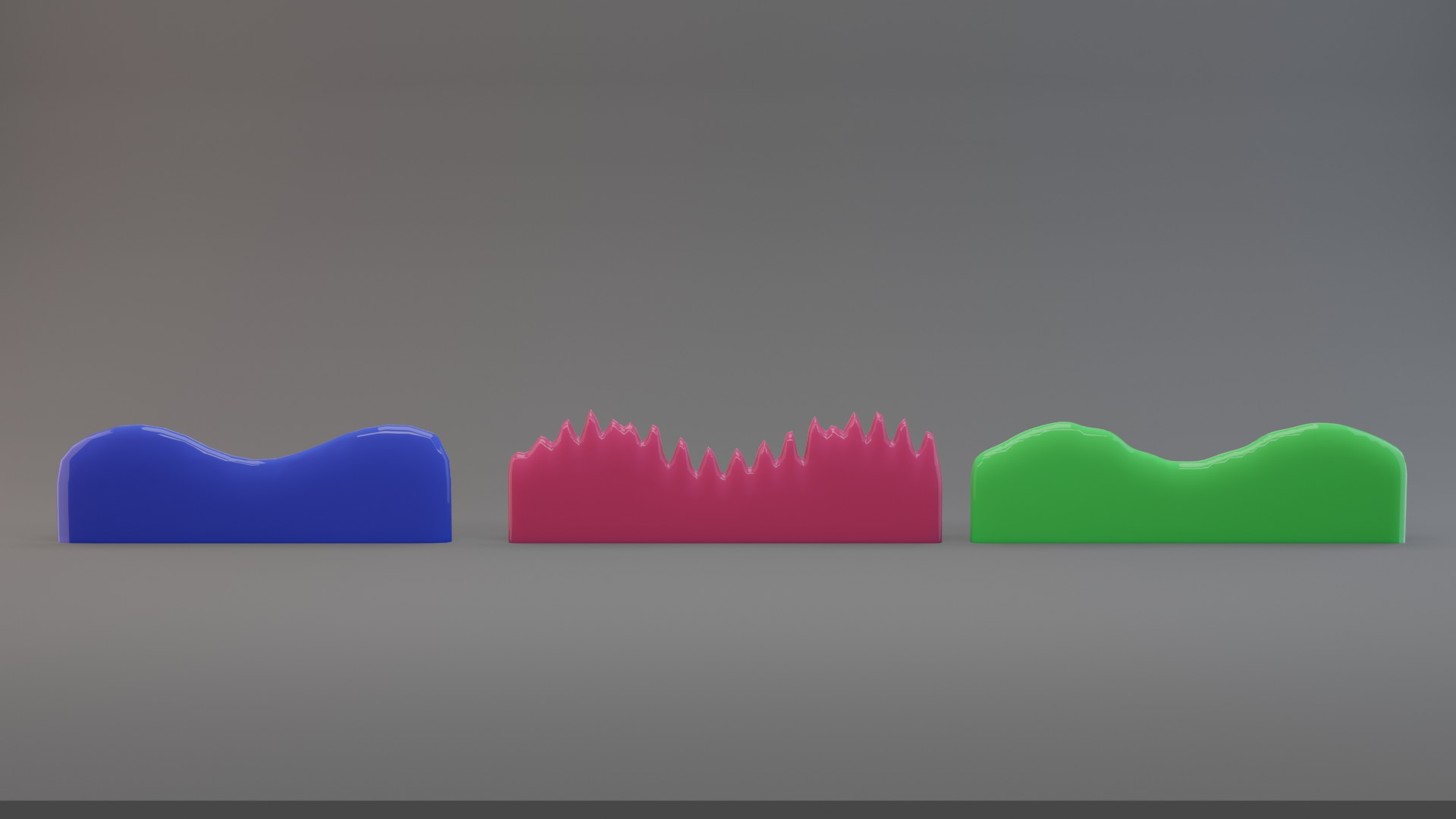} 
	    \put(-50,50){\small \color{black}{$b)$}}\end{overpic}
	    
    \vspace{0.5mm}
	\begin{overpic}[width=\linewidth, trim= 1260 200 20 400, clip]{data/2d/wlt/110.jpg} 
	    \put(-50,50){\small \color{black}{$c)$}}\end{overpic}~
				\subcaption*{}
	\end{subfigure}
	\hspace{1mm}
	\vline
	\hspace{1mm}
    \begin{subfigure}[c]{0.7\textwidth}
	\includegraphics[width=0.1\linewidth, trim= 103 38 87 60, clip]{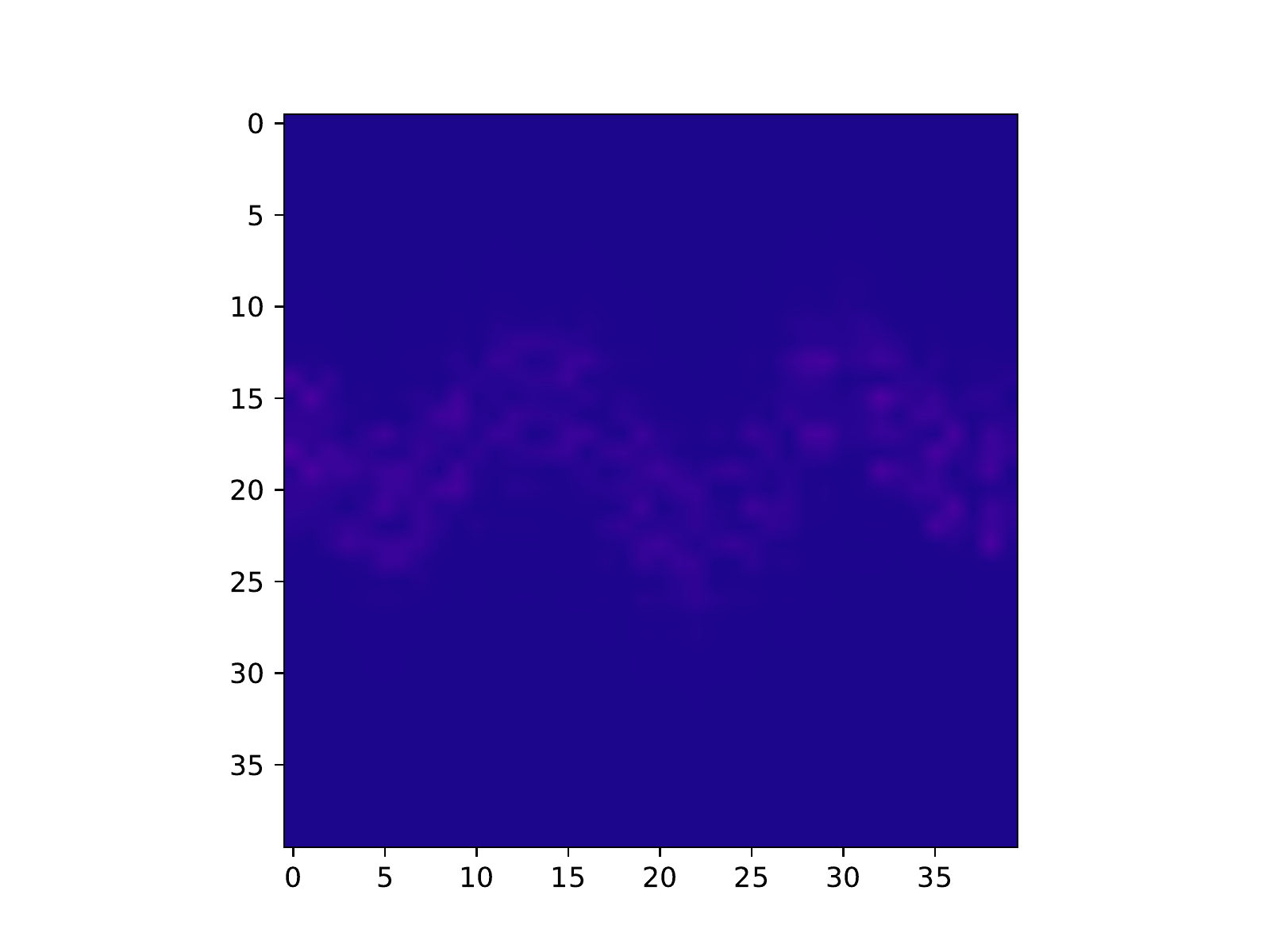}
	\includegraphics[width=0.1\linewidth, trim= 103 38 87 60, clip]{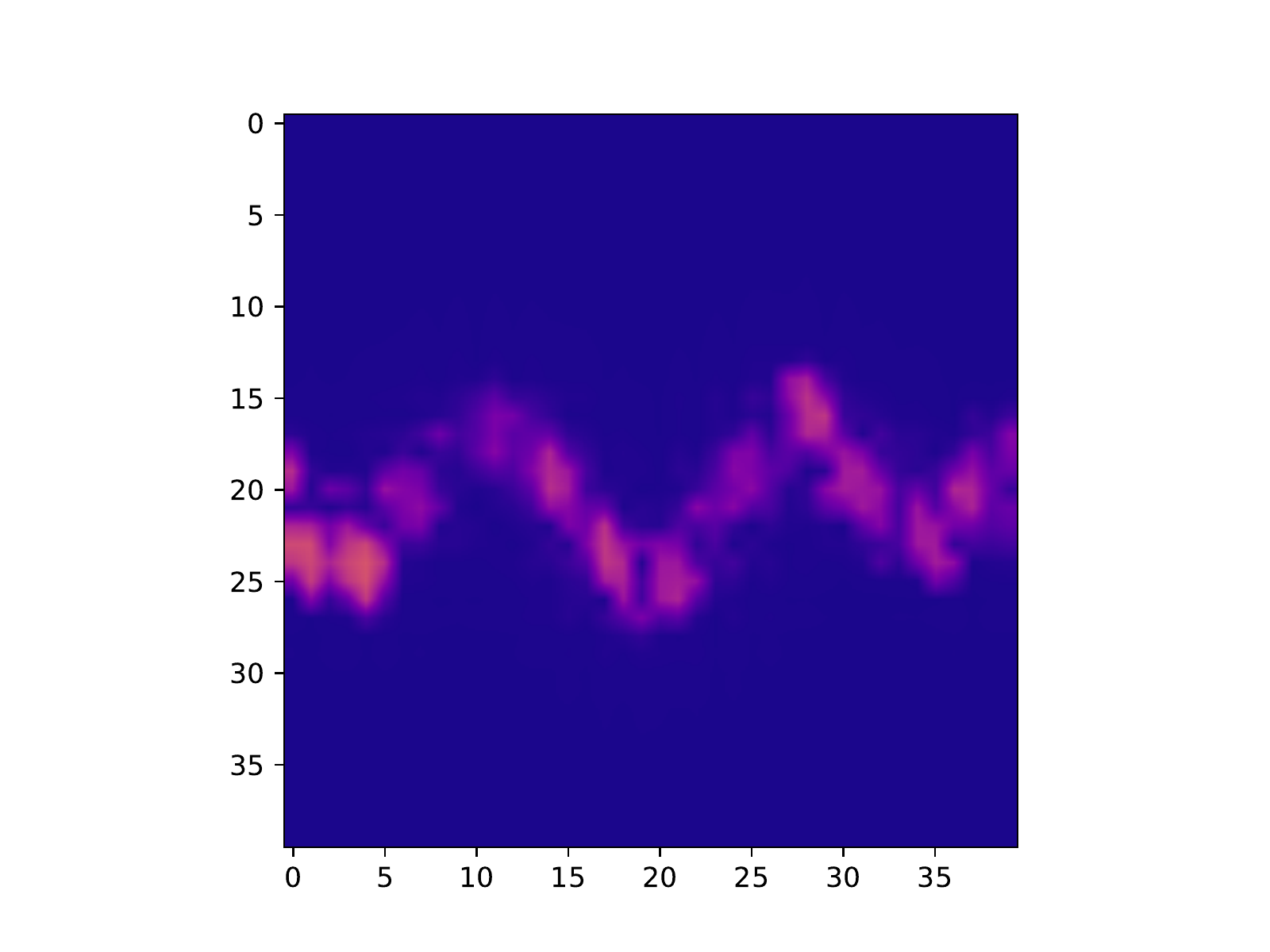}
	\includegraphics[width=0.1\linewidth, trim= 103 38 87 60, clip]{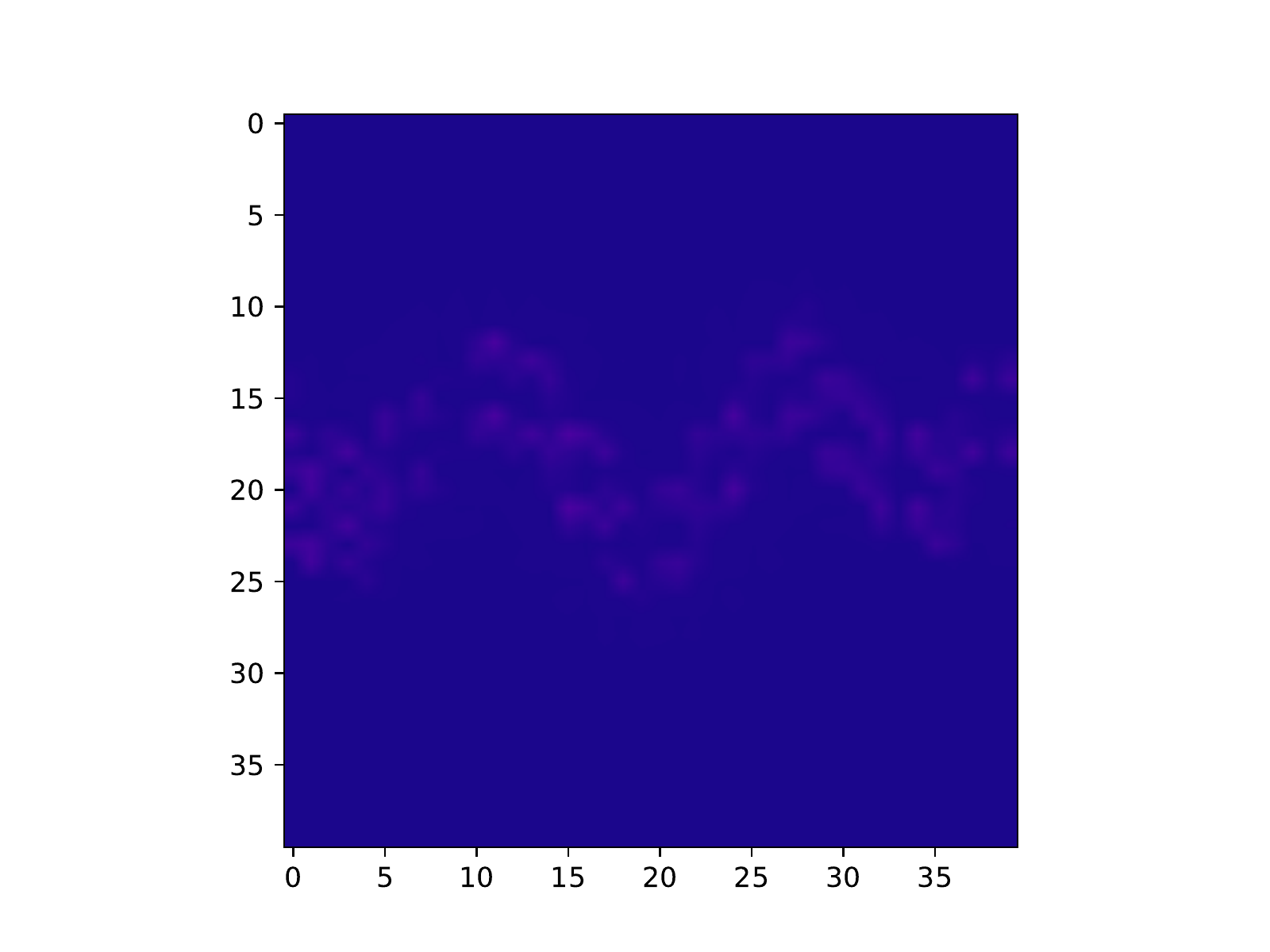}
	\includegraphics[width=0.1\linewidth, trim= 103 38 87 60, clip]{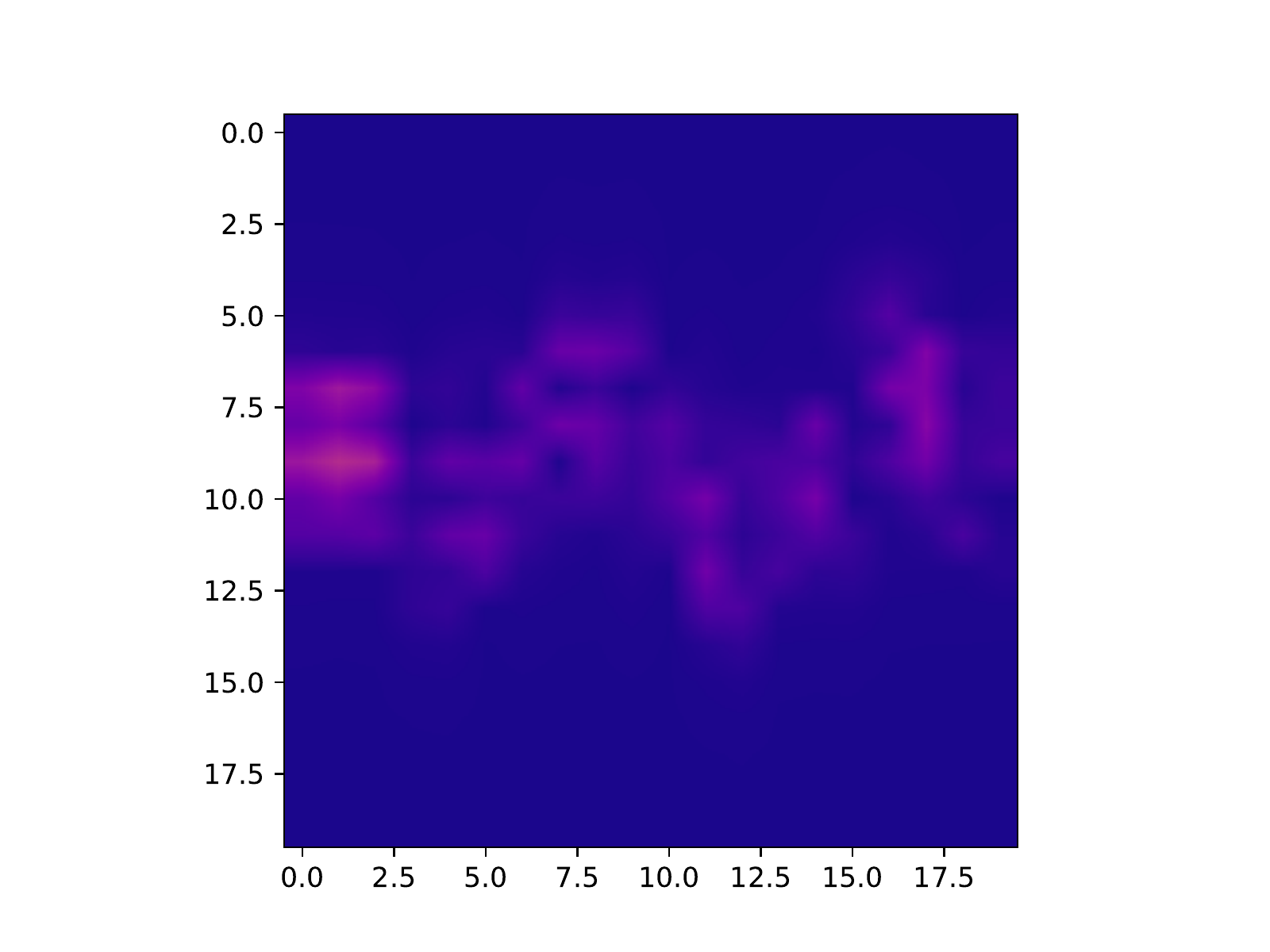}
	\includegraphics[width=0.1\linewidth, trim= 103 38 87 60, clip]{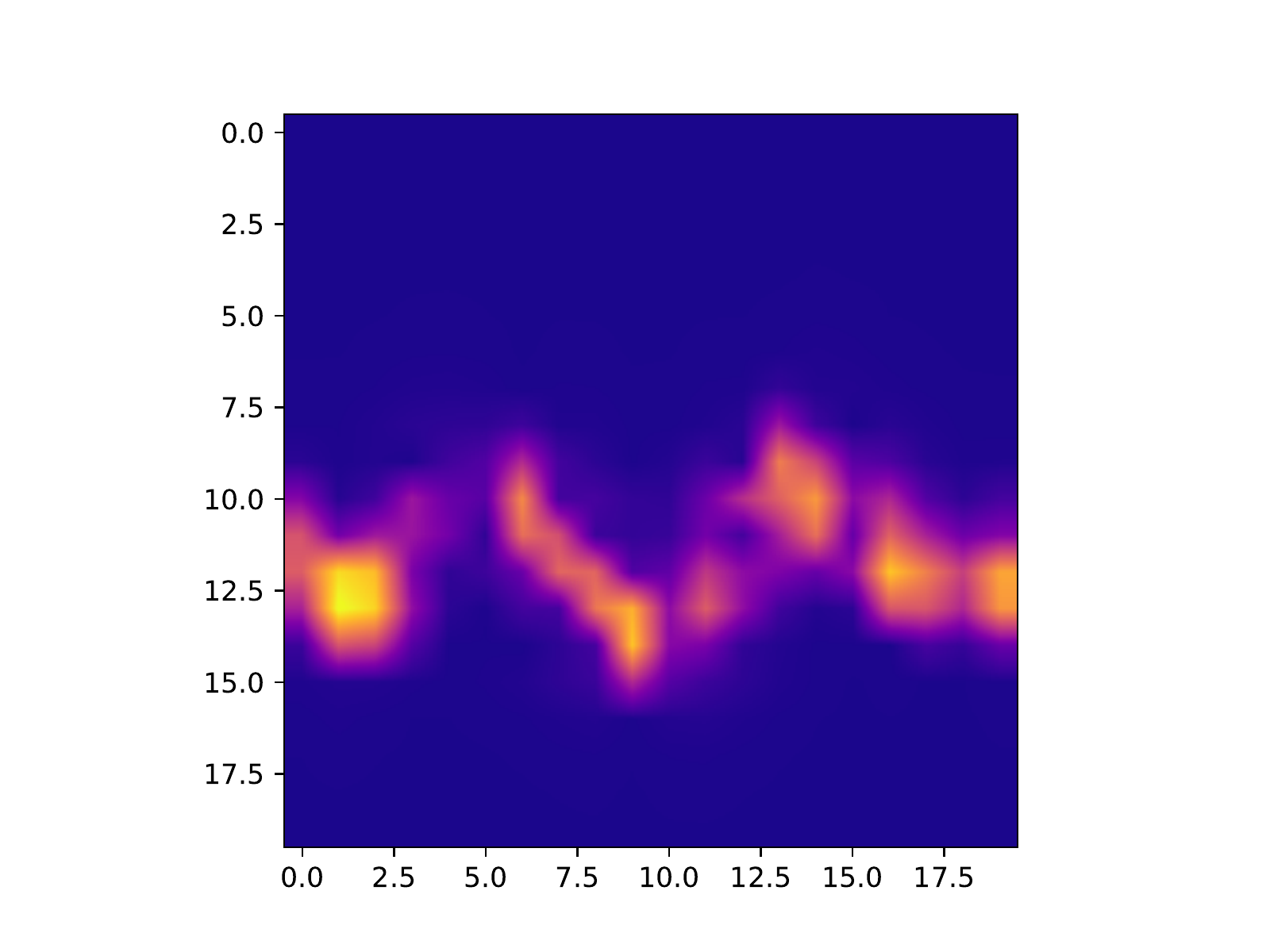}
	\includegraphics[width=0.1\linewidth, trim= 103 38 87 60, clip]{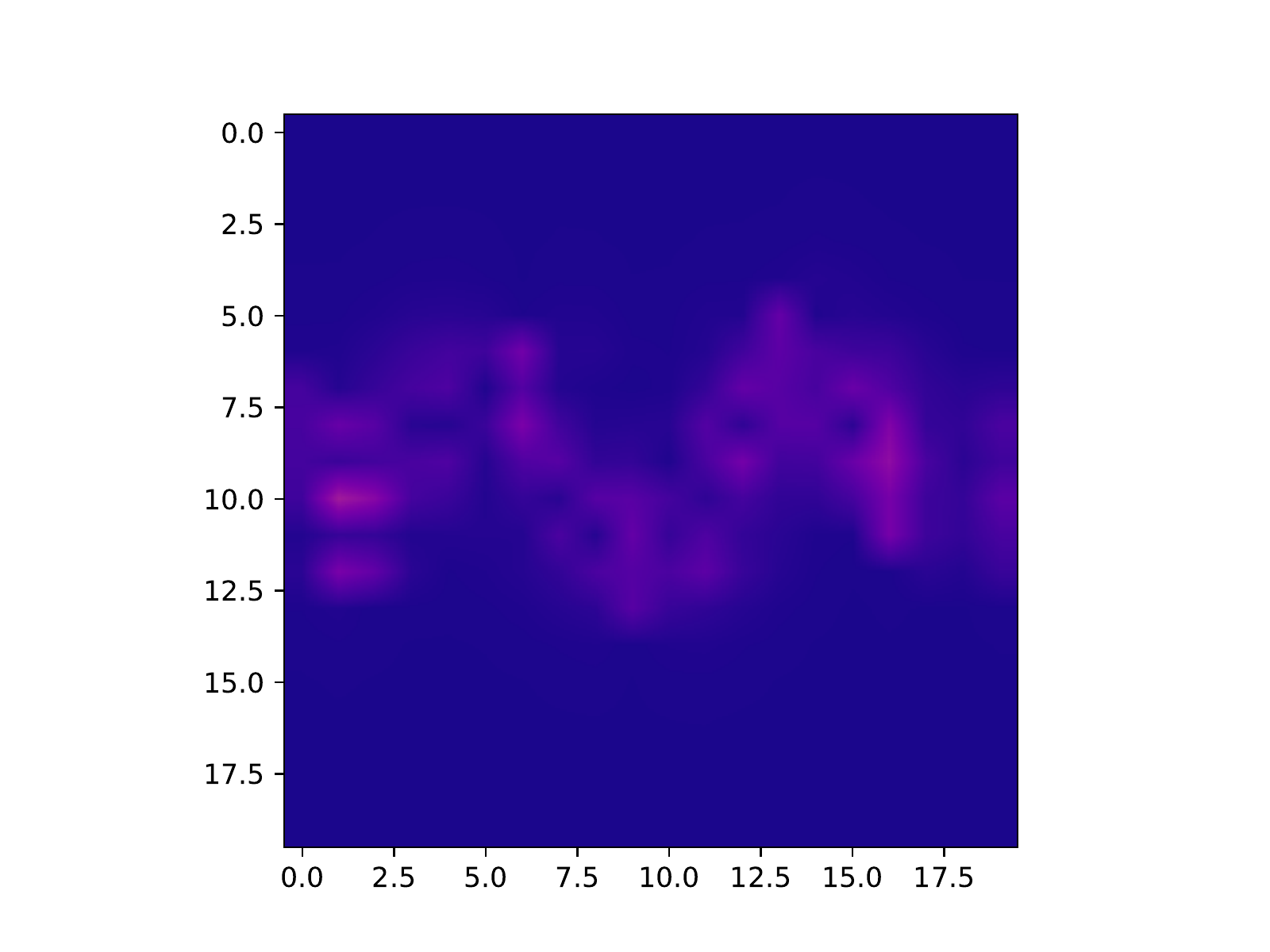}
	\includegraphics[width=0.1\linewidth, trim= 103 38 87 60, clip]{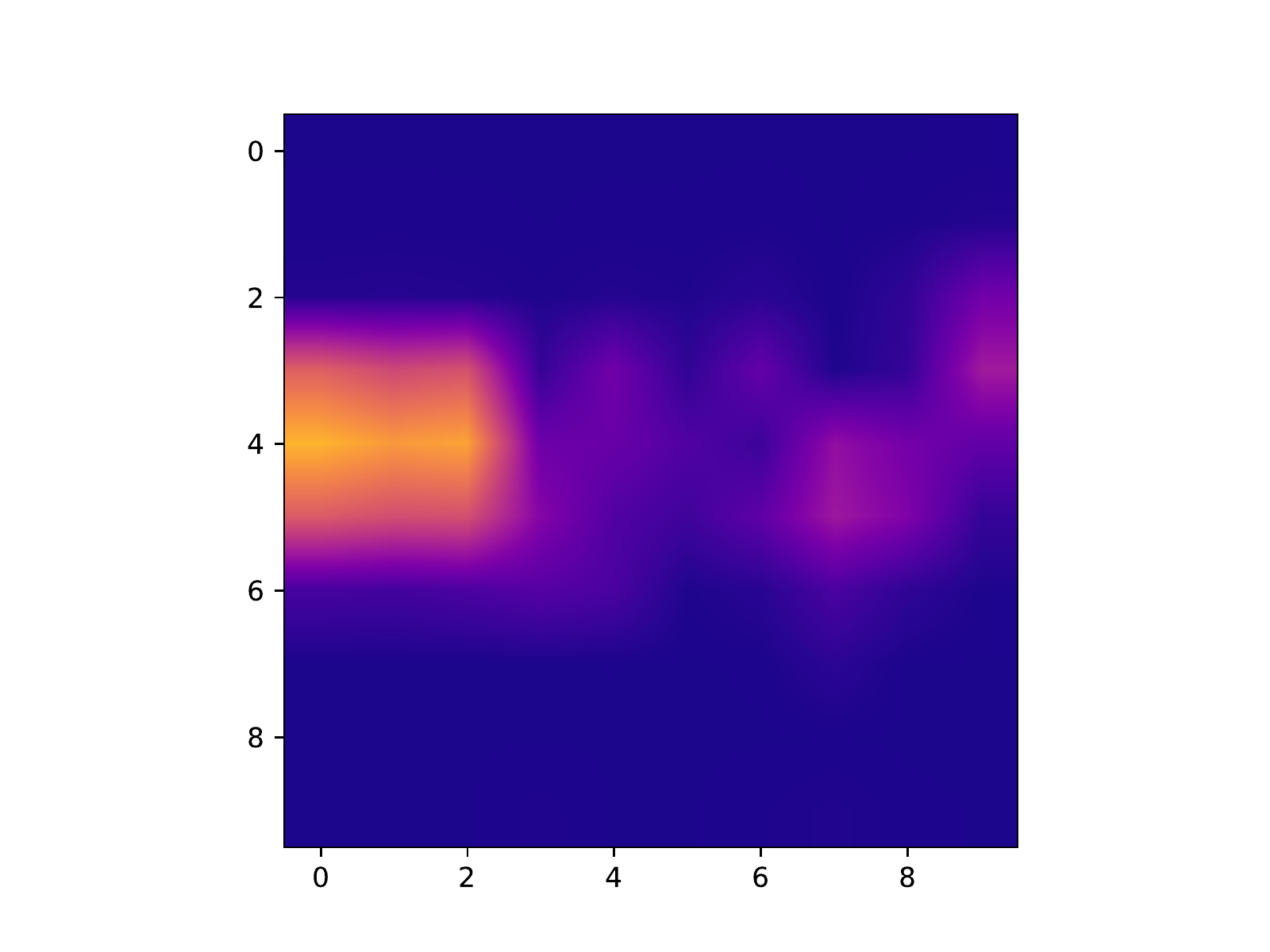}
	\includegraphics[width=0.1\linewidth, trim= 103 38 87 60, clip]{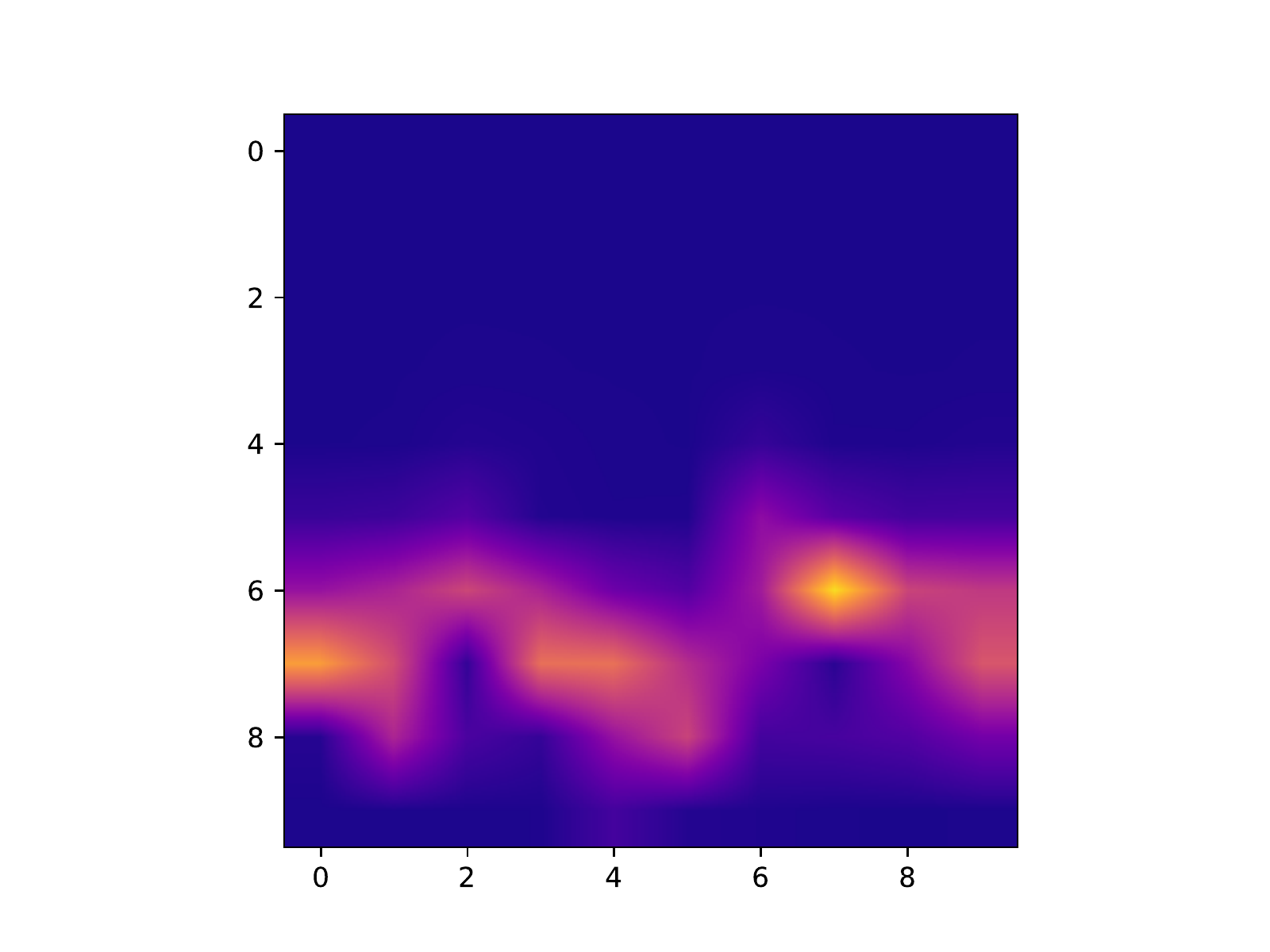}
	\includegraphics[width=0.1\linewidth, trim= 103 38 87 60, clip]{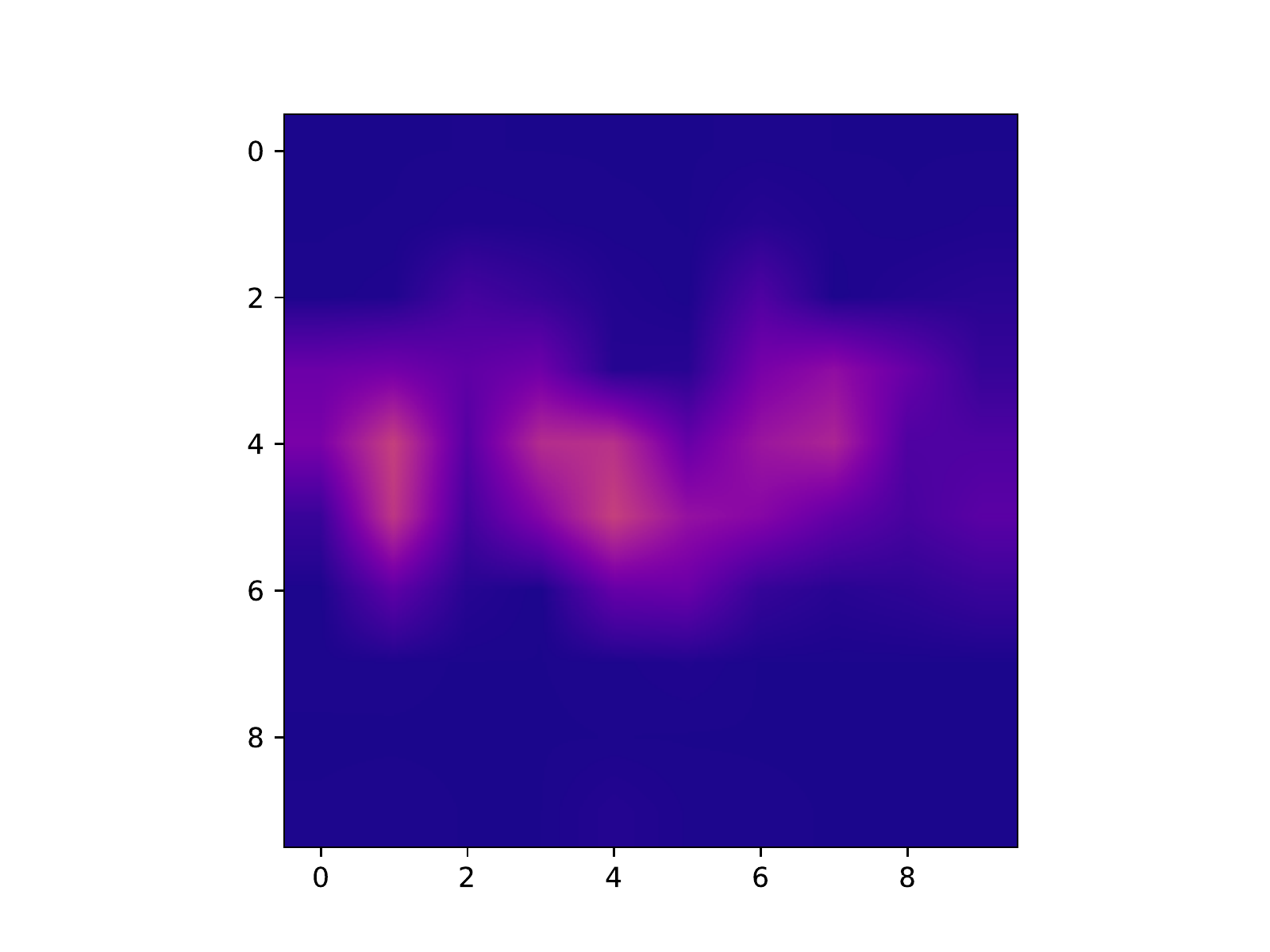}
	
	\includegraphics[width=0.1\linewidth, trim= 103 38 87 60, clip]{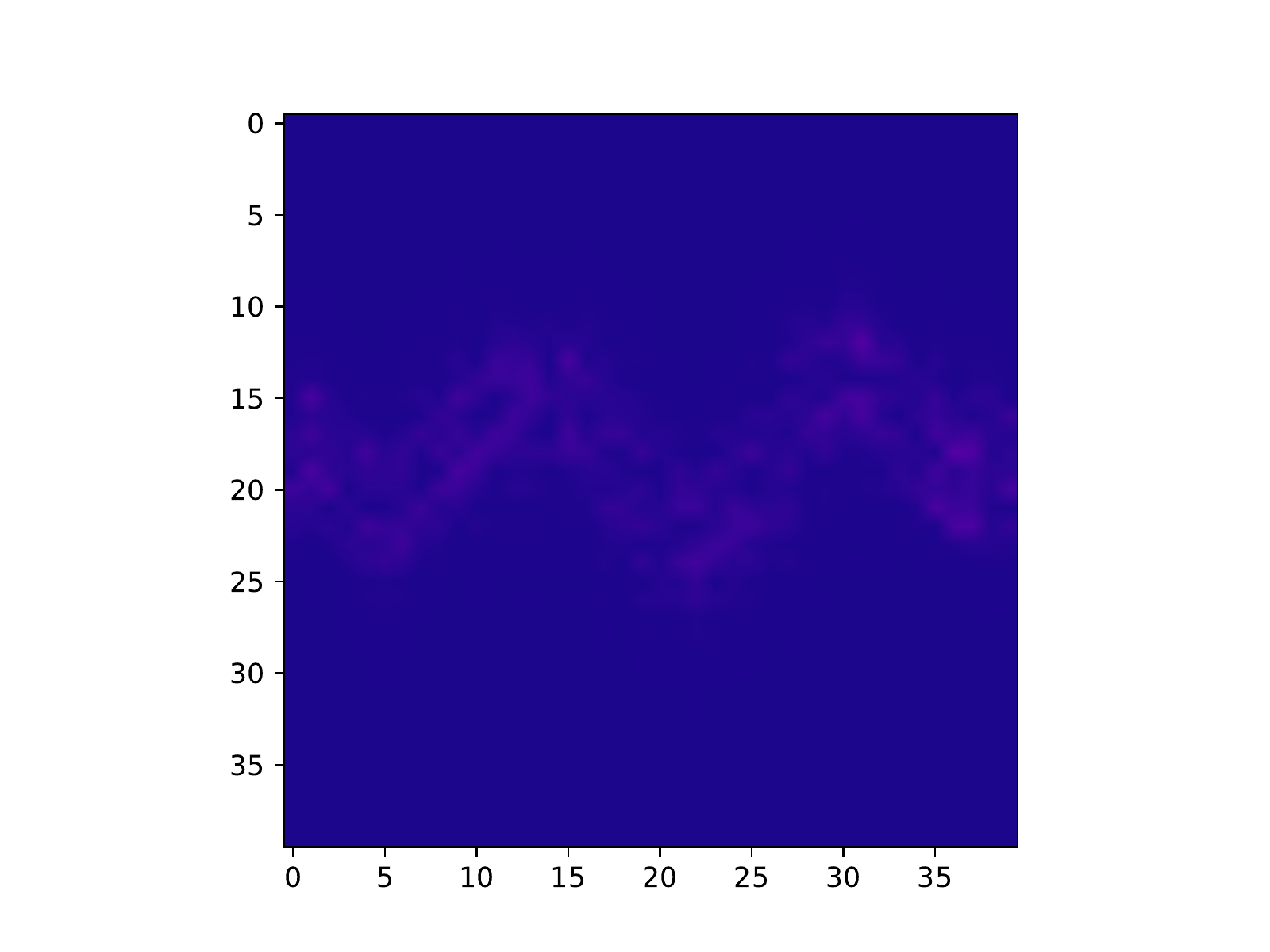}
	\includegraphics[width=0.1\linewidth, trim= 103 38 87 60, clip]{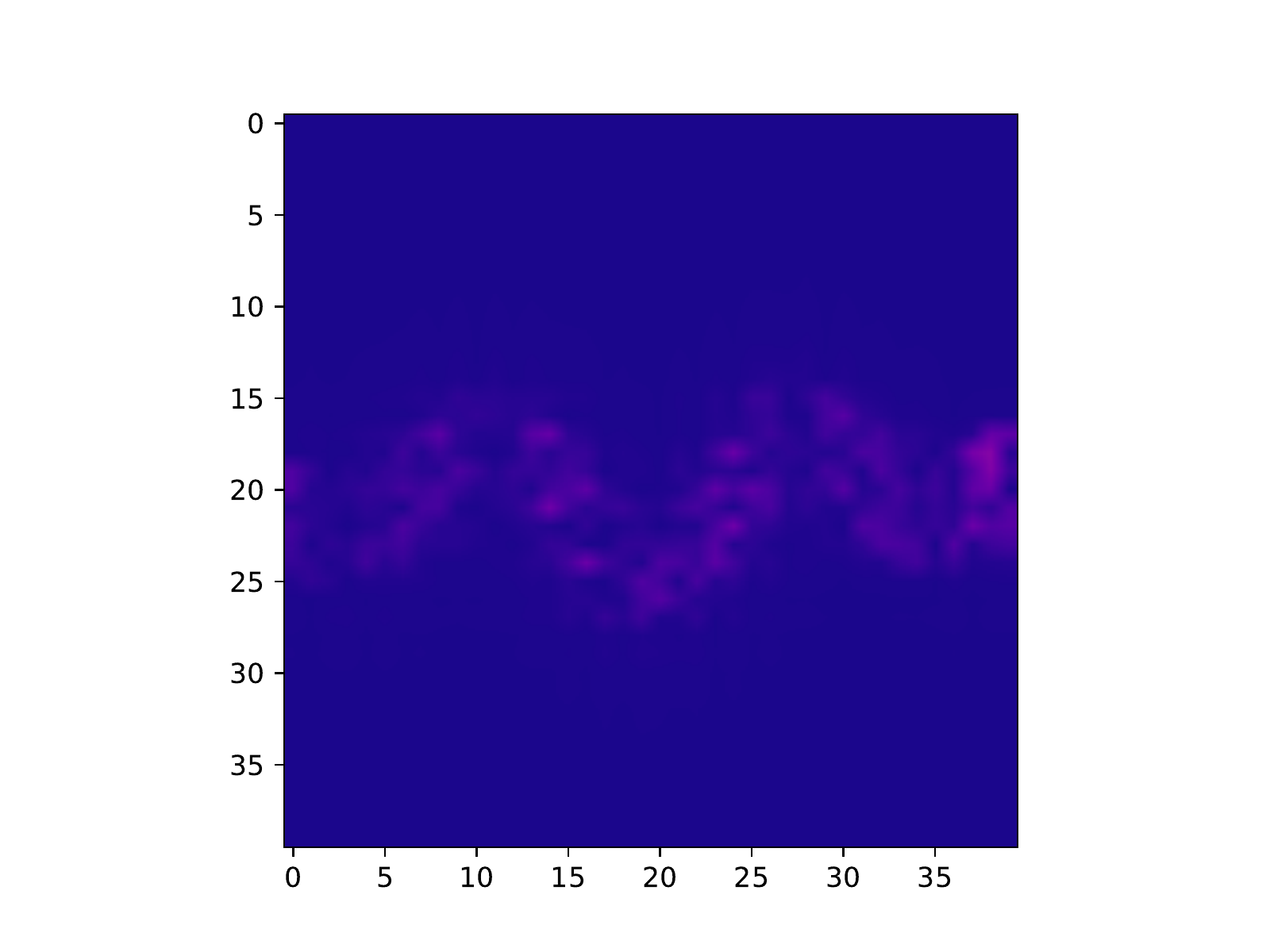}
	\includegraphics[width=0.1\linewidth, trim= 103 38 87 60, clip]{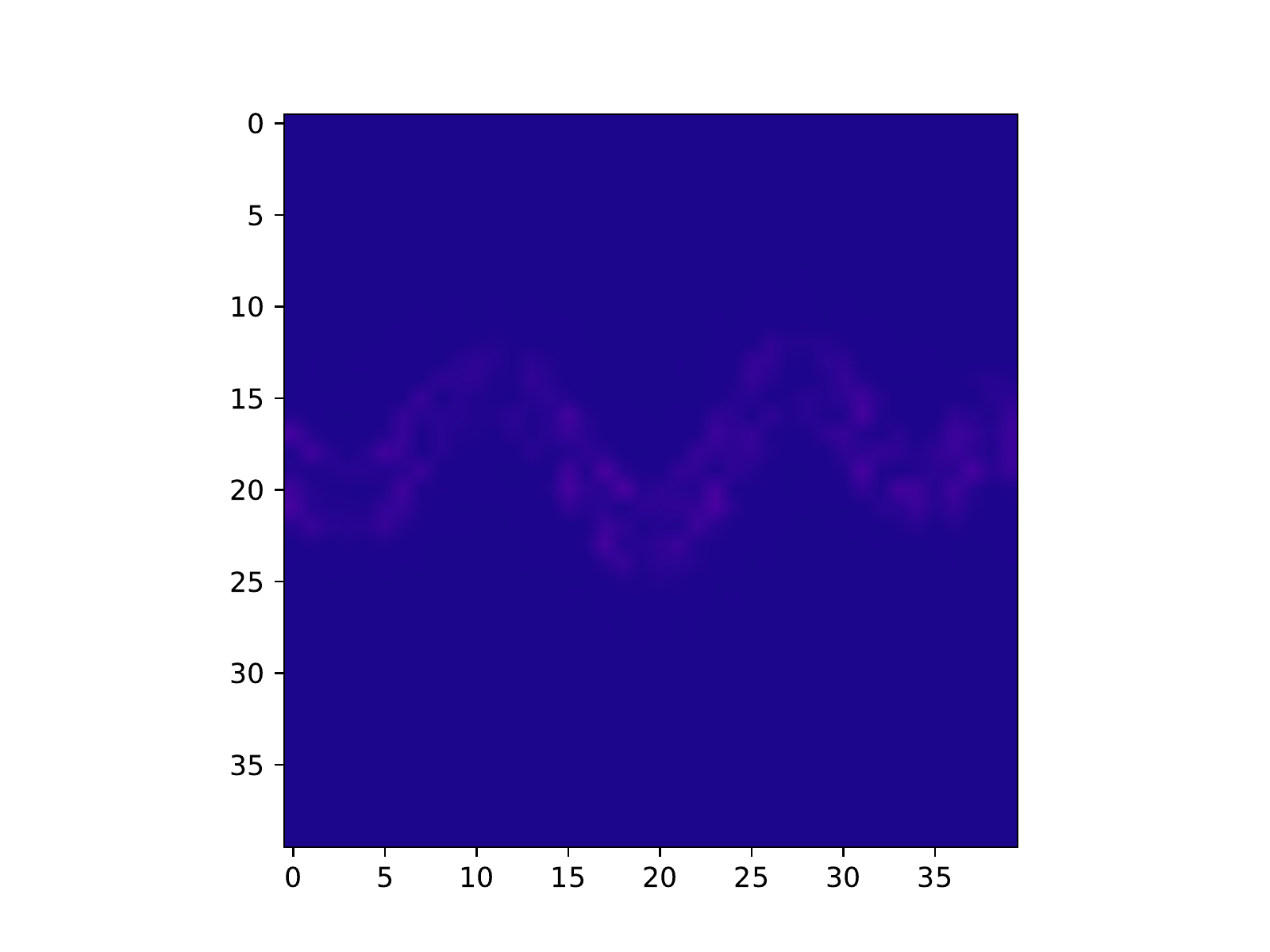}
	\includegraphics[width=0.1\linewidth, trim= 103 38 87 60, clip]{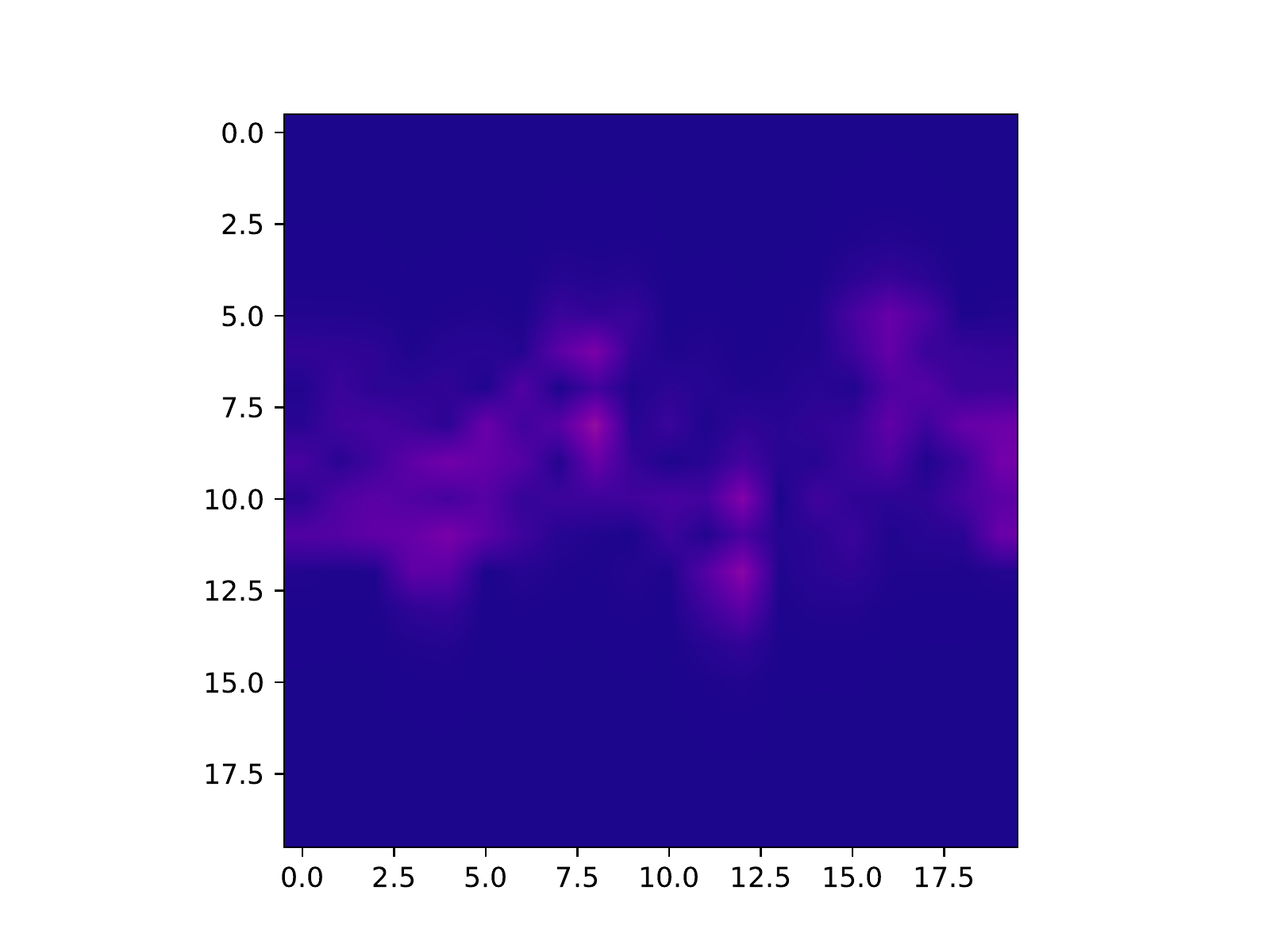}
	\includegraphics[width=0.1\linewidth, trim= 103 38 87 60, clip]{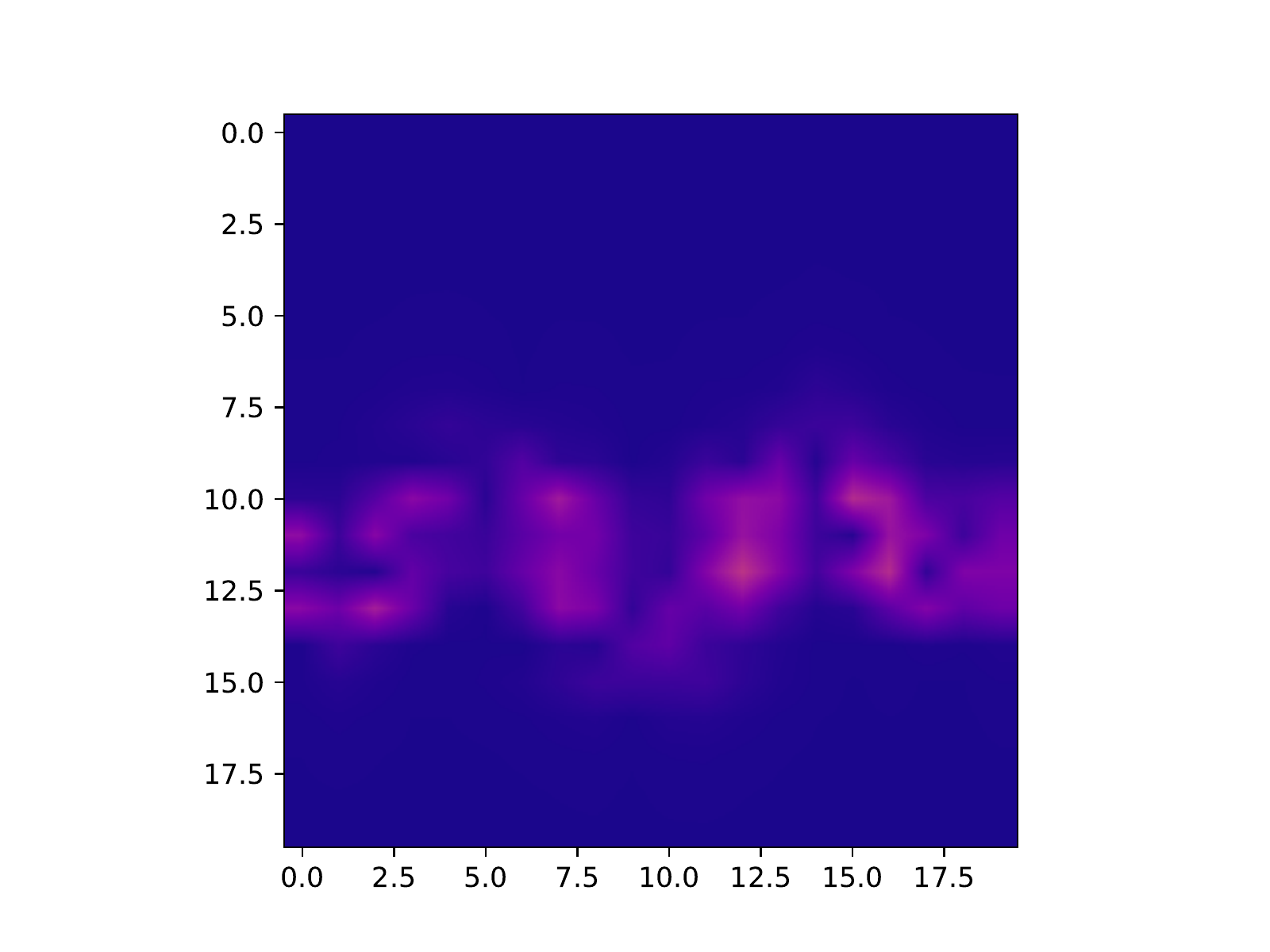}
	\includegraphics[width=0.1\linewidth, trim= 103 38 87 60, clip]{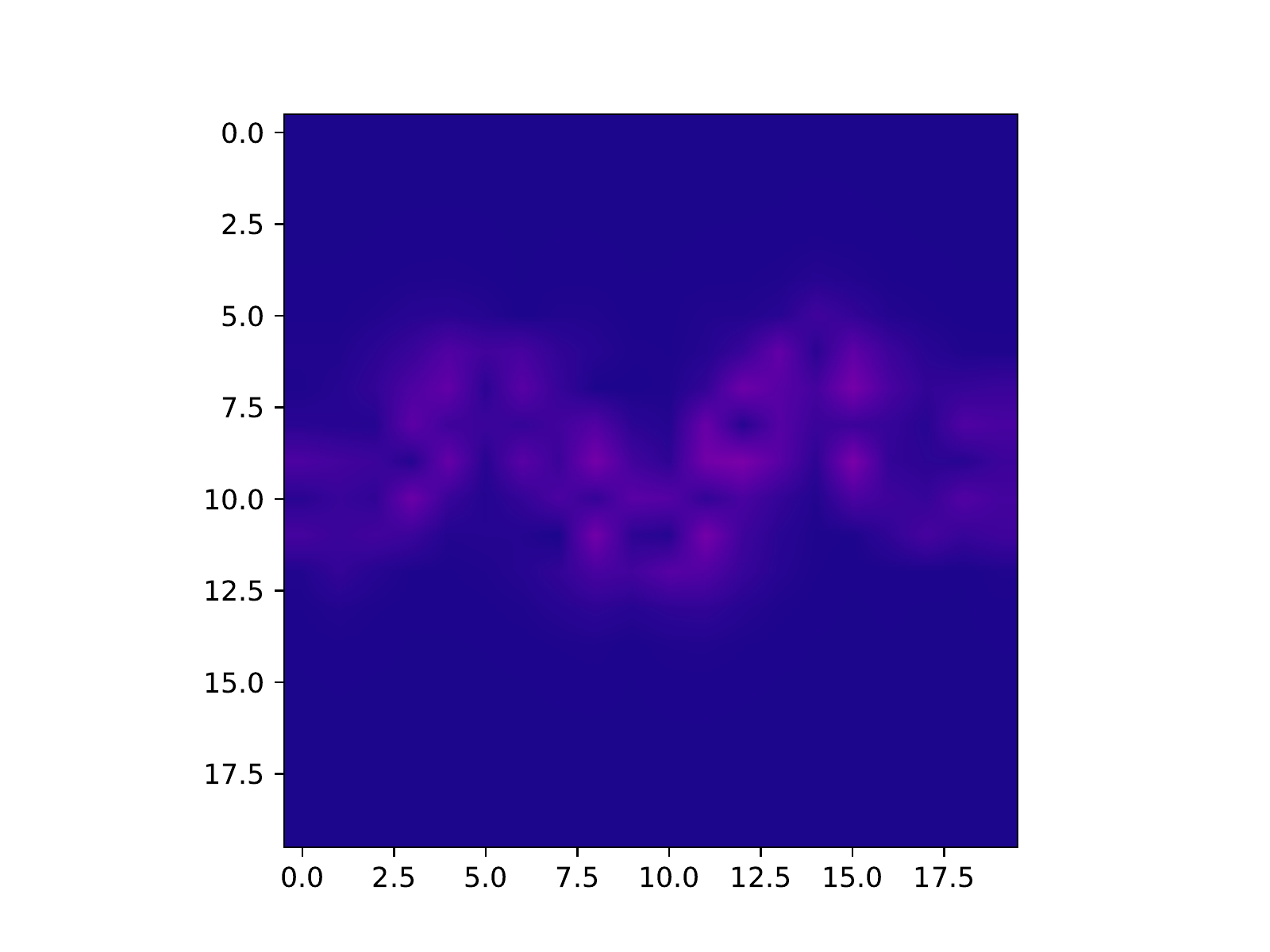}
	\includegraphics[width=0.1\linewidth, trim= 103 38 87 60, clip]{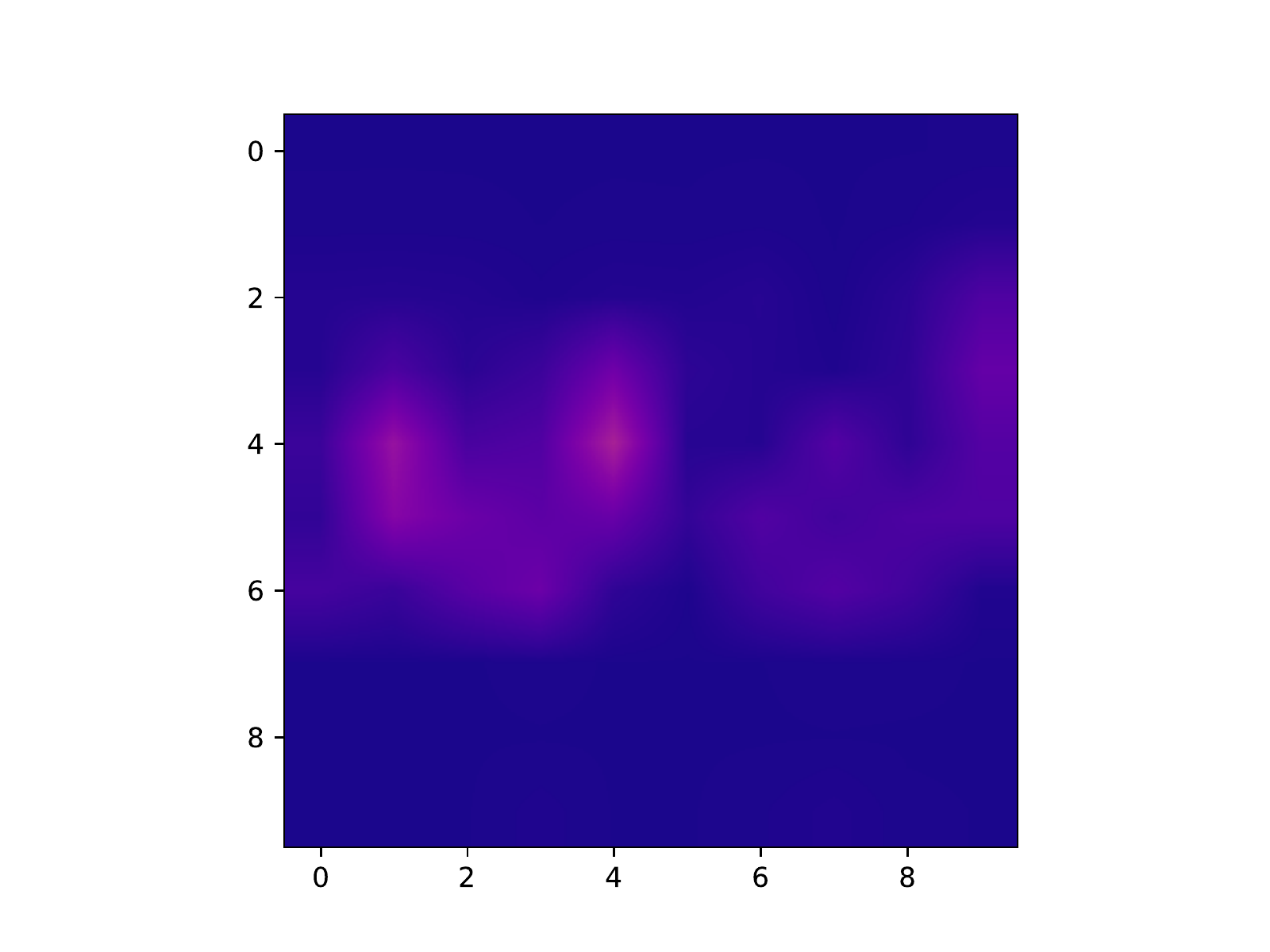}
	\includegraphics[width=0.1\linewidth, trim= 103 38 87 60, clip]{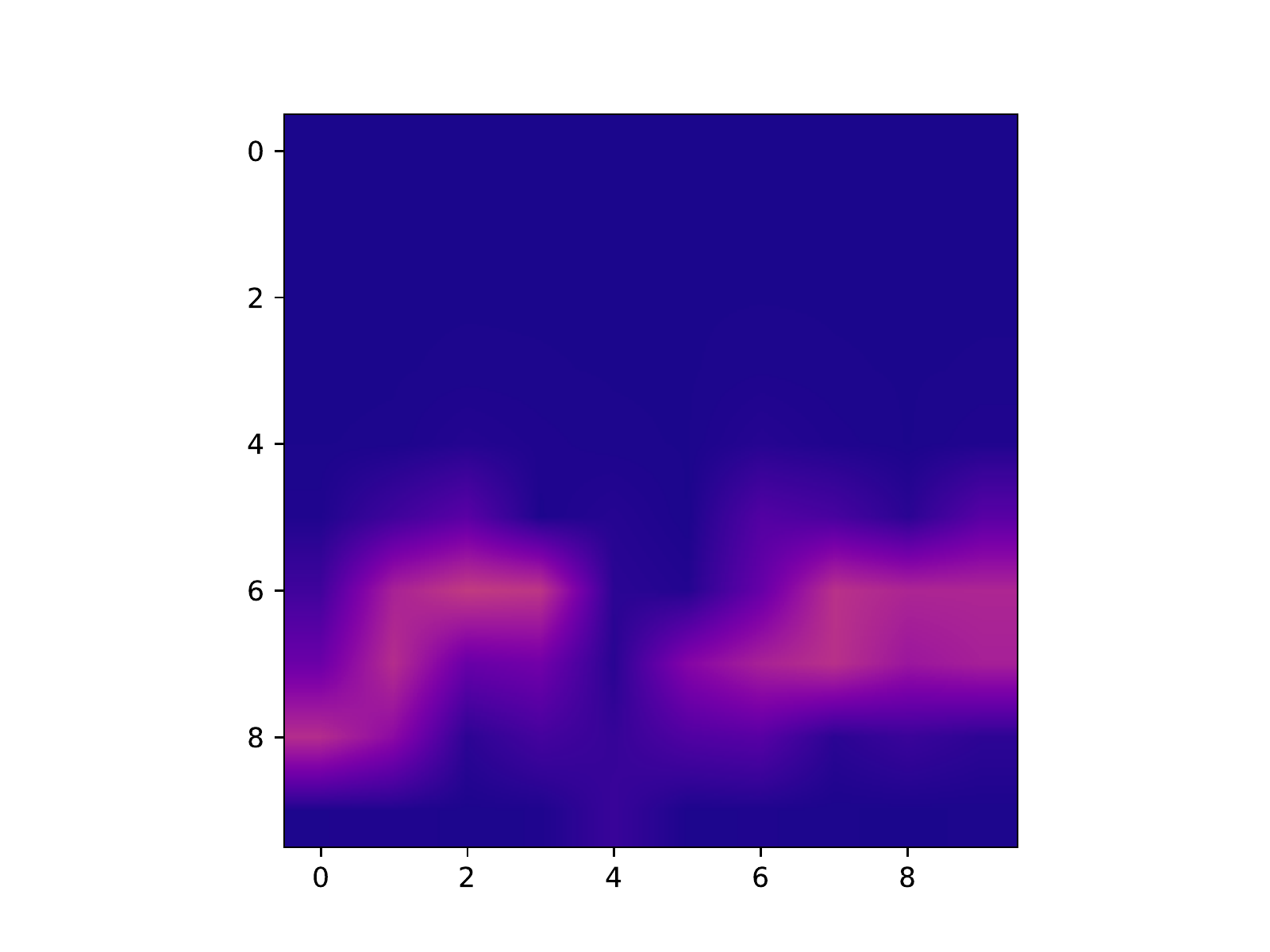}
	\includegraphics[width=0.1\linewidth, trim= 103 38 87 60, clip]{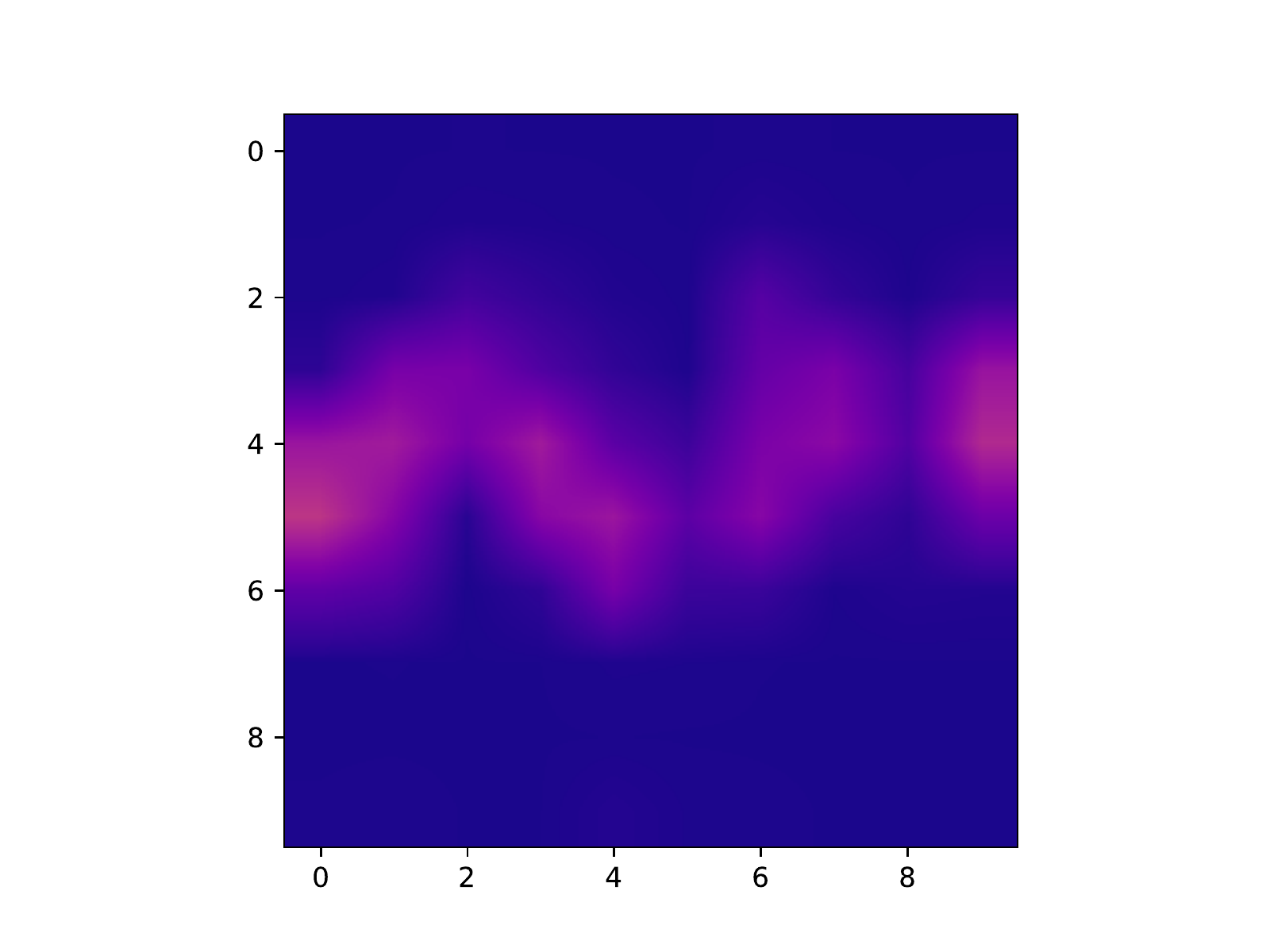}
	
	\includegraphics[width=0.1\linewidth, trim= 103 38 87 60, clip]{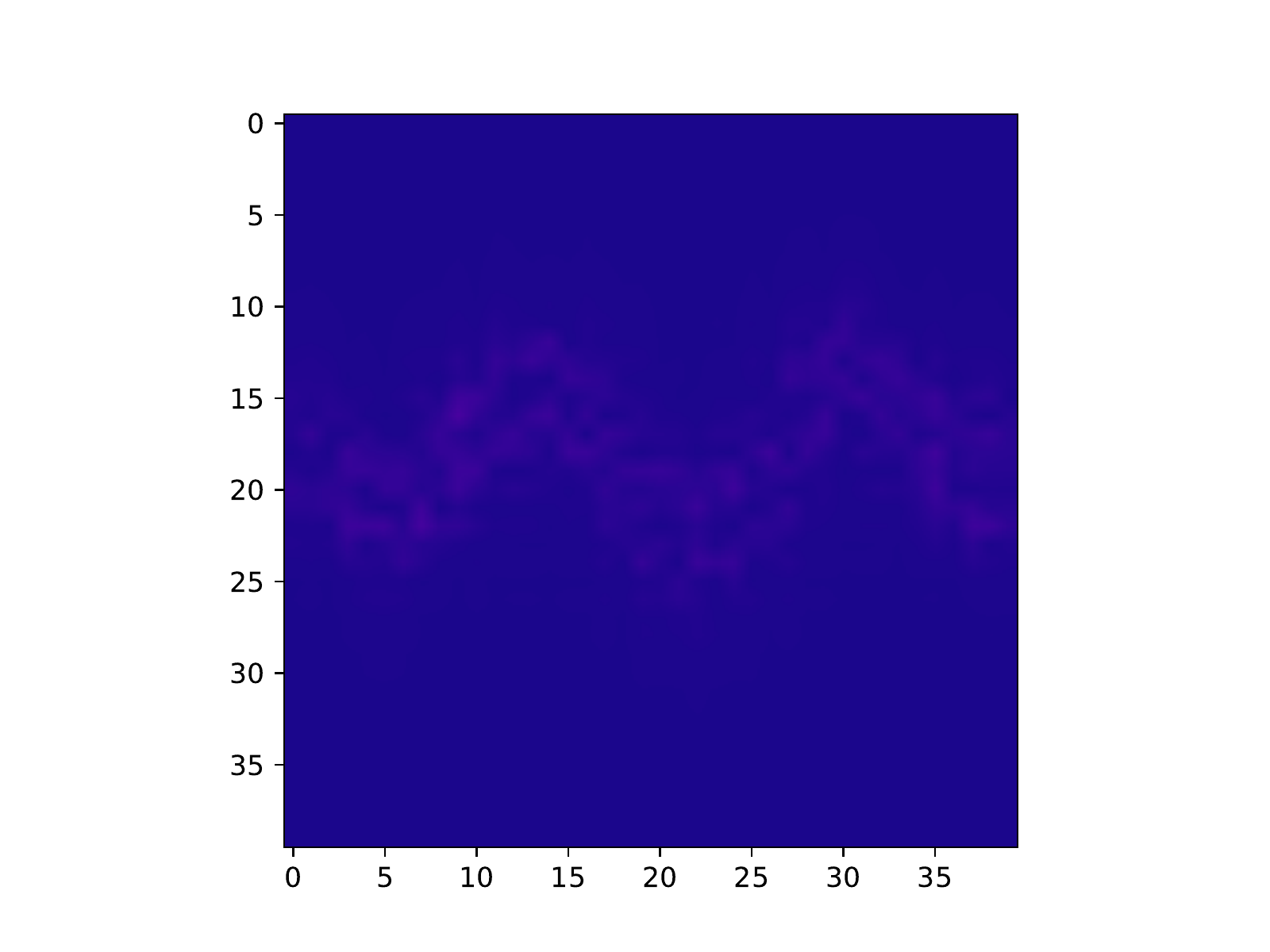}
	\includegraphics[width=0.1\linewidth, trim= 103 38 87 60, clip]{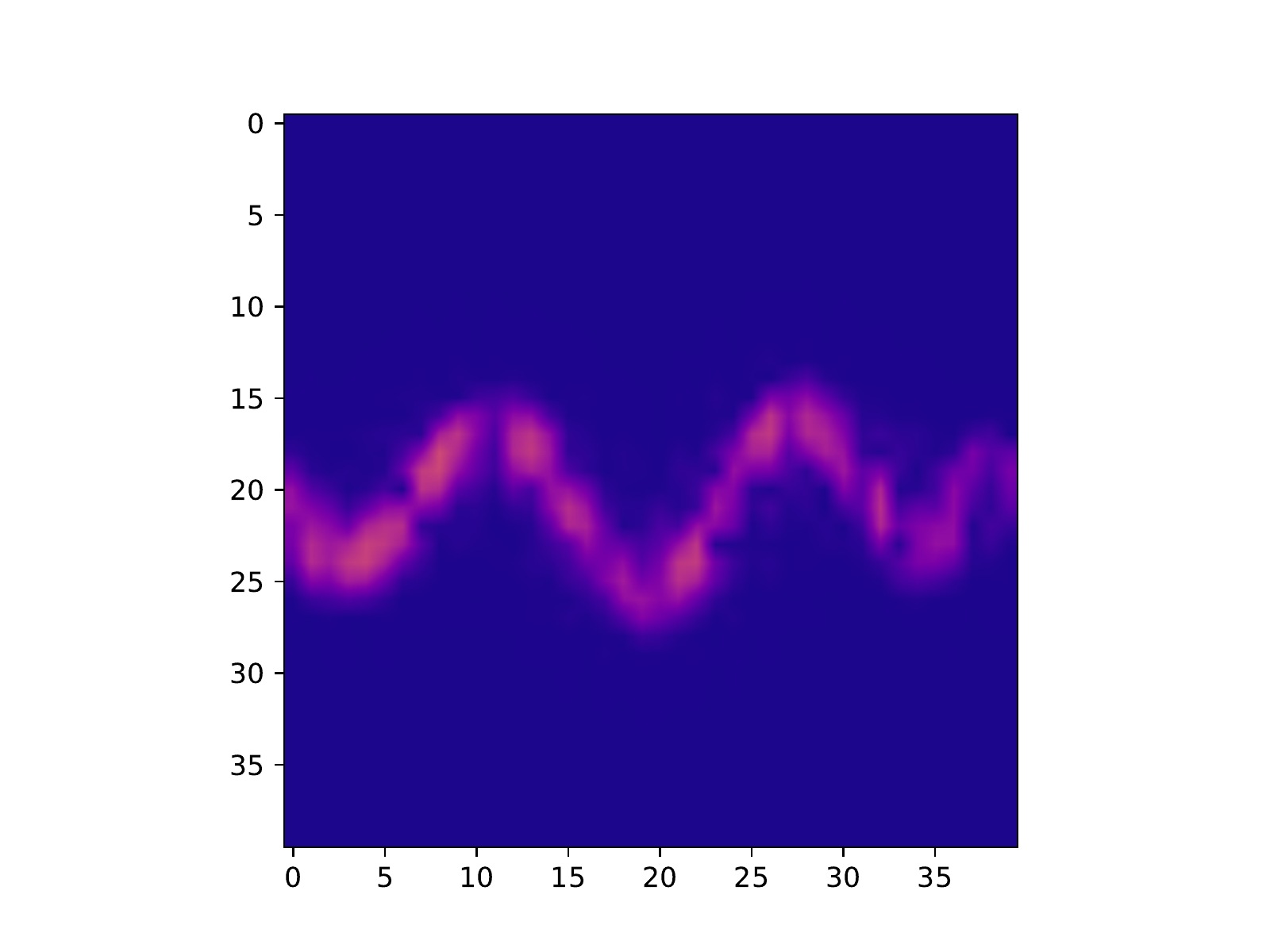}
	\includegraphics[width=0.1\linewidth, trim= 103 38 87 60, clip]{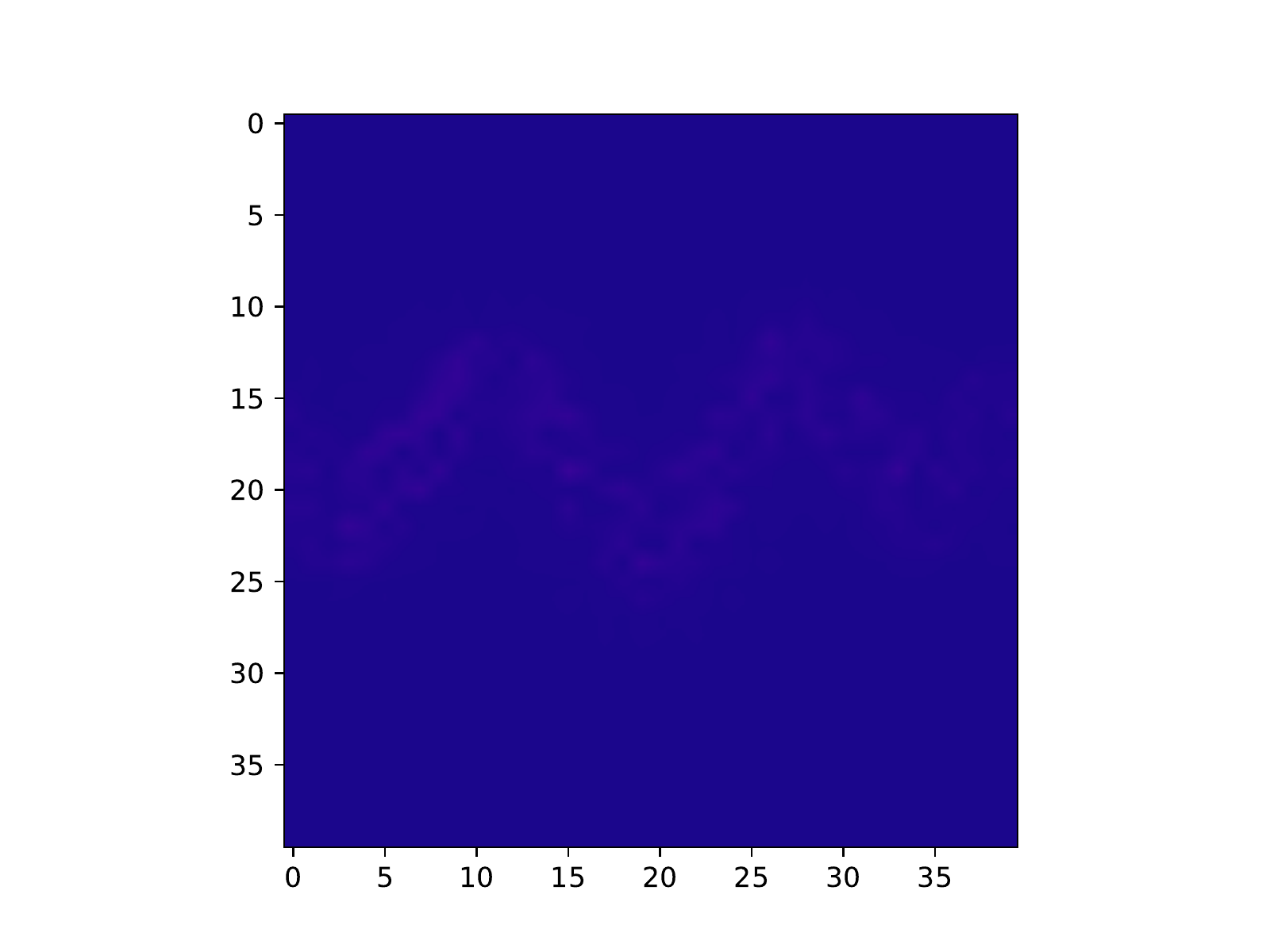}
	\includegraphics[width=0.1\linewidth, trim= 103 38 87 60, clip]{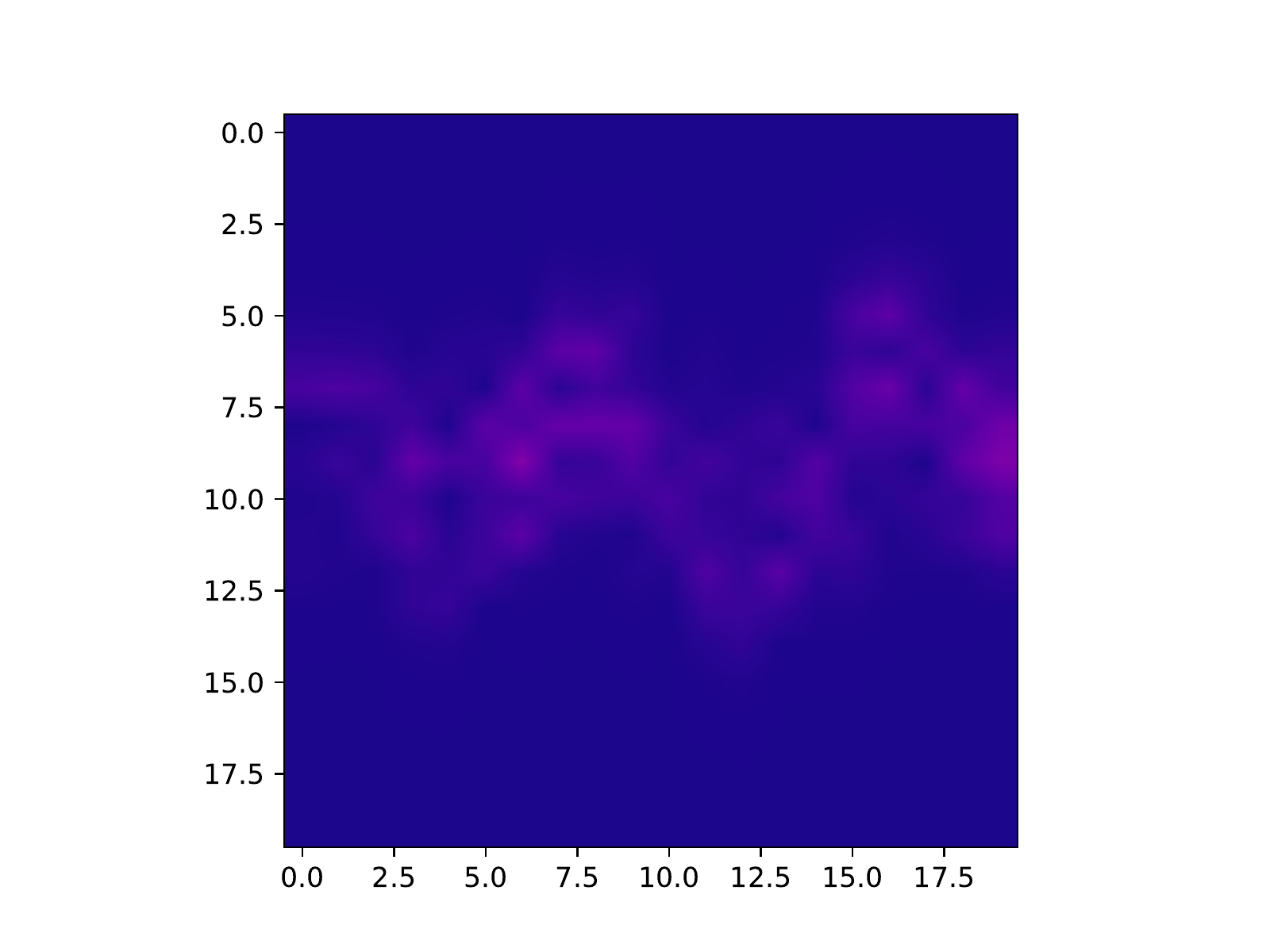}
	\includegraphics[width=0.1\linewidth, trim= 103 38 87 60, clip]{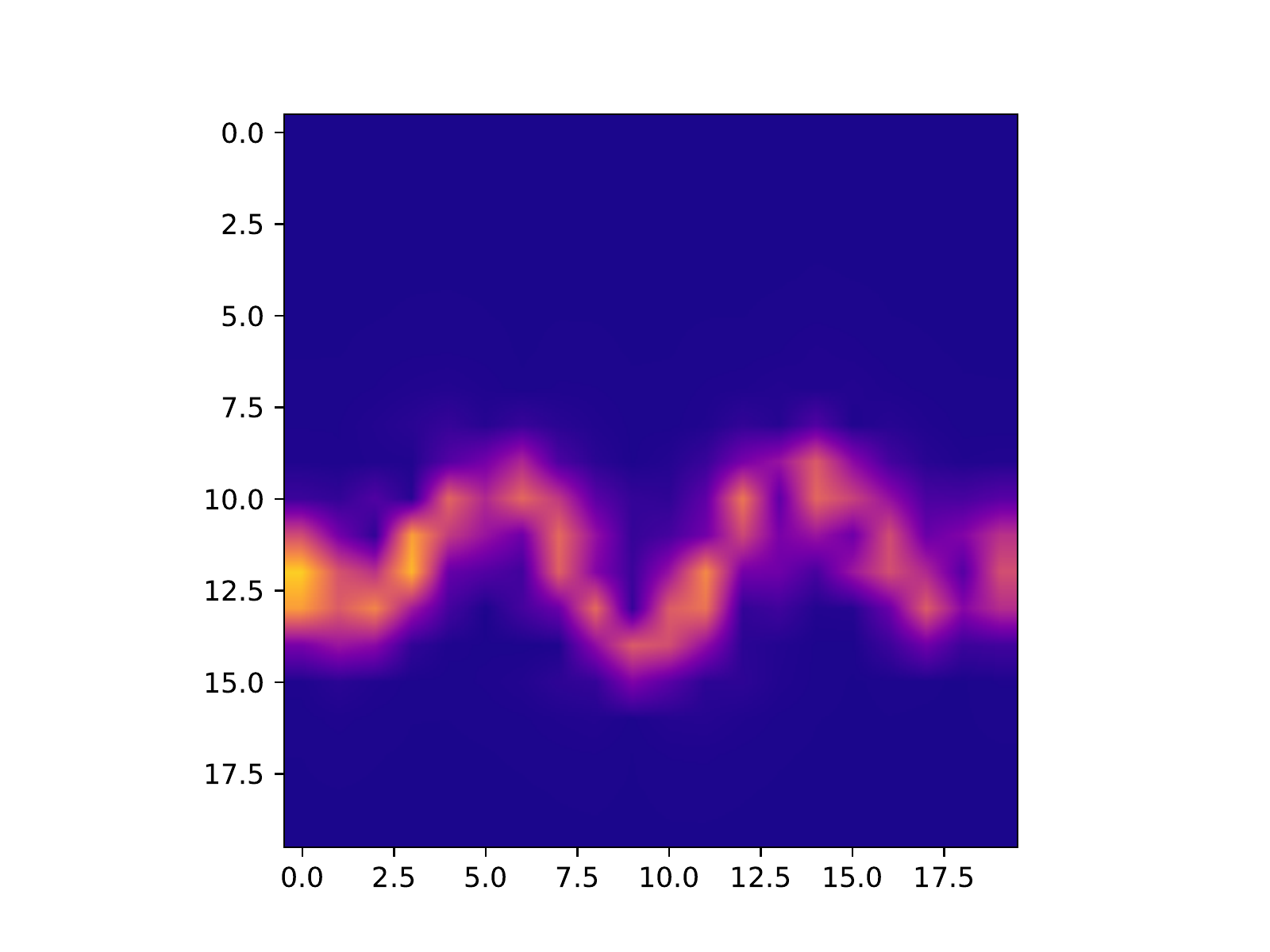}
	\includegraphics[width=0.1\linewidth, trim= 103 38 87 60, clip]{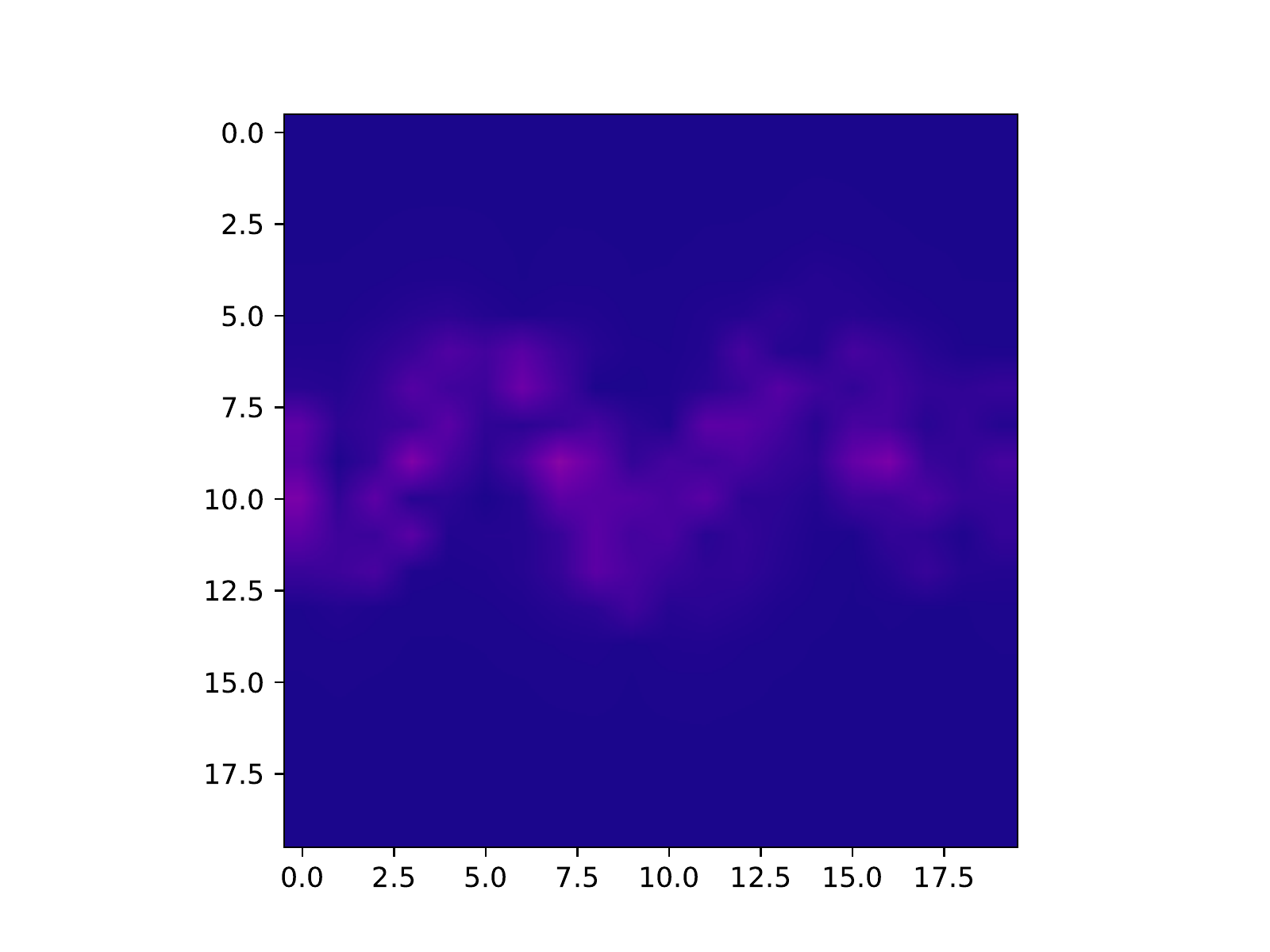}
	\includegraphics[width=0.1\linewidth, trim= 103 38 87 60, clip]{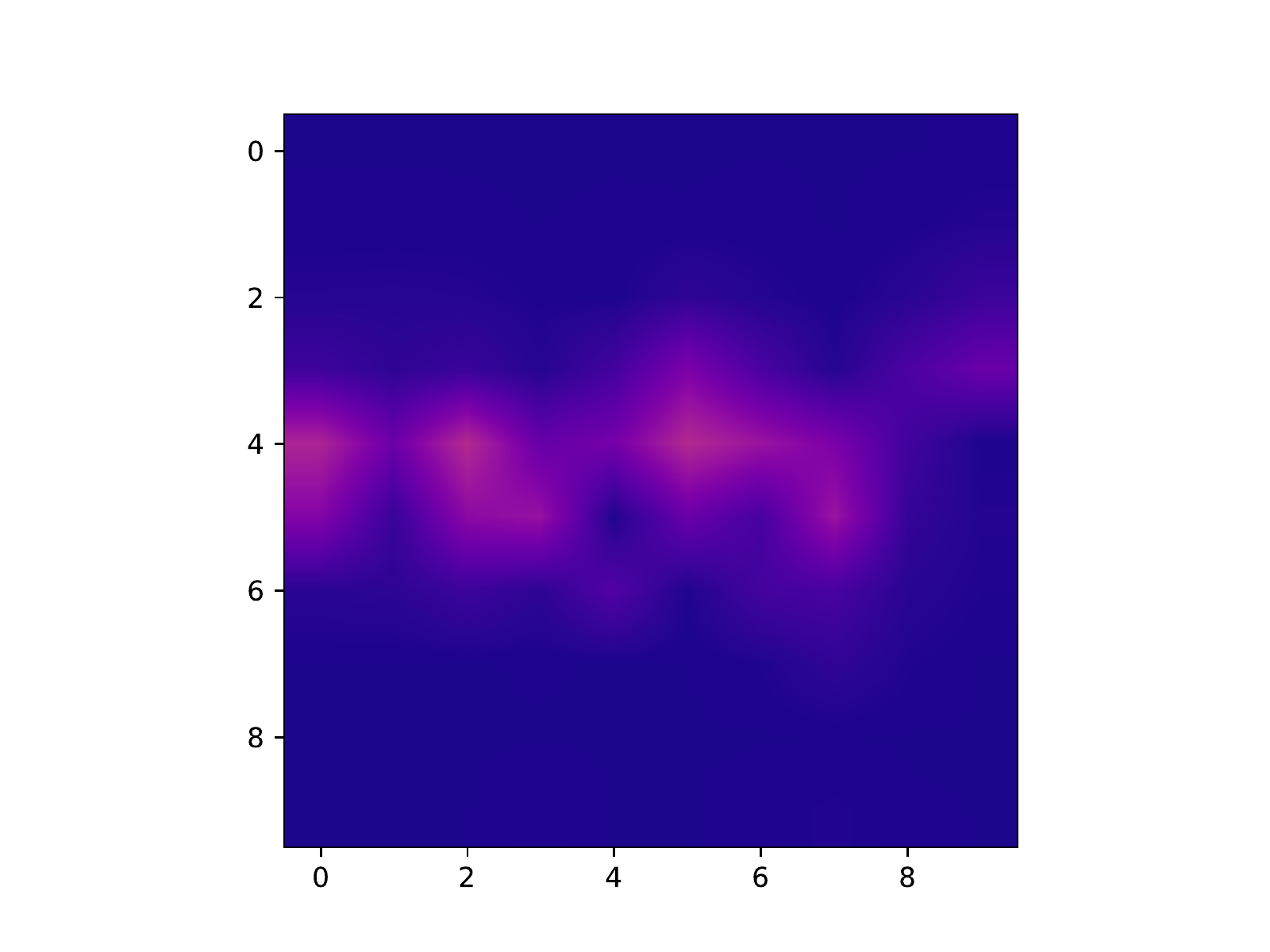}
	\includegraphics[width=0.1\linewidth, trim= 103 38 87 60, clip]{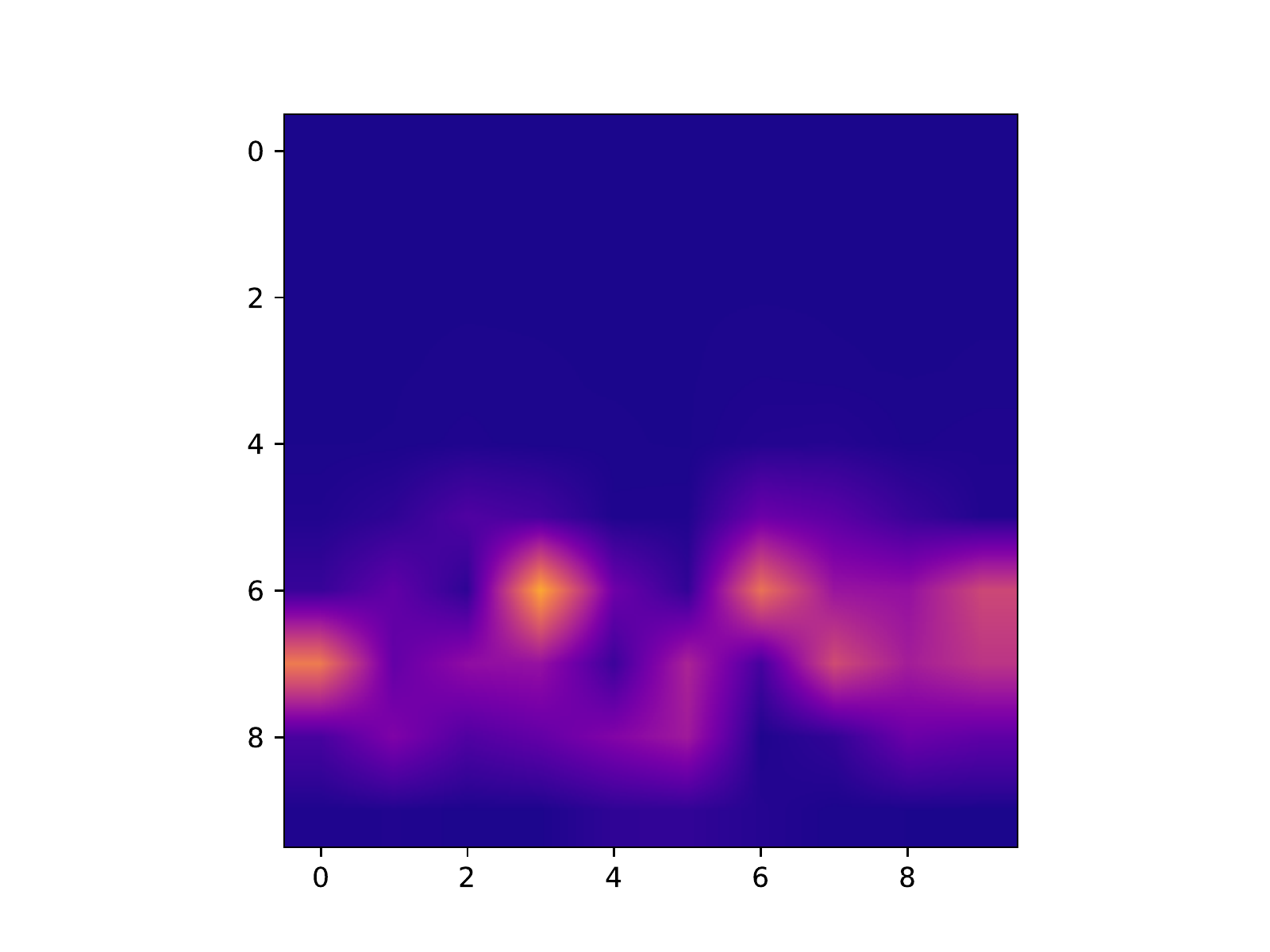}
	\includegraphics[width=0.1\linewidth, trim= 103 38 87 60, clip]{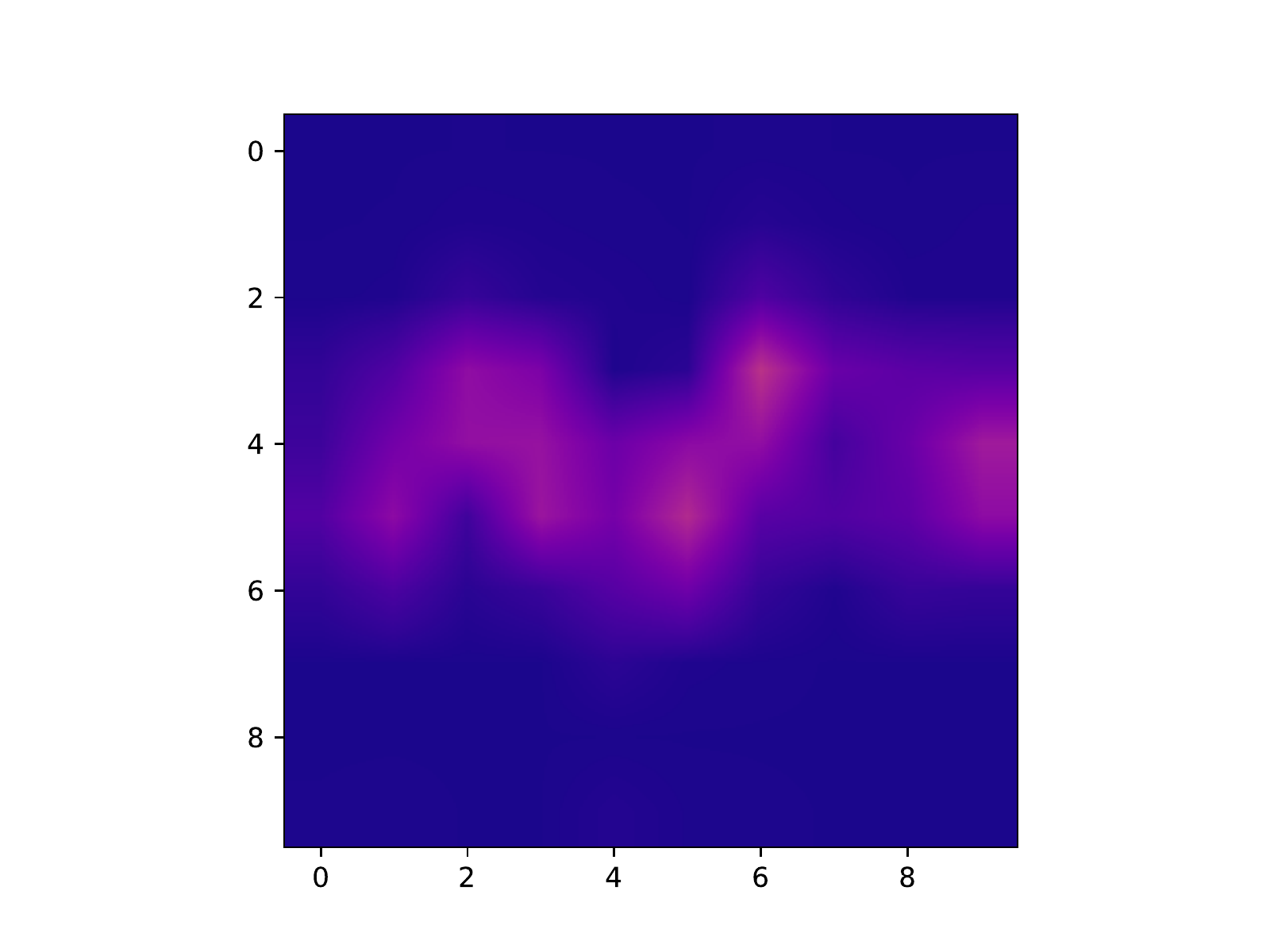}
	\subcaption*{Spatial frequency.}
	\end{subfigure}
	
    \vspace{2mm}
	\centering
    \begin{subfigure}[c]{0.082\textwidth}
	\begin{overpic}[width=\linewidth, trim= 640 200 640 400, clip]{data/2d/wlt/110.jpg} 
	    \put(-50,50){\small \color{black}{$a)$}}\end{overpic}
	    
    \vspace{0.5mm}
	\begin{overpic}[width=\linewidth, trim= 1260 200 20 400, clip]{data/2d/mse/110.jpg} 
	    \put(-50,50){\small \color{black}{$b)$}}\end{overpic}
	    
    \vspace{0.5mm}
	\begin{overpic}[width=\linewidth, trim= 1260 200 20 400, clip]{data/2d/wlt/110.jpg} 
	    \put(-50,50){\small \color{black}{$c)$}}\end{overpic}~
				\subcaption*{Data.}
	\end{subfigure}
	\hspace{1mm}
	\vline
	\hspace{1mm}
    \begin{subfigure}[c]{0.7\textwidth}
	\includegraphics[width=0.56\linewidth, trim= 58 148 50 150, clip]{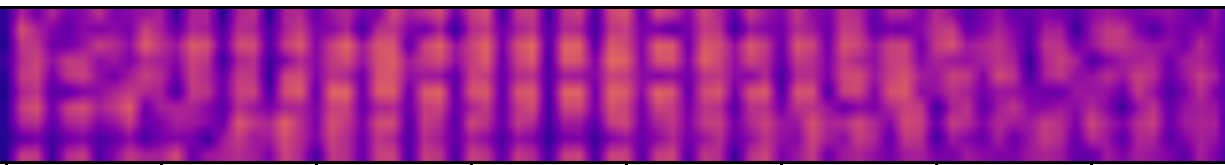}
	\includegraphics[width=0.28\linewidth, trim= 58 125 50 120, clip]{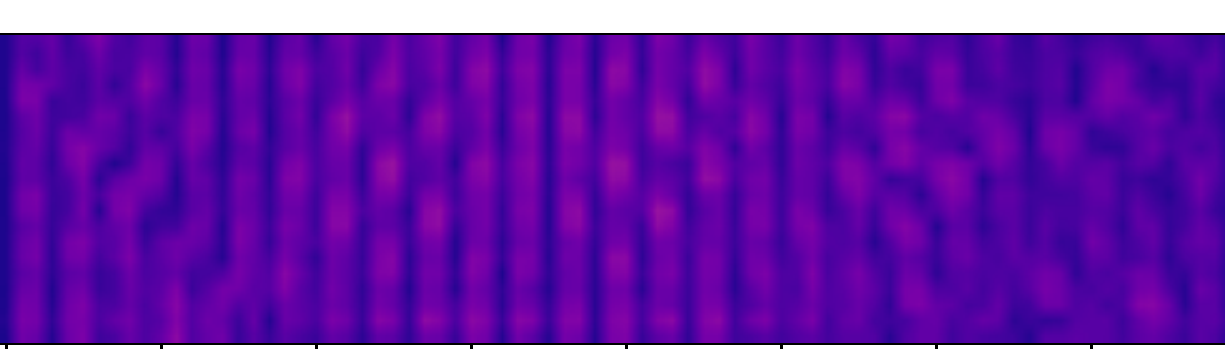}
	\includegraphics[width=0.14\linewidth, trim= 58 80 50 80, clip]{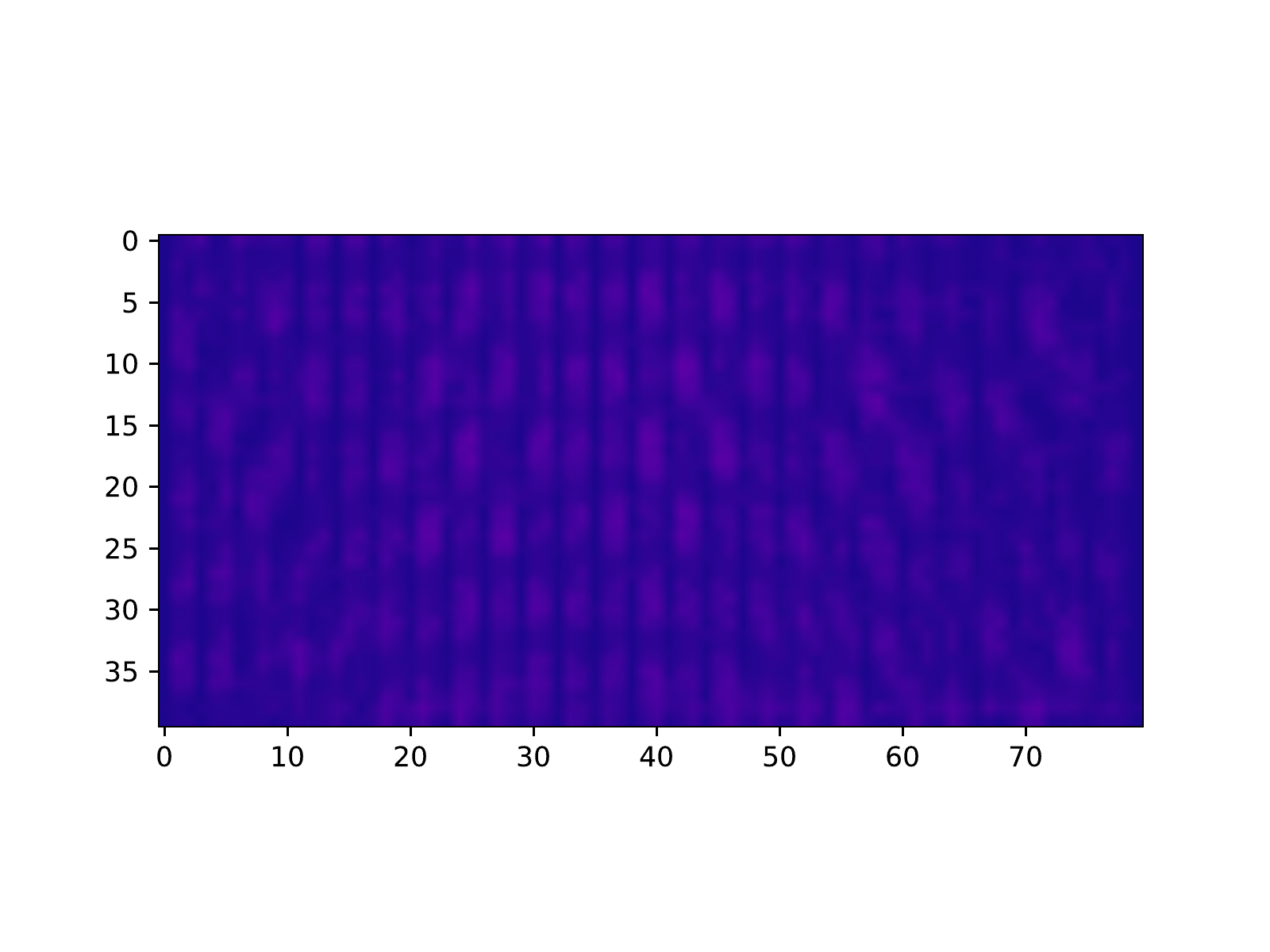}
	
	\vspace{1.5mm}
	\includegraphics[width=0.56\linewidth, trim= 58 148 50 150, clip]{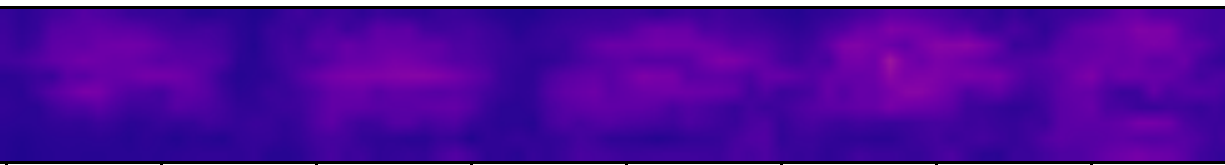}
	\includegraphics[width=0.28\linewidth, trim= 58 125 50 120, clip]{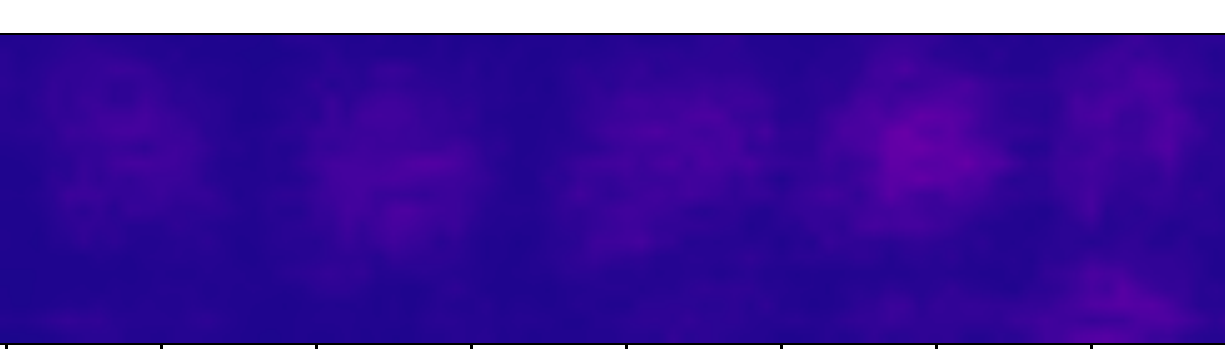}
	\includegraphics[width=0.14\linewidth, trim= 58 80 50 80, clip]{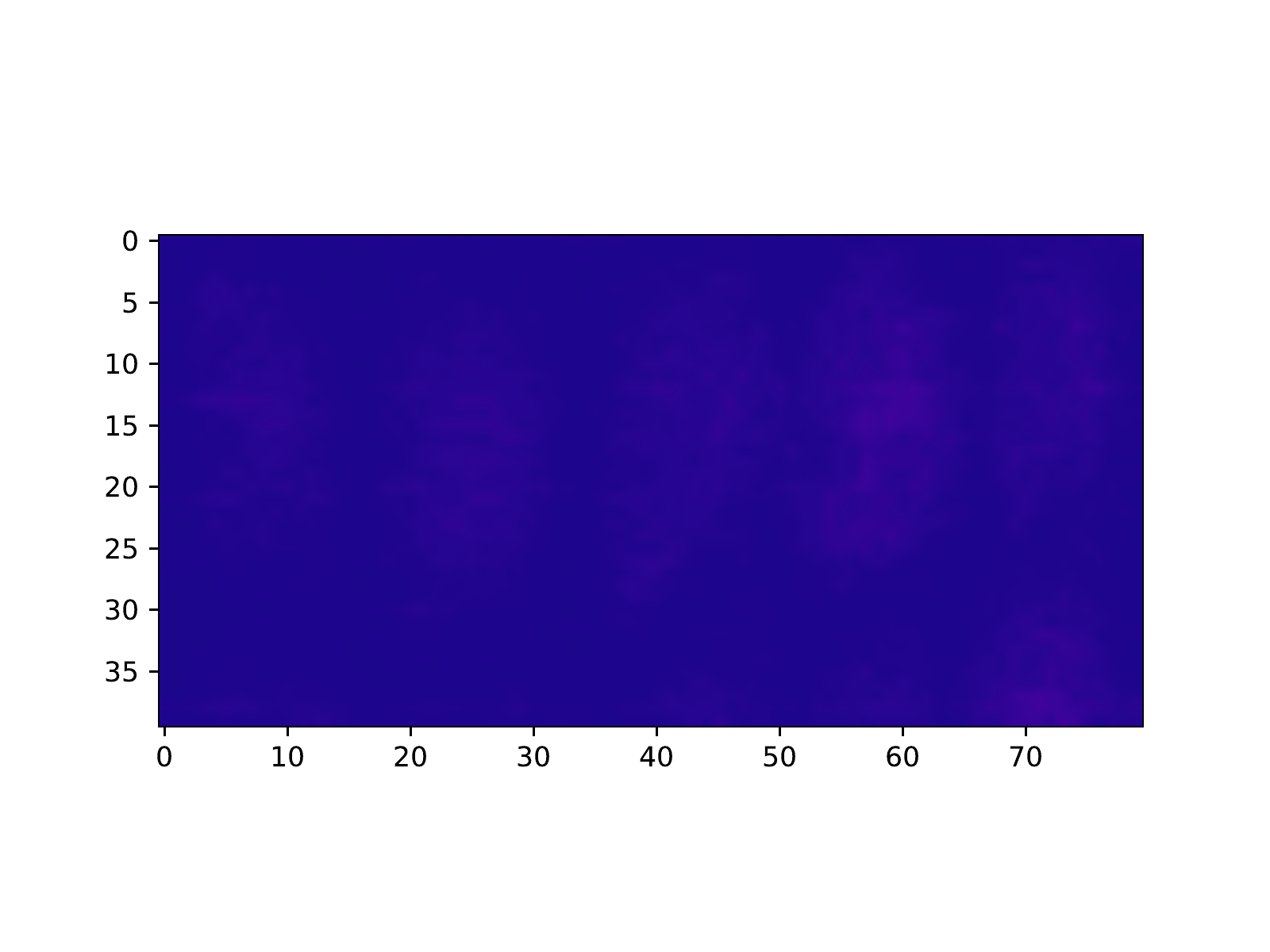}
	
	\vspace{1.5mm}
	\includegraphics[width=0.56\linewidth, trim= 58 148 50 150, clip]{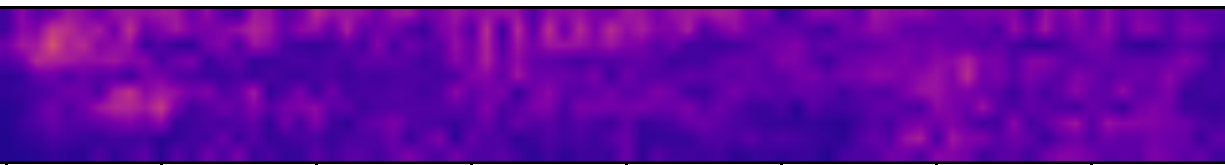}
	\includegraphics[width=0.28\linewidth, trim= 58 125 50 120, clip]{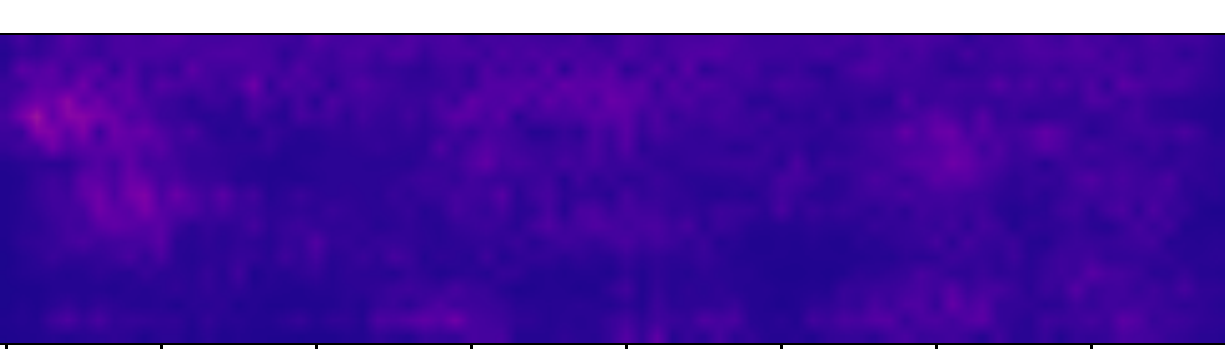}
	\includegraphics[width=0.14\linewidth, trim= 58 80 50 80, clip]{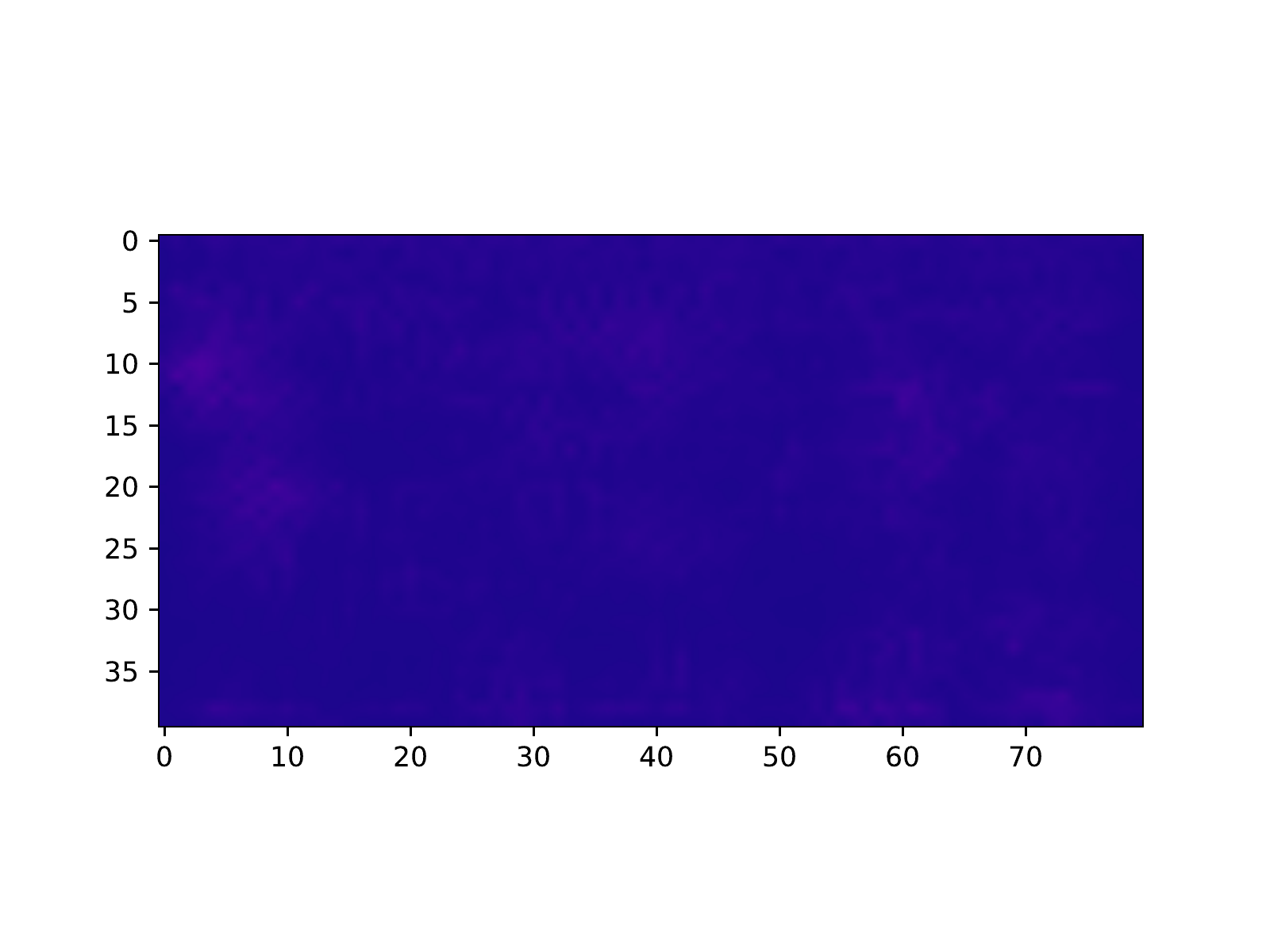}
	\subcaption*{Temporal frequency.}
	\end{subfigure}
	\caption{ 
	    A comparison of wavelet-based features between ground-truth data (a) and the generated data with MAE loss (b) and our surfNet (c). For this, the spatial spectrum of a selected frame was generated, together with the temporal spectrum for the corresponding sequence. The respective columns implicitly indicate the frequency (from left, higher to right, lower frequency). This clearly shows how the high-frequency features are revealed by the wavelet transform and partially reconstructed by our method.
	}
    \vspace{-3mm}
\label{fig:2d_freq_sample}
\end{figure*}

\begin{figure*}[t]
	\centering
    			    \begin{subfigure}[c]{0.91\textwidth}
	\begin{overpic}[width=\linewidth, trim= 50 280 50 450, clip]{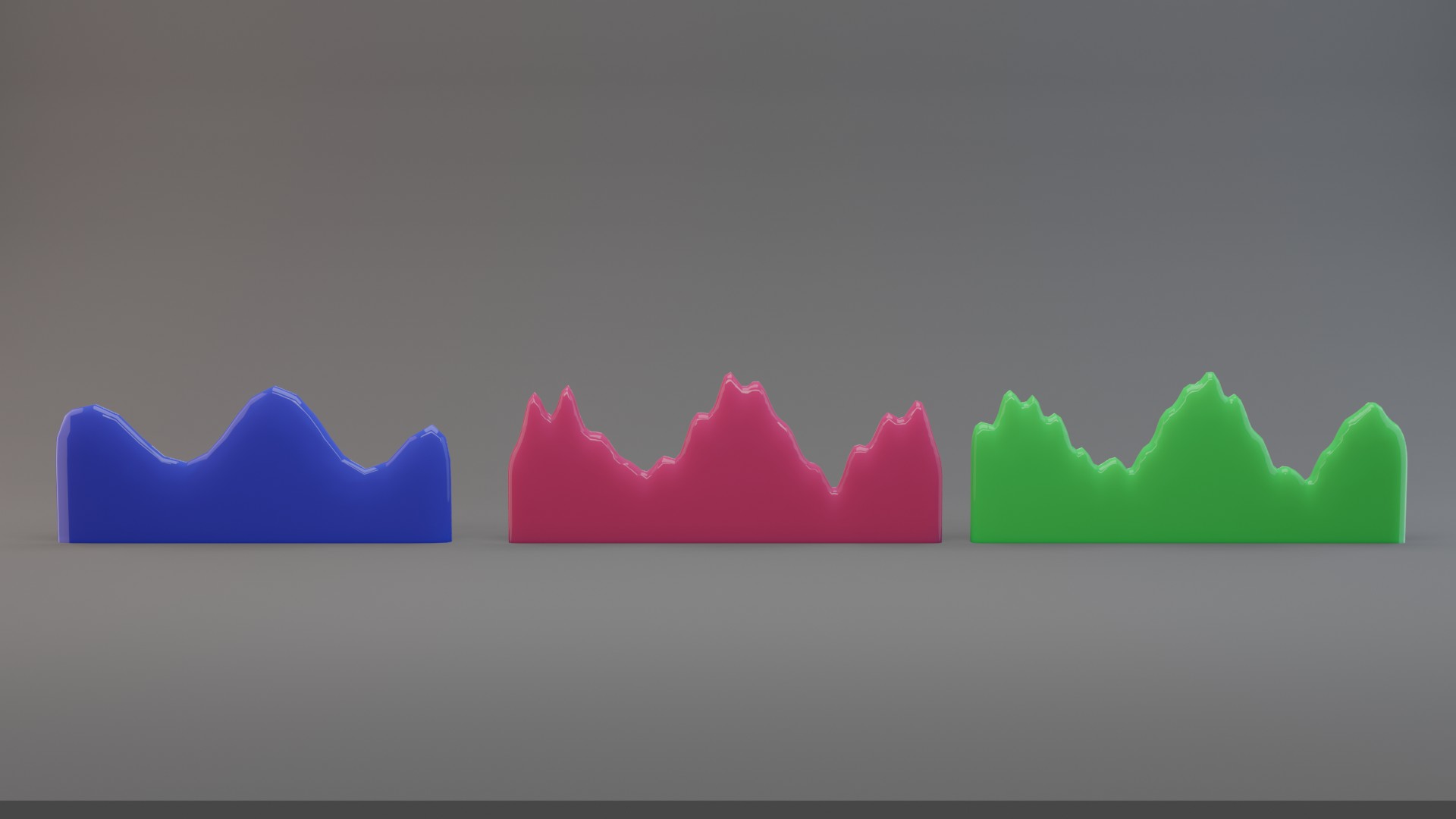} 
	    \put(12,1.5){\small \color{white}{\textit{Source}}}
	    \put(43,1.5){\small \color{white}{\textit{Ground-truth}}}
	    \put(80,1.5){\small \color{white}{\textit{surfNet}}}\end{overpic}
	    
	\vspace{1mm}
		\end{subfigure}
    \begin{subfigure}[c]{0.30\textwidth}
	\begin{overpic}[width=\linewidth, trim= 1260 280 50 450, clip]{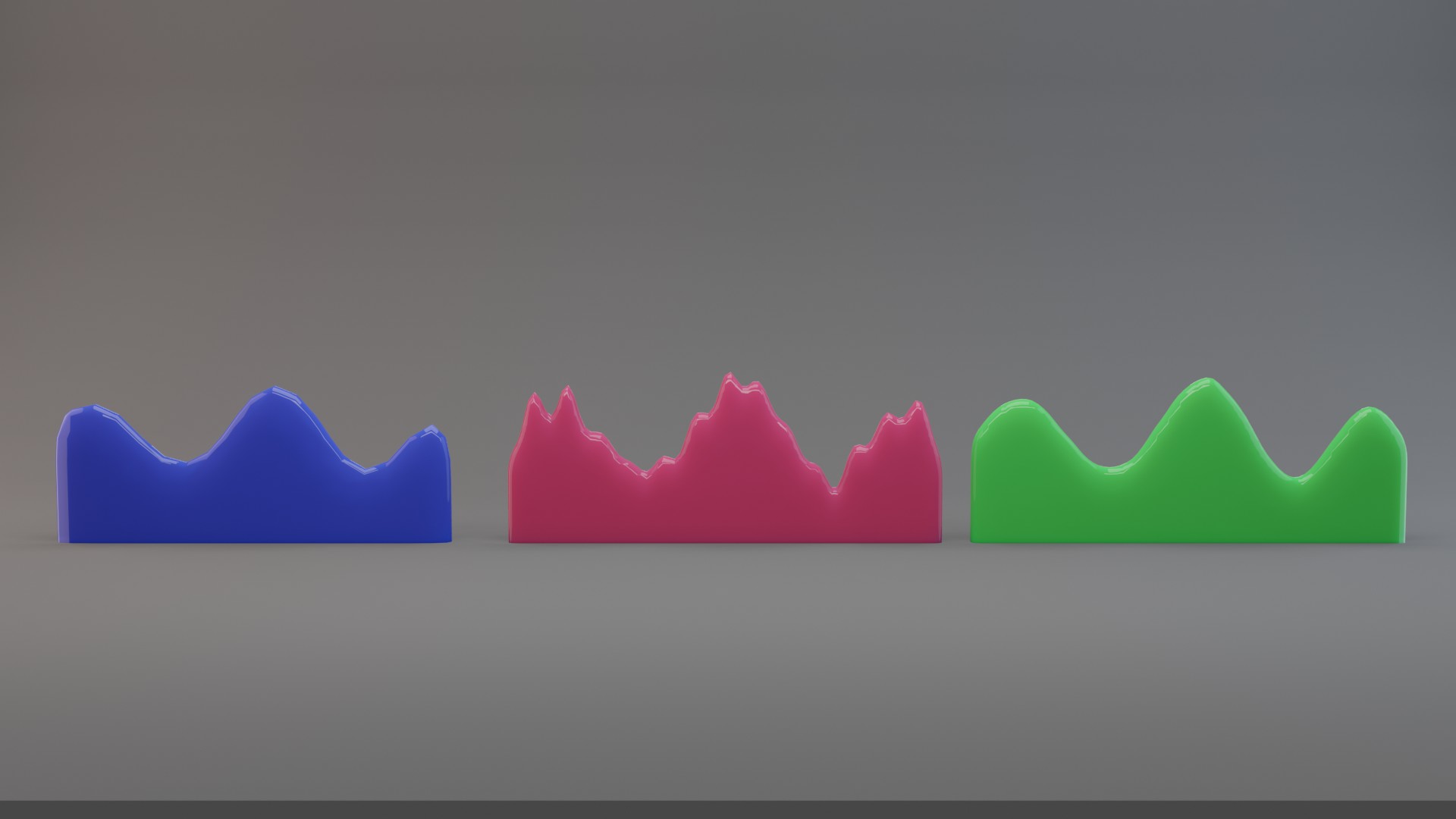} 
	    \put(45,5){\small \color{white}{\textit{MAE}}}\end{overpic}
	\end{subfigure}
    \begin{subfigure}[c]{0.30\textwidth}
	\begin{overpic}[width=\linewidth, trim= 1260 280 50 450, clip]{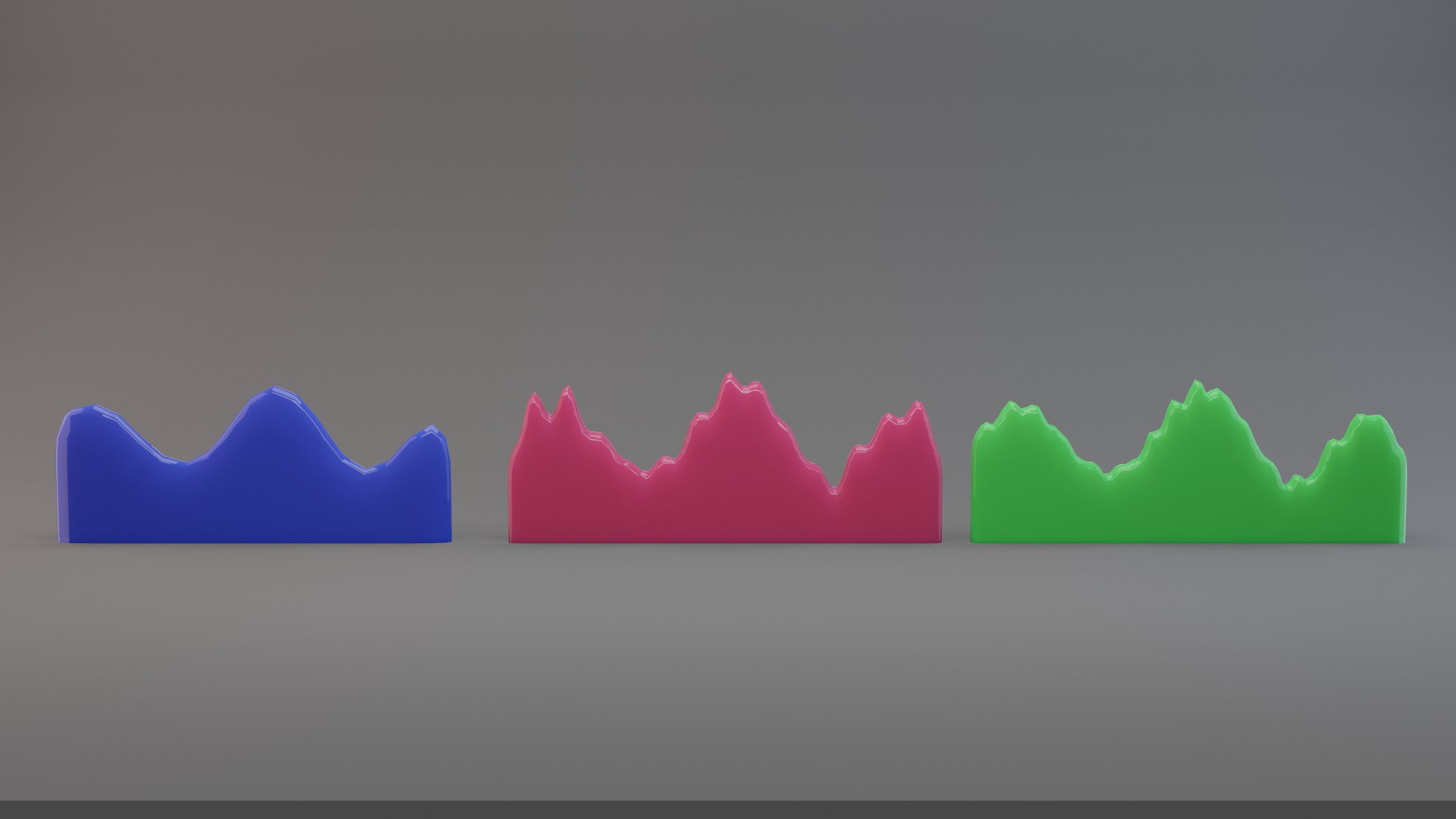} 
	    \put(40,5){\small \color{white}{\textit{RFFT}}}\end{overpic}
	\end{subfigure}
    \begin{subfigure}[c]{0.30\textwidth}
	\begin{overpic}[width=\linewidth, trim= 1260 280 50 450, clip]{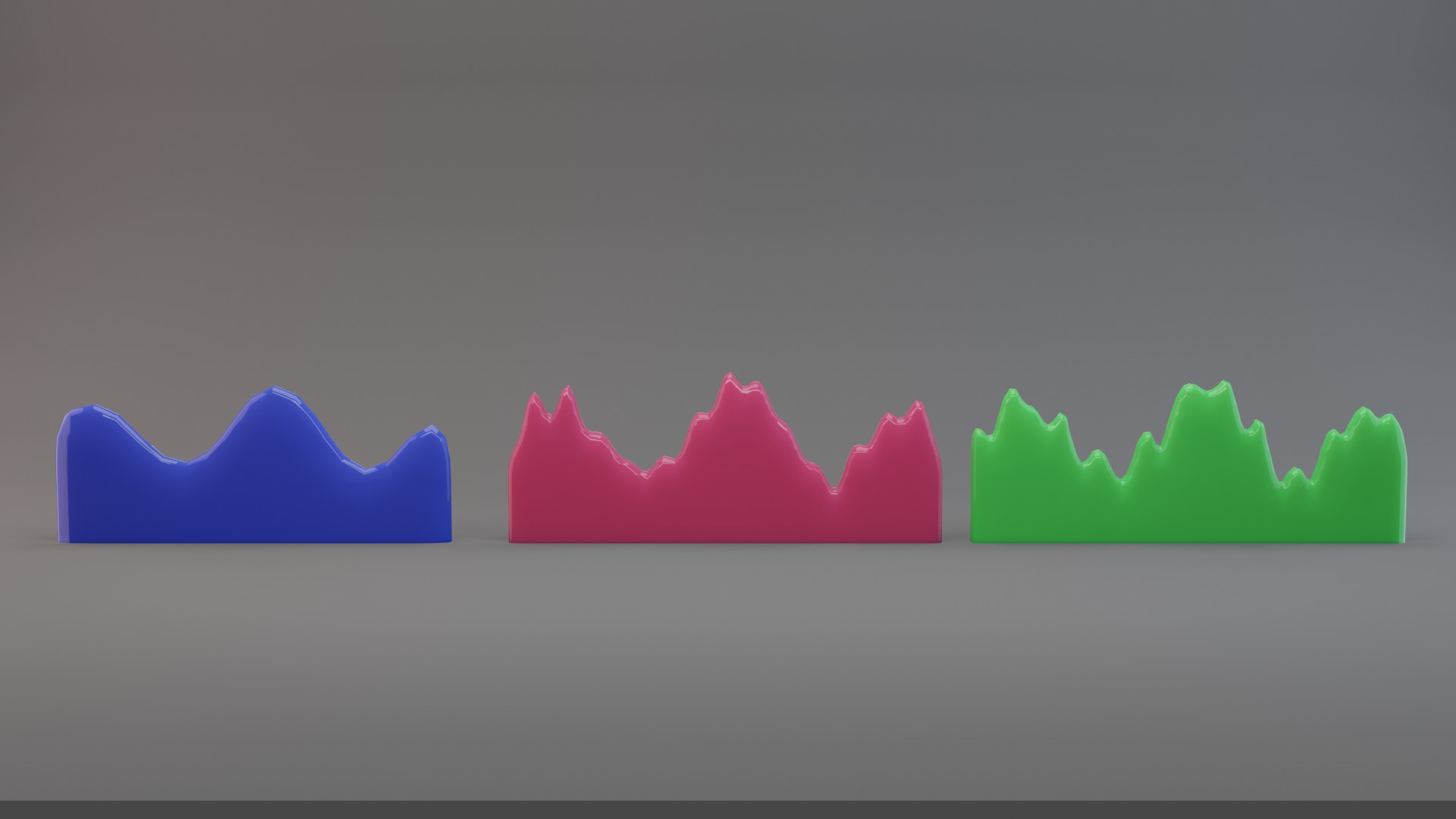}
	    \put(42,5){\small \color{white}{\textit{GAN}}}\end{overpic}
	\end{subfigure}
	\caption{ 
	   Comparison of a frame from a test run with the synthetic 2D data set. The input is shown in blue, ground-truth in red, and predictions in green.
	}
	\label{fig:qual_2d}
    \vspace{-1mm}
\end{figure*}

\subsection{Surface Waves}

We first consider the controlled 2D data. Training is performed with the full data set of 10000 samples with 10 frames, and
we evaluate the resulting models on 10 simulations that are not part of the training set with 40 frames each. The data consists of one-dimensional sine waves, from which a 2D surface was generated. This intuitive setup allows us to easily evaluate the spectrum of the surface. 

In the following evaluations we compare our approach with a Fourier and a GAN-based loss formulation. 
The Fourier loss works similarly to our wavelet Loss, except that we use a real Fourier transform (RFFT) instead of the wavelet transform. The RFFT version serves mainly as a sanity check, since the setup is based on a combination of periodic sine waves and thus falls exactly within the scope of the Fourier transform. This makes the task much easier for the RFFT. With more complex problems, the Fourier transformation comes to its limits, since the signals no longer occur periodically. Thus, our goal is to achieve comparable results with the more flexible wavelet loss.
The comparison with the GAN version~ \citep{tempoGAN}, however, serves to show that our method is similarly powerful while retaining the advantage that the loss function is interpretable. 
As part of our evaluation, we analyse the frequency of the surface. In \Figref{fig:freq_eval}, we compare the spectrum of the data generated by the networks with those from the ground truth. 
\Figref{fig:freq_eval}(a) shows the spectrum of generated data of a variant of our network trained with a MAE loss. It hardly shows any deflection in the high frequency range. 
If we now use our wavelet loss (\Figref{fig:freq_eval}(d)), we can clearly see that higher frequencies are more present in the generated data, making the spectrum similar to that of the ground-truth.
In \Figref{fig:freq_eval}(b) and \Figref{fig:freq_eval}(c) you can see the spectrum obtained with the RFFT setup and the GAN setup. 

In Table~\ref{tab:freq_eval} we also compare the mean absolute error values of the data quantitatively in Euclidean space and in frequency space.
The mean errors likewise illustrate that the MAE version
is not able to reconstruct the high frequencies of the target function. 

Another illustration of the superiority of our method over the baseline is shown in \Figref{fig:2d_freq_sample}. It shows the wavelet transformation from a chosen sample. From top to bottom, it compares ground-truth, a network with MAE loss, and the surfNet result. While the MAE-based variant is not able to reconstruct fine details, the surfNet can recover these, as highlighted by the wavelet representation. The high-frequency blocks clearly illustrate the differences into terms of reconstruction accuracy.

As a qualitative evaluation,
\Figref{fig:qual_2d} shows visual examples of the 2D test data set. While a version based on MAE generates very smooth images in the random data set, the surfNet versions are able to reconstruct the jagged edges. In direct comparison with the ground-truth data there are differences, but this is due to the randomness of the data. Therefore, the exact solution cannot be reconstructed, but the surfNet is still able to generate a very plausible solution.

\subsection{Mass-Spring System}
\begin{table}[ht!]
	\centering \footnotesize
\begin{tabular}{l|c|c}
        & MAE & surfNet \\
    \hline
    \hline
    \textbf{MAE} & 0.017  & 0.020 \\
    \hline
	\textbf{Spatial freq. MAE}  & 0.813 & 0.809 \\
    \hline
	\textbf{Temp. freq. MAE}  & 0.523 & \textbf{0.466} \\
\end{tabular}
\caption{
    Mean error measurements for 2D test scenarios.
}
\label{tab:2d_eval}
\end{table}

\begin{figure*}[h]
	\centering
    \begin{subfigure}[c]{0.2\textwidth}
	\includegraphics[width=\textwidth, trim= 0 300 0 300, clip]{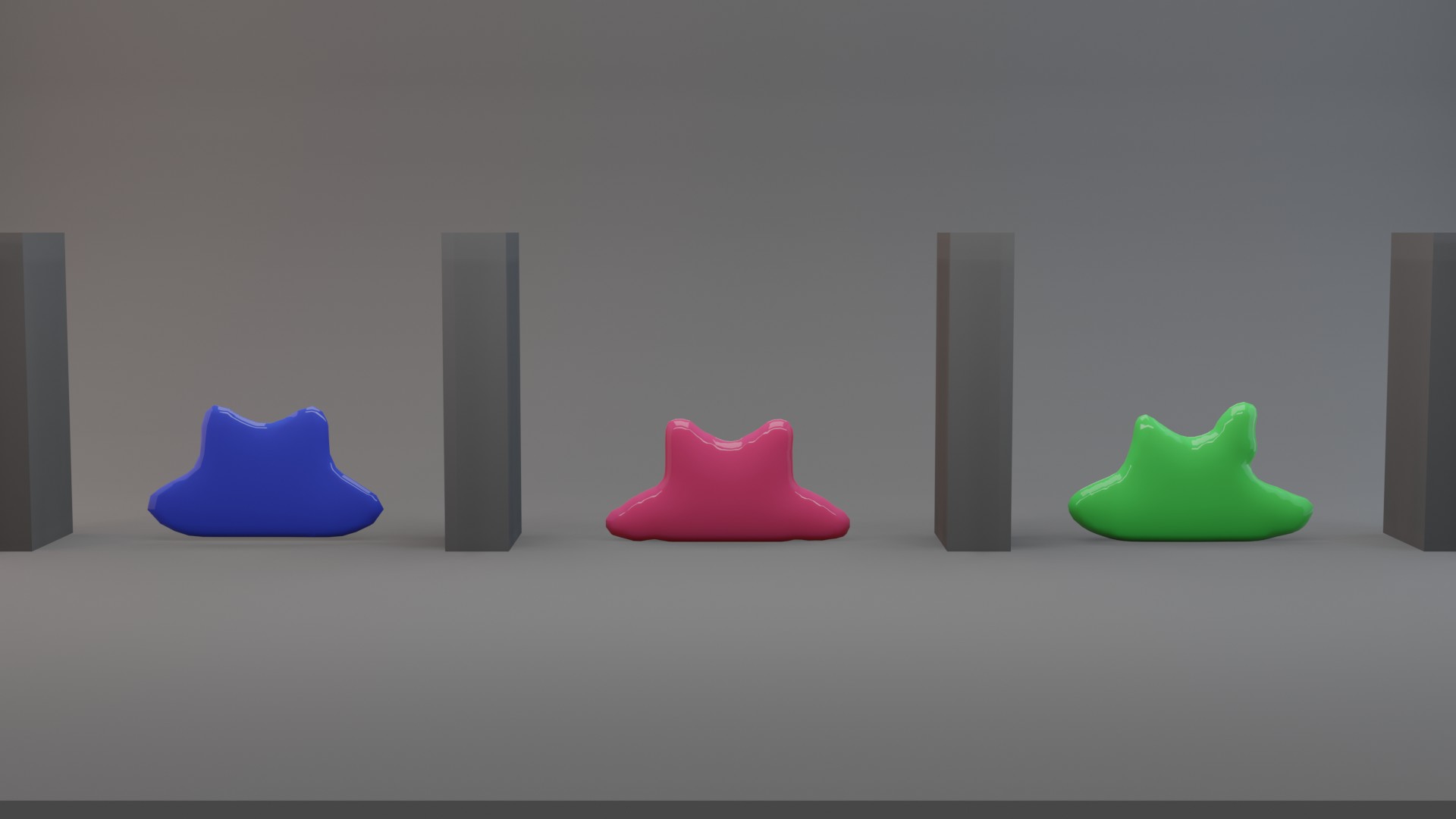}  
	\caption*{\scriptsize (a) Frame 45.}
	\includegraphics[width=\textwidth, trim= 0 300 0 300, clip]{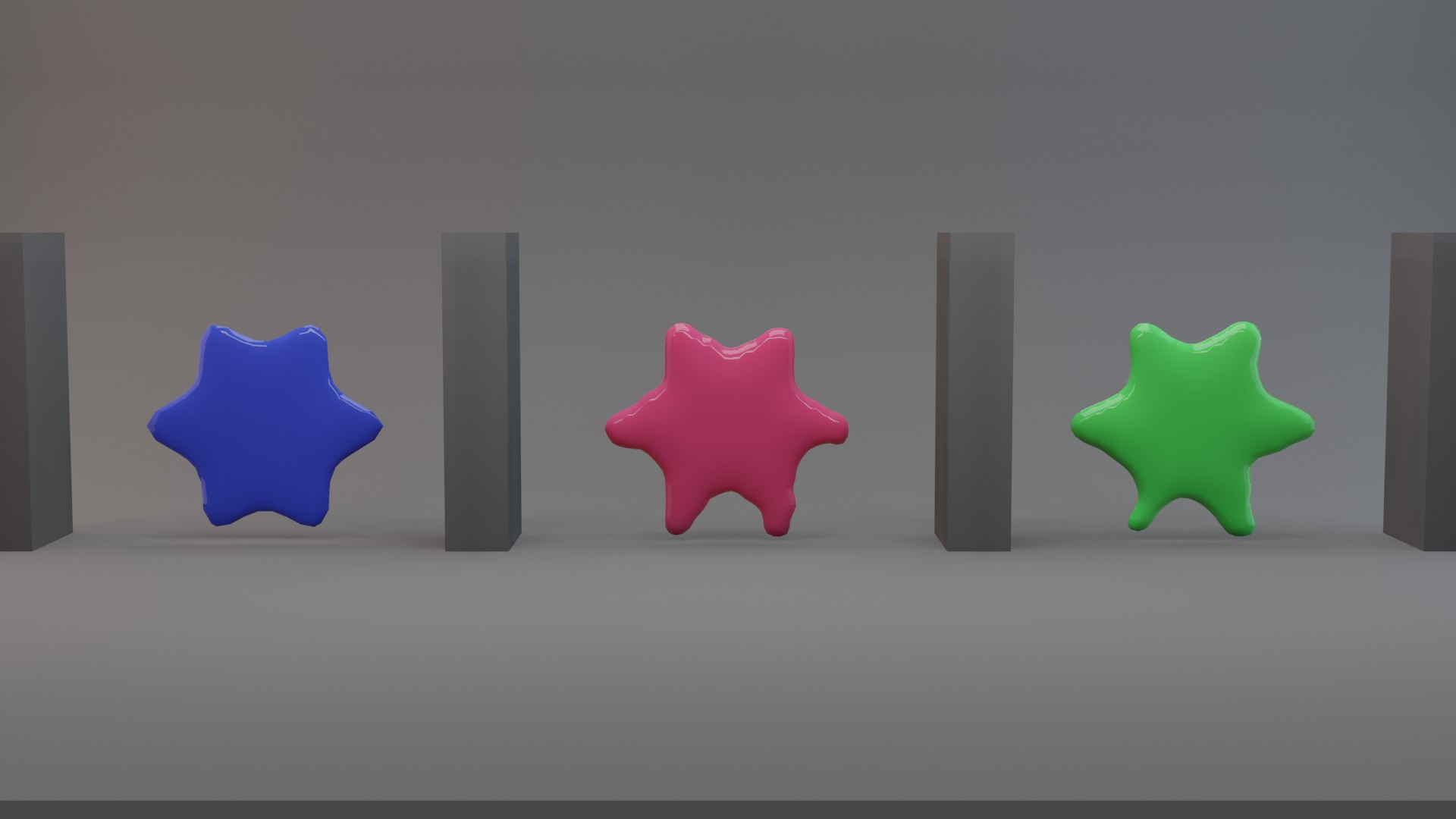}  
	\caption*{\scriptsize (a) Frame 66.}
	\end{subfigure}
    \begin{subfigure}[c]{0.57\textwidth}
	\includegraphics[width=\textwidth, trim= 0 300 0 300, clip]{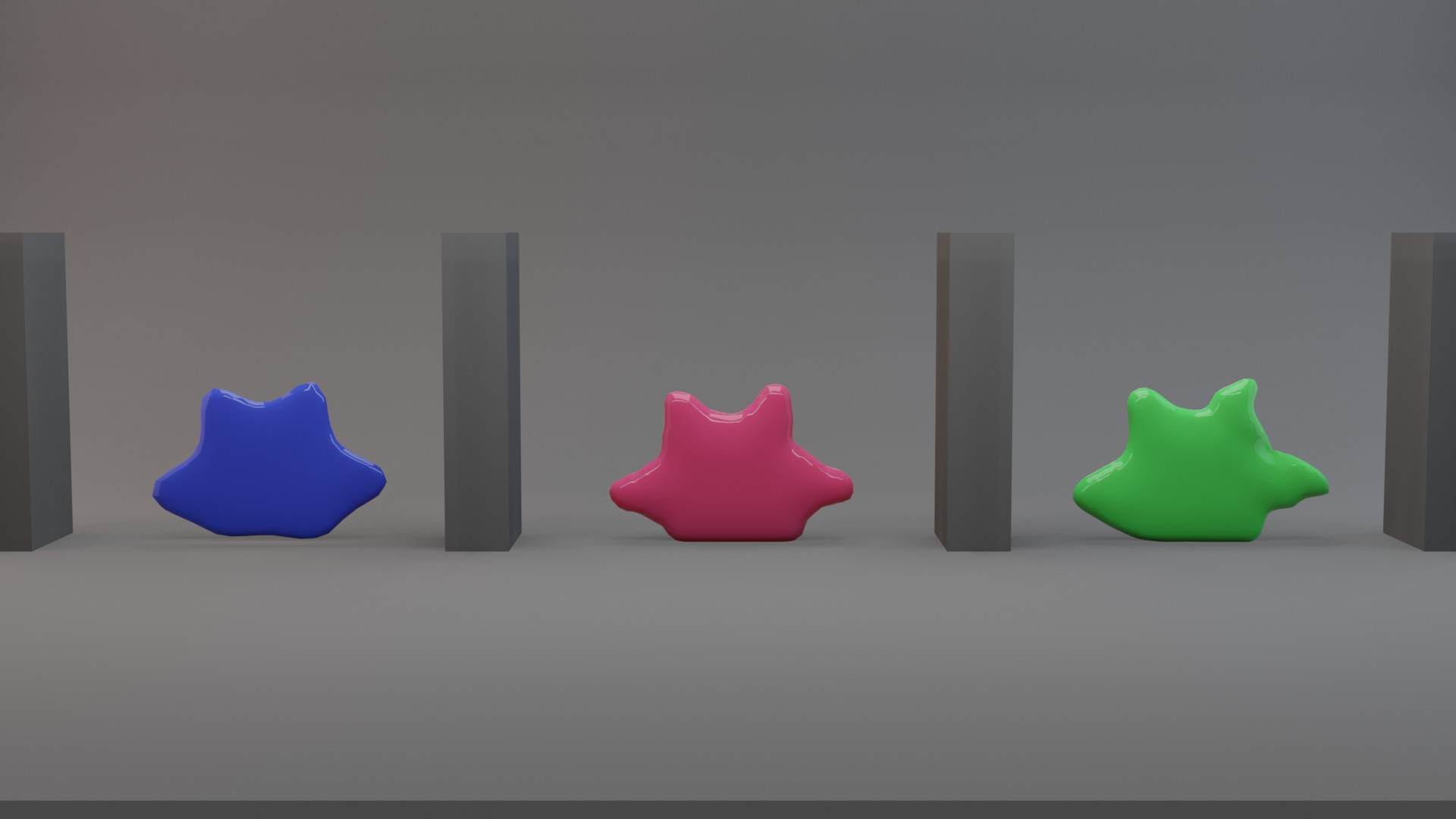} 
	\caption{\scriptsize Frame 123 surfNet.} 
	\end{subfigure}
    \begin{subfigure}[c]{0.2\textwidth}
	\includegraphics[width=\textwidth, trim= 1250 300 0 300, clip]{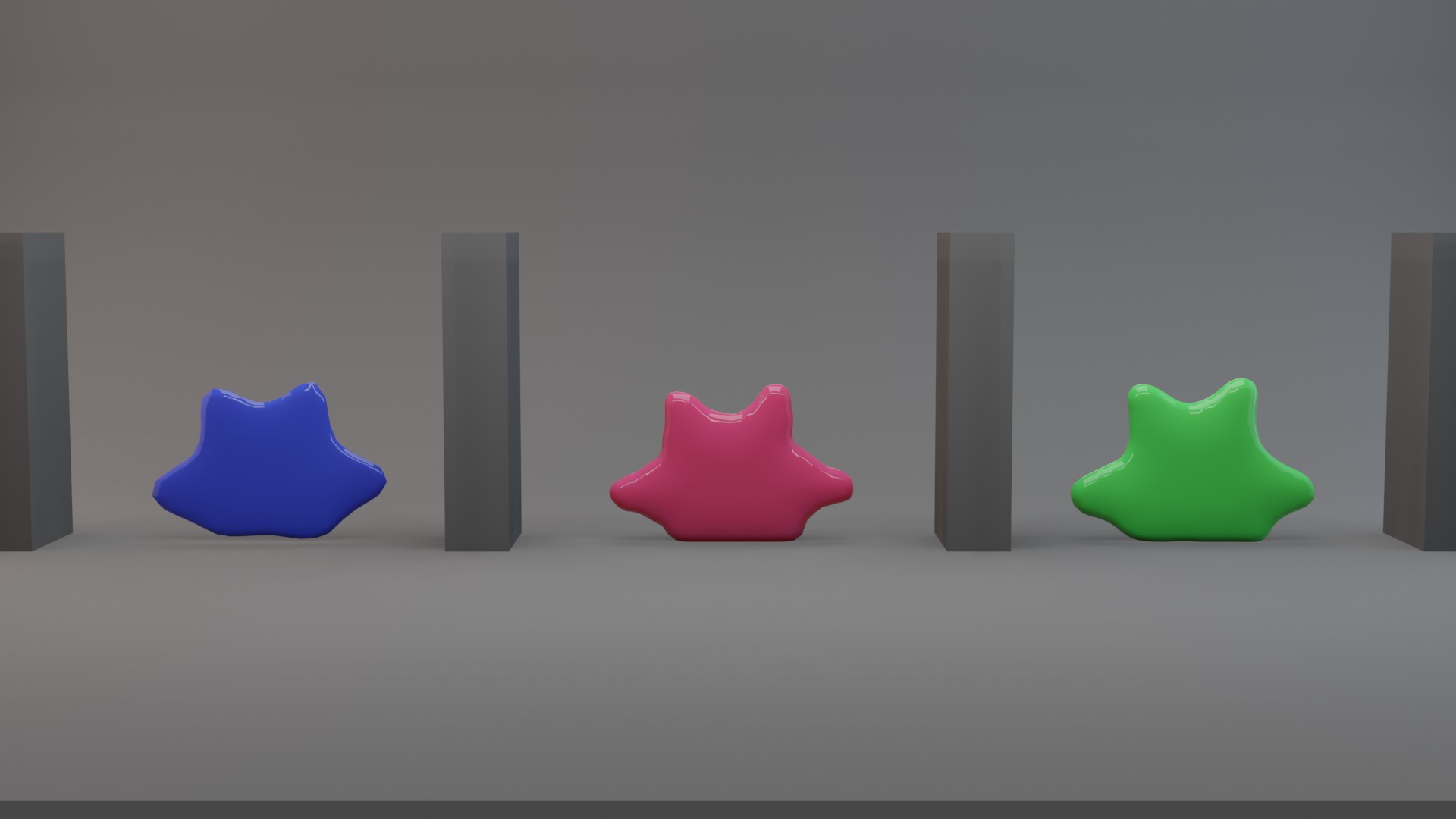} 
	\caption{\scriptsize Frame 123 MAE.} 
	\end{subfigure}
	\caption{ 
	   Comparison of 3D frames with a surfNet model (a) and with a MAE-based generator (b).
	   The input is shown in blue, ground-truth in red, and predictions in green. 
	}
	\label{fig:qual_2d2}
\end{figure*}

For a more complex test, we considered the behavior of different elastic 2D shapes simulated on the basis of mass-spring systems. The low-resolution version, which serves as input, provides the rough motion with a static shape. The network has the task to reconstruct the oscillations caused by the elastic material.
As you can see in \Figref{fig:qual_2d2}, the generator is able to reconstruct the details to some extent for both variants. The results with the wavelet loss (\Figref{fig:qual_2d2}) have, as expected, more high-frequency features, while they are more subtle with MAE. Nevertheless, the generated details of our method do not always seem to match the physical behavior of the data. Apart from that, the differences to the ground truth are still very large. It seems like the network in this case is overfitting to certain high-frequency details rather than generalizing to the underlying physical behavior.

\begin{figure*}[h]
	\centering
    \begin{subfigure}[c]{0.168\textwidth}
	\includegraphics[width=\textwidth, trim= 0 0 250 0, clip]{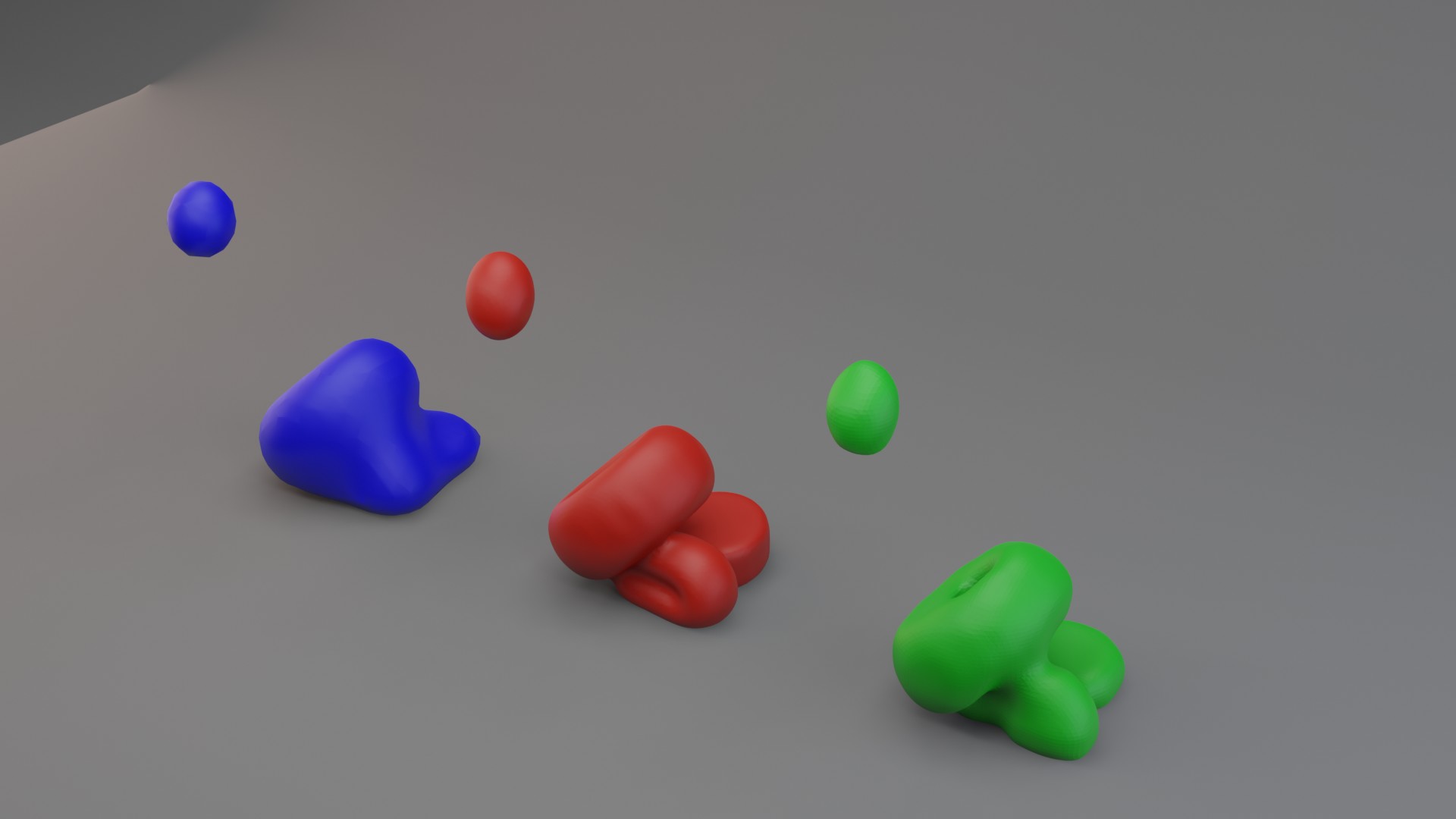}  
	\caption*{\scriptsize (a) Frame 98.}
	\includegraphics[width=\textwidth, trim= 0 0 250 0, clip]{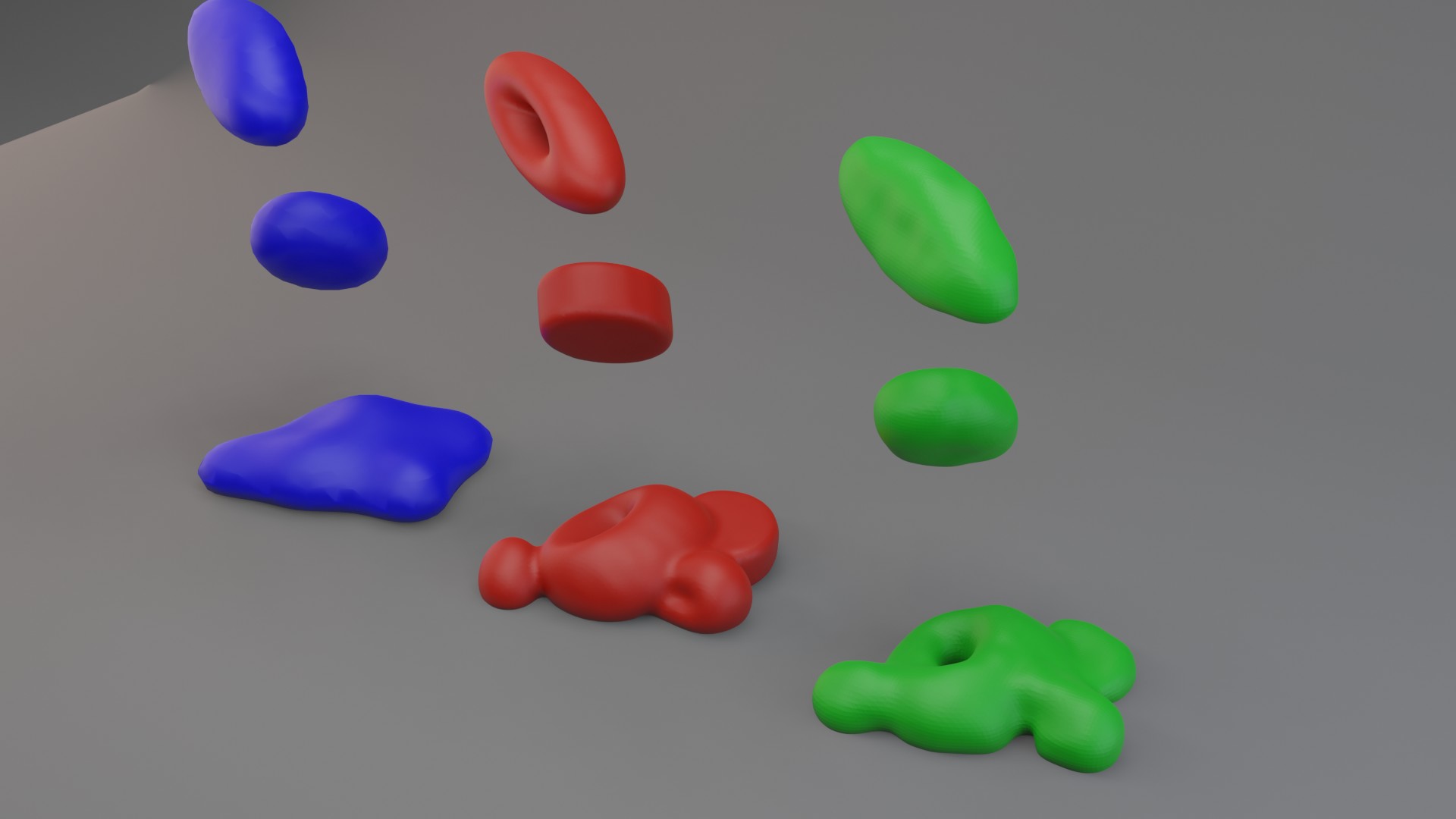}  
	\caption*{\scriptsize (a) Frame 128.}
	\end{subfigure}
    \begin{subfigure}[c]{0.4\textwidth}
	\includegraphics[width=\textwidth, trim= 0 0 250 0, clip]{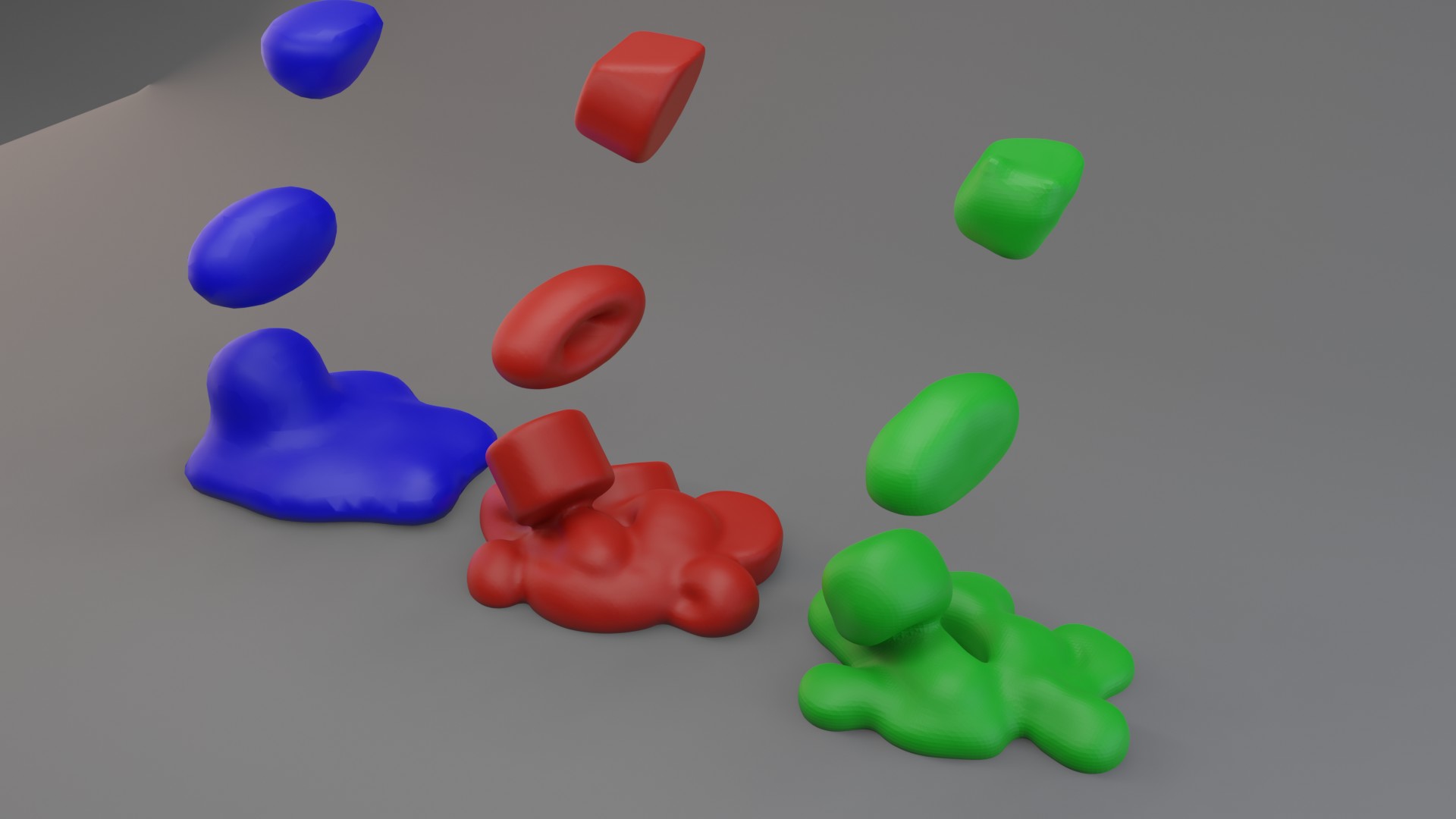} 
	\caption{\scriptsize Frame 164 surfNet.} 
	\end{subfigure}
    \begin{subfigure}[c]{0.4\textwidth}
	\includegraphics[width=\textwidth, trim= 0 0 250 0, clip]{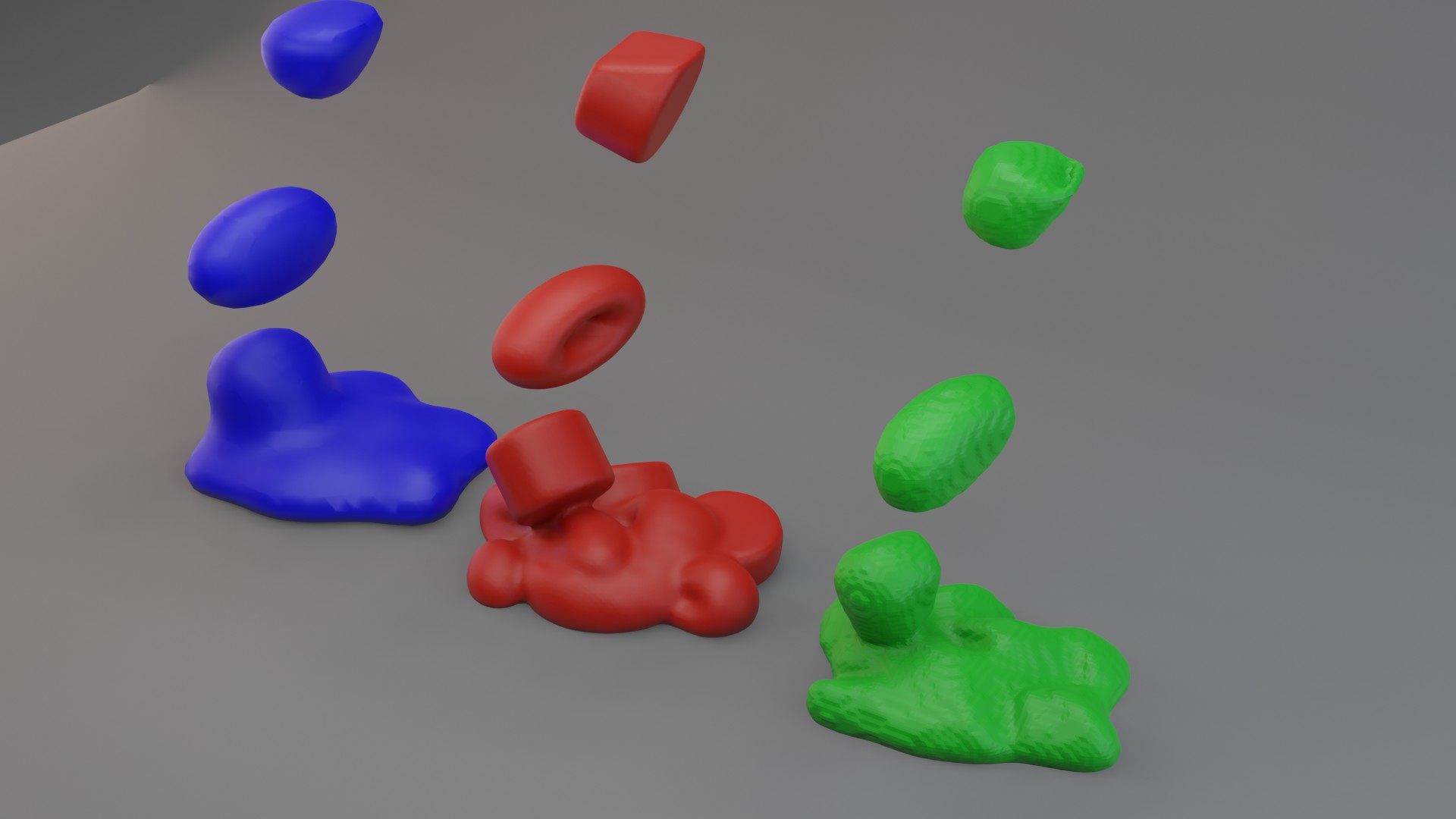} 
	\caption{\scriptsize Frame 164 MAE.} 
	\end{subfigure}
	\caption{ 
	   Comparison of 3D frames with a surfNet model (a) and with a MAE-based generator (b).
	   The input is shown in blue, ground-truth in red, and predictions in green. 
	}
	\label{fig:qual_3d}
\end{figure*}
\subsection{SPH Simulation}
\begin{table}[ht!]
	\centering \footnotesize
\begin{tabular}{l|c|c}
        & MAE & surfNet \\
    \hline
    \hline
    \textbf{MAE} & 0.087  & 0.107 \\
    \hline
	\textbf{Spatial freq. MAE}  & 0.962 & \textbf{0.695} \\
    \hline
	\textbf{Temp. freq. MAE}  & 0.464 & \textbf{0.250} \\
\end{tabular}
\caption{
    Mean error measurements for 3D test scenarios.
}
\label{tab:3d_eval}
\end{table}

A further test case based on physical simulation we use data based on a high viscosity 3D SPH simulation. We compare our method with a MAE based approach, whose is not able to reconstruct the outlines and details of the ground-truth data and is temporally incoherent. 
Our method achieves slightly better results, as can also be seen in Table \ref{tab:3d_eval}. Both, errors in the spatial and temporal frequencies are smaller for our method.

\Figref{fig:qual_3d}(a) shows that our prediction is able to reconstruct and preserve most of the details, even if they are not present in the input. We have deliberately chosen frames after a long run-time (98-164 time-steps) to show that details can persist and that they are the result of complex, physically-based behavior over multiple frames. Again, we compared our method with a simpler MAE-based generator. 
With MAE, the results tend to have unsightly artifacts that lead to flickering over time, as can be seen in \Figref{fig:qual_3d}(b). The wavelet variant (\Figref{fig:qual_3d}(c)) delivers much more coherent results, but similar to the 2D case, the amount of additional detail was not much higher.

\section{Conclusion}
\vspace{-1mm}
Our method sets the foundation for a generic, frequency-based loss function that can find application in many spatial and temporal geometric learning problems. This is interesting especially for physical problems due to the structural nature of the data. The effectiveness of our method has been well demonstrated by evaluations on a synthetic data set. The evaluation of more complex problems was less successful, but we are convinced that due to the generic and simple nature of our method it can still be useful for more complex problems if applied correctly.
Our method provides a first step towards evaluation and synthesis of physical space-time processes, and could be employed for other phenomena such as turbulence \cite{ling2016reynolds} or weather \cite{zaytar2016sequence}. Furthermore, it will be interesting to employ it in conjunction with other frequency-based representations  \cite{sitzmann2020implicit}.

\pagebreak

\bibliography{fluids}
\bibliographystyle{iclr2022_conference}

\clearpage

\appendix

\section{Appendix}

\subsection{Implementation Details}
\label{sec:impl_details}
The main part of the generator consists of four blocks containing two convolution layers and a residual connection. The feature count of the convolution layers per block is as follows: $[8, 16], [32, 32], [16, 8], [1, 1]$. The kernel size is $5\times5\times5$ for all layers.
It is important to mention that for the first convolution of our network we do not use zero padding as usual but mirror padding. This is because we work with tiles from the training data and not with the complete frame and zero padding falsifies the values at the transitions between the tiles.

\subsection{Training Details}
\label{sec:train_details}
For the training we have implemented our network with the Tensorflow Framework. We use an Adam optimizer with a learning rate from 0.0005 to 0.00001 with piece-wise constant decay and a batch size of 16 for the 2D tests and 4 for the 3D tests for 50k iterations. All other weights are initialized with the respective standard initializers of Tensorflow version 2.1. The weighting factors $\alpha$ and $\beta$ of \Eqref{eq:total} are set to $100.0$ and $10.0$ correspondingly.

\begin{figure*}[h]
	\centering
	\includegraphics[width=1.0\textwidth,]{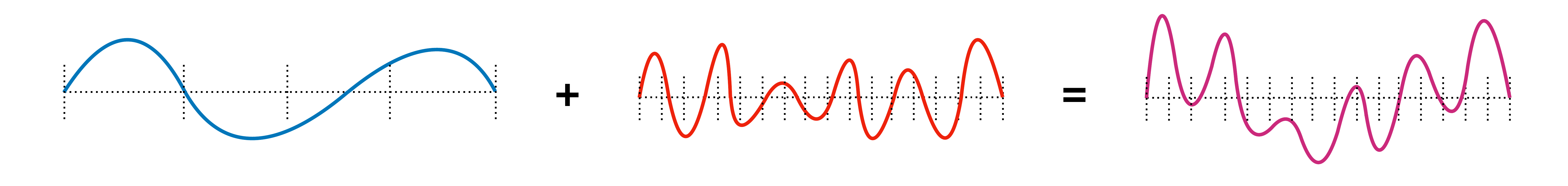}    
	\caption{ 
	    In blue a wave with a randomly varying low frequency $f_l$, which can still be represented by the low-resolution sampling and serves as input data set for our synthetic data set. In red, on the other hand, we see a wave with a high frequency $f_h$ that can only be correctly represented with much higher resolution. Combined with the low-frequency version, we get the ground-truth data (violet) for the synthetic data set. 
	}
	\label{fig:synthetic1}
\end{figure*}

\subsection{Synthetic Data Set}
\label{sec:synth_data}
We use a synthetic data set to test and evaluate different aspects of our approach. The data set is designed to have a simple, clearly defined behavior, and such that the frequency spectrum of the surface can be evaluated reliably. Therefore, we use a horizontal wavy surface based on a 1D sine function $s_t(x)$ with randomly varying frequency $f_l \in (0,\frac{M}{2})$:
\begin{equation}
    s_0(x) = sin(\frac{\pi f_l(x) x}{M}), x \in [0,M),
\end{equation}
where $M$ is the resolution of the low-resolution data.

To make a time sequence out of this, we use a simple wave equation:
\begin{equation}
    \frac{\delta^2 s}{\delta t^2} = \frac{\delta^2 s}{\delta x^2} ,
\end{equation}
Discretizing this we can calculate the vertical velocity of our surface as follows:
\begin{equation}
    v_t(x) = v_{t-1}(x) + \frac{2*s_t(x)-s_t(x-\Delta x)+s_t(x+\Delta x)}{\Delta x} ,
\end{equation}
where $v_0(x)=0$. Given the velocity we can now calculate the next frame as follows:
\begin{equation}
    s_t(x) = s_{t-\Delta t}(x) + \Delta t v_{t}(x) ,
\end{equation}

For the high resolution data set, i.e, the targets to be learned,
we use the same low resolution wave as base, and modulate it with a high frequency component (\Figref{fig:synthetic1}):
\begin{equation}
    t_0(x) = sin(\frac{\pi f_l(x) x}{kM}) + sin(\frac{\pi f_h(x) x}{kM}) , x \in [0,kM),
\end{equation}
where $k$ is the chosen up-sampling factor and the frequency $f_h(x)$ is chosen so that it cannot be represented by the low-resolution version. According to the Nyquist-Shannon sampling theorem, the frequency should be higher than $\frac{M}{2}$ and below $\frac{kM}{2}$  (\Figref{fig:2d_sample}(a)). The generation of a sequence is done in the same way like for the low-resolution data.

Based on this setup, we generate two different data sets of high resolution data: one where we modulate with a fixed high frequency $f_h(x) = const$, whereas in the second version we vary the this high frequency component (\Figref{fig:2d_sample}(b)). Thus the first version represents a deterministic up-sampling, which a generator should be able to reconstruct perfectly, whereas the second version is ill-posed, i.e. several solutions are possible, and hence poses a much more difficult learning target. Both data sets consist of 10000 sequences with 10 frames each at the end.
While the deterministic version only serves as a sanity check, the second, randomized version shows how well the method can approximate the ground-truth distribution for ill-posed tasks like the actual super-resolution problem for physical simulation data. 

\subsection{Mass-Spring System Data}
\label{sec:sim_data_ms}
As a first, complex, physics-based example, we use data from an 2D elasto-plastic simulation with a mass-spring system. For this, we generate mass particles in star-shaped shapes with different scale and rotation and connect them with springs.
We then simulate how the generated objects fall to the ground. We map the particle-based simulations onto a grid to make them compatible with the network. We use a high-resolution grid for our ground-truth data and a low-resolution grid for our input data. The network thus has to learn to reconstruct from the low-resolution data the high-frequency details generated at the surface of the ground-truth simulation.
For the data, 900 particles are simulated over 400 time steps in 1000 different setups. The generated grids have a size of $64\times 64$ for the high resolution and $32 \times 32$ for the low resolution.

\subsection{Simulation Data}
\label{sec:sim_data}
For the generation of simulation data we use an SPH solver of the SPlisHSPlasH framework \citep{splishsplash}. There are different materials to choose from. With materials that exhibit high-frequency physical behavior, chaotic behavior occurs in some cases, such as splashes in water, which are typically very difficult to reconstruct. With more viscous materials, such as gel, details are mainly distinguished by folds and fine waves on the surface. The chaotic behaviour is very difficult to reconstruct and it is sometimes very difficult to understand how correct the behaviour is. For these reasons we focus more on materials like gel.
With gel, fine wrinkles can form on the surface which allows a good evaluation of the method. Another special feature is that details are persistent over time. Methods such as \citet{tempoGAN} cannot represent such details because they do not provide feedback in the network. This means there is no memory. Therefore, we use an plastic material with high viscosity, simulated with an advanced viscosity solver \citep{WKBB18}.

For the training we generate 60 different scenes with 300 frames per simulation. The time step corresponds to 100 frames per second. The scenes consist of randomly generated shapes that fall into a pool from different heights at random times. This creates interesting waves and folds on the surface. The ground-truth resolution is $160^3$. 
For the generation of the low-resolution data 
the ground-truth data is scaled down by the desired up-sampling factor $k$ and then smoothed with a Gaussian blur. This results in synchronous data pairs that can be used in a supervised setup. 
Before the training, the data is normalized over the whole data set, so that the data is in the value range between -1 and 1. 
Finally, we take advantage of the locality of our problem and only use excerpts from the training data frames in training. On the one hand, this saves memory, because the data is sometimes very large and can cause problems with the GPU memory, on the other hand it allows us to extract only relevant parts of the data. So in our example we can only consider the data in places where there is a surface. Finally we have the advantage to augment the data by overlapping the tiles we extract, so we can get a lot of information from only a few frames.

\end{document}